%% file: neurips_2026.tex
\documentclass{article}

\usepackage{etoolbox}
\usepackage{comment}
\newtoggle{neurips}
\togglefalse{neurips}
\newcommand{\neurips}[1]{\iftoggle{neurips}{#1}{}}
\newcommand{\arxiv}[1]{\iftoggle{neurips}{}{#1}}
\newcommand{\loose}{\looseness=-1}

\iftoggle{neurips}{
  \usepackage{neurips_2026}
}{
  \usepackage[preprint]{neurips_2026}
}

\usepackage[utf8]{inputenc} 
\usepackage[T1]{fontenc}    
\usepackage{hyperref}       
\usepackage{url}            
\usepackage{booktabs}       
\usepackage{multirow}       
\usepackage{graphicx}       
\usepackage{wrapfig}        
\usepackage{longtable}      
\usepackage{amsfonts}       
\usepackage{nicefrac}       
\usepackage{microtype}      
\usepackage{xspace}         
\usepackage{enumitem}       
\usepackage{xcolor}         
\usepackage{thm-restate}
\usepackage{caption}
\usepackage[nameinlink,capitalize]{cleveref}

\usepackage[suppress]{color-edits}
\definecolor{jtblue}{RGB}{25,95,170}
\definecolor{davidemagenta}{RGB}{180,30,140}
\addauthor{kn}{ForestGreen}
\addauthor{bc}{violet}
\addauthor{jt}{jtblue}
\addauthor{davide}{davidemagenta}
\addauthor{be}{red}
\addauthor{todo}{red}

\newcommand{\bc}[1]{\bccomment{#1}}
\newcommand{\jt}[1]{\jtcomment{#1}}
\newcommand{\be}[1]{\becomment{#1}}
\newcommand{\davide}[1]{\davidecomment{#1}}
\newcommand{\todo}[1]{\todocomment{#1}}

\newcommand{\skillLibraryName}{\texttt{CodeHack}\xspace}

\newcommand{\stderr}[1]{{\textcolor{gray}{\scriptsize$\pm$\,#1}}}

\declaretheorem[name=Remark, parent=section]{remark}

\title{Up and Down the Abstraction Ladder: \\ Code-Based Skills for Language Agents}

\author{%
  \textbf{Bartłomiej Cupiał\textsuperscript{1,6,*} \qquad
  Jens Tuyls\textsuperscript{2,*} \qquad
  Maciej Wolczyk\textsuperscript{3}} \\
  \textbf{Davide Paglieri\textsuperscript{4} \qquad
  Martin Klissarov\textsuperscript{5,7} \qquad
  Benjamin Eysenbach\textsuperscript{2}} \\
  \textbf{Piotr Miłoś\textsuperscript{1,8,9} \qquad
  Karthik R. Narasimhan\textsuperscript{2}}
}

\begin{document}

\maketitle

\makeatletter
\arxiv{%
  \begingroup
  \renewcommand{\thefootnote}{}
  \long\def\@makefntext#1{\noindent#1}
  \footnotetext[0]{%
    \raggedright%
    \setlength{\parindent}{0pt}%
    \setlength{\parskip}{0pt}%
    \mbox{\textsuperscript{1}\,University of Warsaw};
    \mbox{\textsuperscript{2}\,Princeton University};
    \mbox{\textsuperscript{3}\,IDEAS NCBR};
    \mbox{\textsuperscript{4}\,University College London};
    \mbox{\textsuperscript{5}\,McGill University};
    \mbox{\textsuperscript{6}\,AKCES NCBR};
    \mbox{\textsuperscript{7}\,Mila};
    \mbox{\textsuperscript{8}\,Mistral AI};
    \mbox{\textsuperscript{9}\,Institute of Mathematics, Polish Academy of Sciences}.
    \par
    \textsuperscript{*}\,Equal contribution. Corresponding author:
    \href{mailto:bartlomiej.cupial@gmail.com}{\texttt{bartlomiej.cupial@gmail.com}}.
    \par
    \hypertarget{title-origin-note}{\textsuperscript{\ensuremath{\dagger}}}\,Our title is inspired by the following blog:
    \url{https://worrydream.com/LadderOfAbstraction/}.
    \todo{maybe say something about how the blog relates}}
  \endgroup
}
\makeatother

\begin{abstract}
\neurips{
Acting and learning in long-horizon environments
\be{Is there a risk that LM readers will construe "environments" to mean that the method is only applicable to game-like tasks? Perhaps "agentic tasks"?}
remains challenging for language agents, especially on tasks requiring low-level action control such as in games~\citep{paglieri2025balrog}. Prior work suggests this is because primitive action control often provides poor grounding for language models and, when combined with long horizons, can be prohibitively expensive and hinder learning, exploration, and planning~\citep{wang2024executable,wangvoyager,ahn2022can,nachum2019does}.
\be{Check grammar in sentence above. What is the subject for the verb "can be"?}
\be{I think NeurIPS allows numbered citations, which can save a bit of space if needed}
Inspired by work on agent interfaces~\citep{yang2024swe} and horizon reduction~\citep{parkhorizon,SUTTON1999181}, we study the value of using higher-level, semantically meaningful code skills in addition to or instead of low-level primitives.
\be{I feel like we're missing a sentence around here stating what the big problem is. E.g., "Why hierarchical methods have proven exceptionally effective in games and robotic control problems, their efficacy remains unclear in agentic tasks. This paper aims to provide a systematic, empirical answer to this question.}
To do so, we focus on NetHack~\citep{kuttler2020nethack}, a long-horizon, challenging game environment. We follow a two-pronged approach. We first develop \skillLibraryName, a rich library of code-based skills with natural-language descriptions.
We compare agents that can only use primitive actions to only use semantic skills and those that can use both primitive actions and semantic skills. We evaluate these agents in three settings:\jt{this previous sentence needs grammar/spelling fix!}
zero-shot prompting, supervised fine-tuning, and reinforcement learning.
\be{Kinda unclear how the list at the end relates to NetHack. E.g., are these 3 different settings one can use NetHack to study?}
Averaged across a broad zero-shot NetHack evaluation, we find that, compared to primitives, skills provide nearly 3x improvement in game progression while cutting inference costs per episode by 86\%.
\be{What's the intuition behind why semantic skills result in less compute? Very superficially, I'd think "semantics = bigger model = more compute."}
Mixed control retains much of this benefit while preserving a path back down to low-level actions. Finally, in RL, we find skill-based agents learn faster
than agents acting on primitives as demonstrated by a 4.8x larger gain in dungeon level on average across models when using the same training budget.
We release \skillLibraryName together with training and evaluation code. \footnote{Our title is inspired by the following blog: \url{https://worrydream.com/LadderOfAbstraction/} \todo{maybe say something about how the blog relates}. \be{I'd recommend cutting this footnote and instead discussing this blog as a serious prior work in the related work section.}}

\be{Adversarial reviewer: you just wrote a bunch of code to solve certain subproblems in the game, so of course that code should be useful... Clarifying the degree to which code was written by LMs might help (and emphasizing that LMs are really really good at code generation, but it's often unclear how to harness that capability for solving non-coding problems).}
}

\arxiv{
Language agents struggle to act and learn in environments that require long sequences of low-level actions.
Code-based abstractions can make these agents more productive by letting them invoke reusable skills instead of repeatedly selecting individual actions. 
The code handles recurring local decisions, while the language model decides which skills to use and how to combine them. 
Yet abstractions are leaky, and situations beyond a skill’s capabilities may require a return to primitive actions. 
Motivated by this tradeoff between productivity and flexibility, we systematically study how code-based action abstraction affects the performance, inference cost, and learning of language agents.
We study this in NetHack, a challenging, long-horizon game environment, using \skillLibraryName, our library of code-based skills with natural-language descriptions.
We use this library to compare agents restricted to primitives with those using semantic skills alone or in combination with primitives.
We evaluate these agents in three settings: zero-shot prompting, supervised fine-tuning, and reinforcement learning. 
Across a broad zero-shot evaluation on NetHack, we find that compared with primitives, skills nearly triple game progression, while reducing inference cost per episode by 86\%.
Combining skills with primitives retains much of this benefit while preserving a path back down to low-level actions.
Finally, in RL, we find that skill-based agents learn significantly faster than agents acting on primitives, achieving a 7.2x larger average gain in dungeon level over the same training budget.
These results show that a supplied skill library can improve performance, efficiency, and learning, while retaining primitives provides flexibility when the library is insufficient. We release \skillLibraryName together with training and evaluation code. \hyperlink{title-origin-note}{\textsuperscript{\ensuremath{\dagger}}}
}

\end{abstract}

{\centering\small
Project page: \href{https://bartekcupial.github.io/abstraction-ladder/}{\nolinkurl{bartekcupial.github.io/abstraction-ladder/}}\par}

\arxiv{
\input{sections/intro}

}
\neurips{
\input{sections/intro_neurips}
}
\input{sections/preliminaries}

\input{sections/skill_library}
\input{sections/experimental_setup2}
\input{sections/results}
\input{sections/related_work}
\input{sections/conclusion}

\bibliographystyle{plainnat}
\bibliography{references}


\appendix
\crefname{appendix}{Appendix}{Appendices}
\Crefname{appendix}{Appendix}{Appendices}
\crefalias{chapter}{appendix}
\crefalias{section}{appendix}
\crefalias{subsection}{appendix}

\input{sections/appendixA}
\input{sections/appendixB}

\input{sections/appendixC}
\input{sections/appendixD}



\end{document}

%% file: sections/intro.tex
\section{Introduction}

\begin{wrapfigure}{R}{0.50\linewidth}
    \centering
    \includegraphics[width=\linewidth]{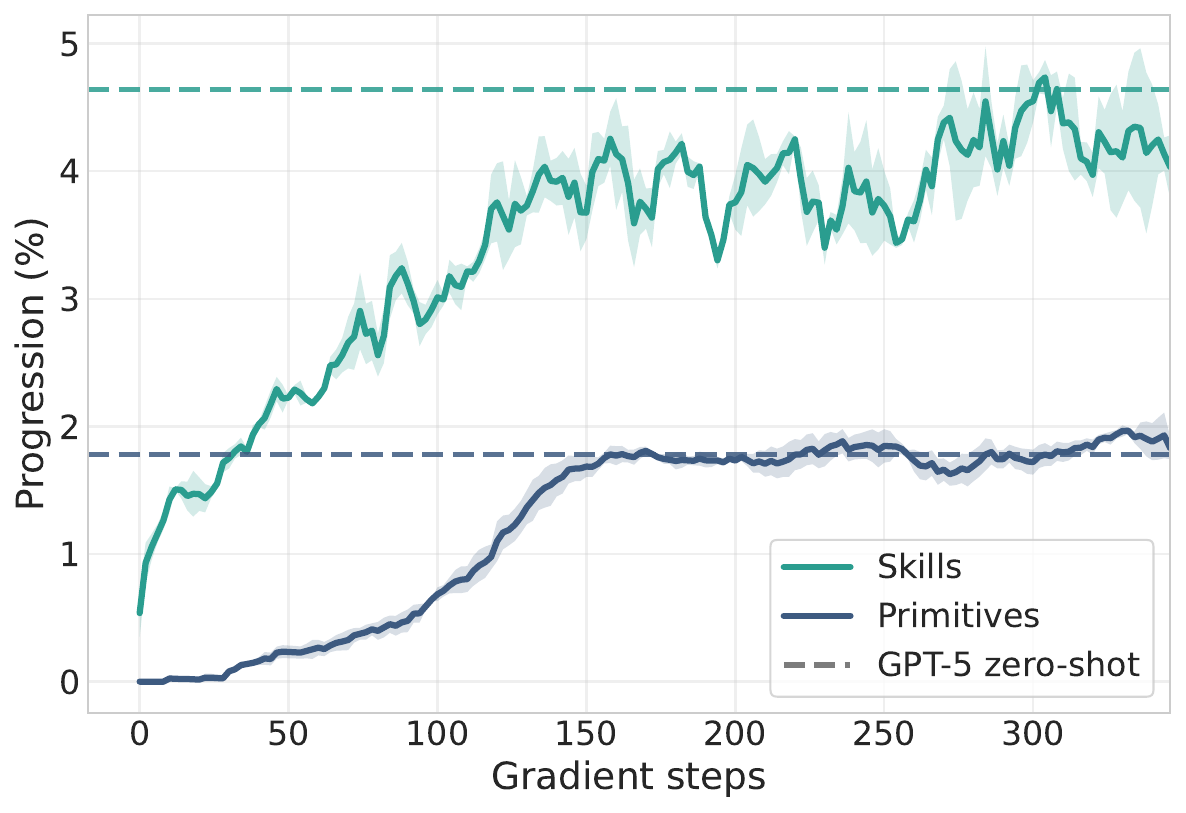}
    \caption{\textbf{The value of skills vs. primitives for language agents in NetHack.} Higher-level, semantically meaningful skills provided by \skillLibraryName improve performance both in zero-shot use and during RL. Learning curves are for \texttt{Qwen-3.5-4B}.}
    \label{fig:teaser}
    \vspace{-0.5em}
\end{wrapfigure}

As language models (LMs) transition into autonomous agents, they are increasingly tasked with complex, long-horizon problems, ranging from software engineering~\citep{yang2024swe} and computer use~\citep{xie2024osworld,yuan2026osworld2} to embodied control~\citep{liang2023code}. Many of these environments require chaining together long sequences of \emph{low-level}, primitive actions, such as precise mouse clicks and keystrokes for computer use agents, or chains of \texttt{cd} and \texttt{ls} commands for software engineering agents. While operating on this primitive action space allows for the finest level of control, it can also be costly, and elevating the action interface to a higher level of temporal abstraction can help facilitate learning and exploration~\citep{SUTTON1999181,nachum2019does}.

Past approaches have addressed these challenges through agent-computer interfaces~\citep{yang2024swe,wangvoyager} and temporal abstractions~\citep{SUTTON1999181}. However, to the best of our knowledge, no study has examined the empirical benefits and tradeoffs of using primitive actions, higher-level skills, or a mixture of both for language agents in long-horizon settings. This paper aims to fill that gap by studying the use of temporally extended code-based actions, or ``skills.'' Skills, closely related to options in hierarchical reinforcement learning~\citep{SUTTON1999181,bacon2017option,klissarov2025discovering}, execute sequences of primitive actions behind a semantically meaningful interface. Rather than learning the skills themselves, we ask how useful a fixed library of such skills is for language-model agents and learned controllers. To make this comparison possible, we build \skillLibraryName, a code-based skill library for NetHack and MiniHack that exposes the same environment through primitive-only, skill-only, and mixed action interfaces. Code-based skills are natural for language agents because they can be described in language and inspected as source code. They also open a path toward agents that construct and refine these skills themselves~\citep{liang2023code,wangvoyager}.


We first compare skills with primitives in MiniHack, where controlled tasks let us isolate navigation, exploration, combat, and item-use. We then use NetHack as our main domain because completing it remains a longstanding challenge, despite extensive work on RL, imitation learning, and language-based control~\citep{piterbarg2023nethack, tuyls2024scaling,wolczyk2024fine,paglieri2025balrog}. Successful games span tens to hundreds of thousands of turns, while a single mistake can end a run~\citep{kuttler2020nethack}. The variety of situations agents can encounter in NetHack also makes pure skill abstraction incomplete in practice as no practical fixed library can anticipate every contingency, so agents may still need to move back down to primitive control when abstractions are poorly matched to the current state. We therefore also evaluate mixed control on NetHack, which allows the agent to move up and down the abstraction ladder.

\paragraph{Contributions.} In this paper, we provide a systematic study of action abstraction for long-horizon language agents. Specifically, our work makes the following contributions:
\begin{enumerate}[leftmargin=*]
    \item \textbf{\skillLibraryName.} We develop and release \skillLibraryName, an open-source library of reusable Python skills for NetHack and MiniHack, together with training and evaluation code for studying primitive-only, skill-only, and mixed control.
    \item \textbf{A systematic study of action abstraction.} We compare agents that use primitives, skills, or both, in zero-shot, SFT, and RL settings, and present careful ablations relevant to skill coverage.
    
    \begin{itemize}[leftmargin=*]
        \item \textbf{Zero-shot control.} Skills improve success rates on every evaluated MiniHack task, with an average gain of 55 percentage points over primitive control. These benefits extend to NetHack, where skills nearly triple average progression across 14 models while reducing inference cost per episode by 86\% relative to primitives. Adding primitives alongside skills preserves most of these performance benefits, although at a higher inference cost than skill-only control.
        
        \item \textbf{Learning with RL.} We study how the action interface affects learning under a fixed RL budget in NetHack. Averaged across two models, skill-only and mixed controllers achieve gains in dungeon depth 7.2 and 8.6 times those of primitive-only controllers, respectively. These larger gains widen the absolute performance gap already present in zero-shot evaluation, showing that the benefits of skills extend to learning.
        
        \item \textbf{Incomplete skill coverage.} We investigate whether retaining access to primitive actions alongside skills helps compensate for incomplete skill coverage, and find that mixed control is generally less affected by missing skills than skill-only control. Selected trajectories illustrate how primitive fallback works in practice, with agents using primitives to resolve situations that the available skills cannot handle before returning to skills.
    \end{itemize}
\end{enumerate}

%% file: sections/intro_neurips.tex
\section{Introduction}

As language models (LMs) transition into autonomous agents, they are increasingly tasked with complex, long-horizon problems, ranging from software engineering~\citep{yang2024swe} and computer use~\citep{xie2024osworld} to embodied control~\citep{liang2023code}. To succeed in these environments, agents must manage extended interactions, adapt to shifting contexts, and recover from inevitable errors, all while carefully managing their inference budgets. Despite this, most agents are still forced to operate through low-level action interfaces, expanding simple behaviors into long sequences of brittle, error-prone decisions.

Agentic coding systems make this tension concrete. If a coding agent had to generate one keystroke at a time, it would risk wasting its reasoning budget on mechanics rather than design. In practice, programmers and coding agents rely on functions, libraries, tools, and editor commands that package many low-level operations into semantically meaningful units~\citep{yang2024swe}. A similar pattern  
\begin{wrapfigure}{r}{0.50\linewidth}
    \vspace{-0.5em}
    \centering
    \includegraphics[width=\linewidth]{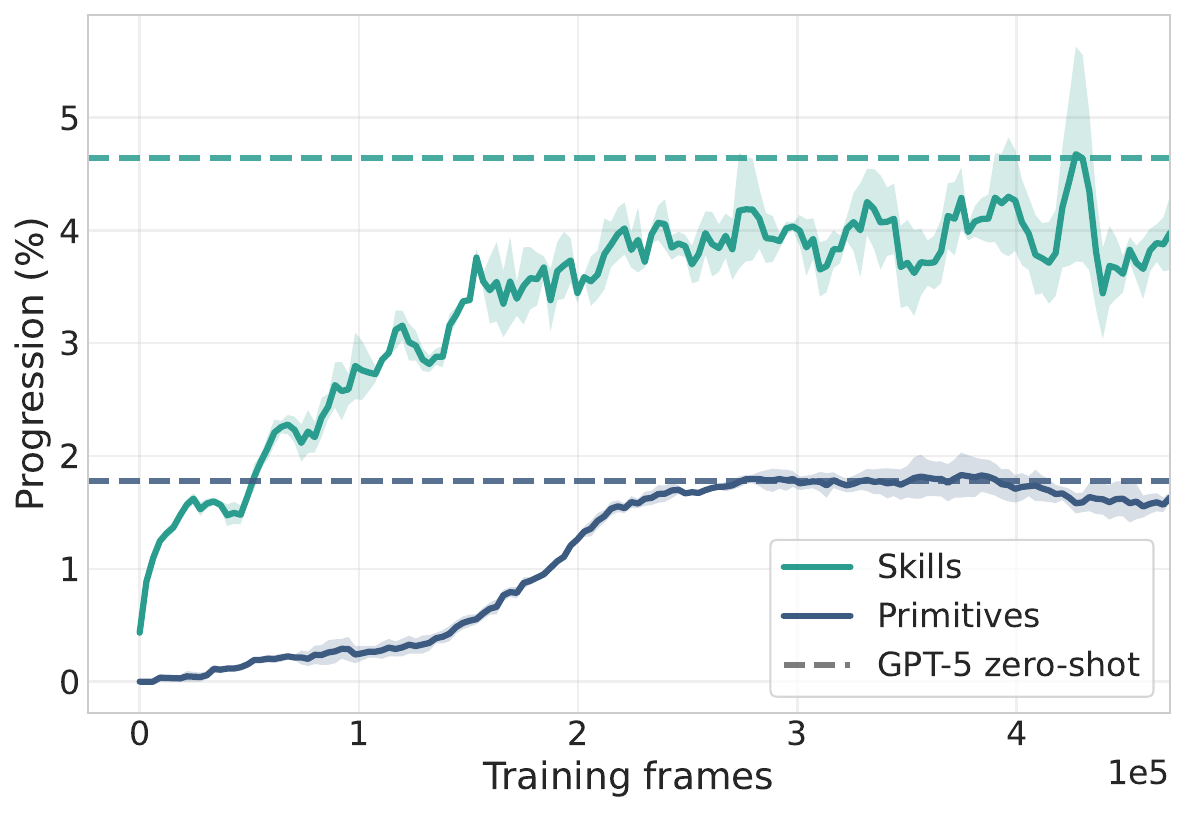}
    \caption{\textbf{The value of skills vs. primitives for language agents in NetHack.} Higher-level, semantically meaningful skills provided by \skillLibraryName improve performance both in zero-shot use and during RL. Learning curves are for \texttt{Qwen-3.5-4B}.}
    \label{fig:teaser}
    \vspace{-1.0em}
\end{wrapfigure}
exists beyond software: a robot may usually navigate by calling a reusable navigation routine, but still needs low-level control when the routine fails in a cluttered or unusual state. This suggests a central question for long-horizon language agents: \emph{what are the benefits and tradeoffs of using primitive actions, higher-level skills, or both?}




In this paper, we study this question through temporally extended code-based actions, or ``skills.'' Skills, closely related to options in hierarchical reinforcement learning~\citep{SUTTON1999181,bacon2017option,klissarov2025discovering}, execute sequences of primitive actions behind a semantically meaningful interface. Rather than learning the skills themselves, we ask how useful a fixed library of such skills is when exposed to language-model agents and learned controllers. To make this comparison possible, we build \skillLibraryName, a code-based skill library for NetHack and MiniHack that exposes the same environment through primitive-only, skill-only, and mixed action interfaces. Code-based skills are natural for language agents: they can be described in language, inspected as source code, executed cheaply on the CPU, and combined with primitive actions when finer control is needed.


We first evaluate these interfaces in MiniHack, where controlled tasks let us isolate navigation, exploration, combat, and item-use. We then use NetHack as our main domain because it captures much of the structure of real long-horizon agentic tasks: agents must act under partial observability, adapt across changing contexts, recover from mistakes, and repeatedly switch among qualitatively different modes of behavior such as exploration, navigation, combat, resource management, and tool use. This complexity also makes pure skill abstraction incomplete in practice: no practical fixed library can anticipate every contingency, so agents may still need to move back down to primitive control when abstractions are poorly matched to the current state. This allows us to ask whether effective long-horizon agents should operate at a single level of abstraction, or move flexibly up and down the abstraction ladder.

Across both MiniHack and NetHack, skills substantially improve performance over primitive-only control. In MiniHack, \texttt{GPT-5} with skills achieves non-zero zero-shot success on every task, including 25\% success on \texttt{MiniHack-Quest-Hard-v0}, where primitive control solves no episodes. In NetHack, skill-only zero-shot agents nearly triple average progression, reach about 2.4x as deep into the dungeon, and reduce inference cost by 85\% relative to primitive-only agents. Mixed control preserves most of these zero-shot gains while restoring access to primitive actions. After RL, skill-based and mixed controllers more than double final progression relative to primitive control; mixed control becomes slightly stronger on average than skill-only control, and our action-distribution analysis shows that the mixed controller continues to use primitives rather than collapsing to skills alone.
We release \skillLibraryName, an open-source, rich library of code-based skills for MiniHack and NetHack, together with training and evaluation code, to support research on skills for language agents in challenging environments.

%% file: sections/preliminaries.tex
\section{Preliminaries}
\label{sec:preliminaries}
While skills and options have a rich history in the classic RL literature (see~\cref{sec:related-work}), they are not as commonly used for language agents. As such, we here give a brief overview of options as the formal framework that underlies our work.

In a fully observable Markov decision process (MDP), an option $\omega$ is usually defined by an initiation set $\mathcal{I}_\omega \subseteq \mathcal{S}$, an intra-option policy $\pi_\omega(a\mid s)$, and a termination function $\beta_\omega(s)$. However, in a partially observable MDP (POMDP) like NetHack, the agent does not observe the latent state $s_t$, so we define these components over the interaction history $h_t$. Thus, each skill $\omega$ has an initiation set $\mathcal{I}_\omega \subseteq \mathcal{H}$, an intra-skill policy $\pi_\omega(a_t\mid h_t)$, and a termination function $\beta_\omega(h_t)$. The initiation set $\mathcal{I}_\omega$ contains all histories that an option $\omega$ can execute from. In our work, $\forall \omega \in \Omega: \mathcal{I}_\omega = \mathcal{H}$ where $\mathcal{H}$ is the set of all histories. The intra-option policy $\pi_\omega(a\mid h)$ is a mapping from a history to an action to take at that history, i.e. $\pi_\omega:\mathcal{H} \to \Delta(\mathcal{A})$. Finally, the termination function $\beta_\omega(h)$ gives the probability with which option $\omega$ stops execution when at history $h$. In our work $\beta_\omega(h)$ will be $1$ whenever the option $\omega$ has finished executing its task (e.g. killed a monster), or when the skill policy runs into an unexpected situation such as when the agent loses health points during execution of the skill. Finally, there is the high-level option policy $\mu: \mathcal{H} \to \Delta(\Omega \cup \mathcal{A})$, which outputs a distribution over all options $\Omega$, all low-level actions $\mathcal{A}$, or both combined. In our work, the high-level option policy $\mu$ will be a language model, which will follow the call-and-return model~\citep{klissarov2025discovering} to switch between high-level decision and low-level option policy control.


\begin{remark}\label{remark}
    Note that our notion of ``skills'' in this work is related to, but different from other popular LLM-centric notions of skills such as agentic markdown files~\citep{han2026swe}, metacognitive descriptions of reasoning procedures~\citep{didolkar2024metacognitive}, or tools~\citep{schick2023toolformer}. Instead, skills are policies that operate in the same environment as the controller itself.
  \end{remark}

%% file: sections/skill_library.tex

\begin{figure}[t]
    \centering
    \includegraphics[width=\linewidth]{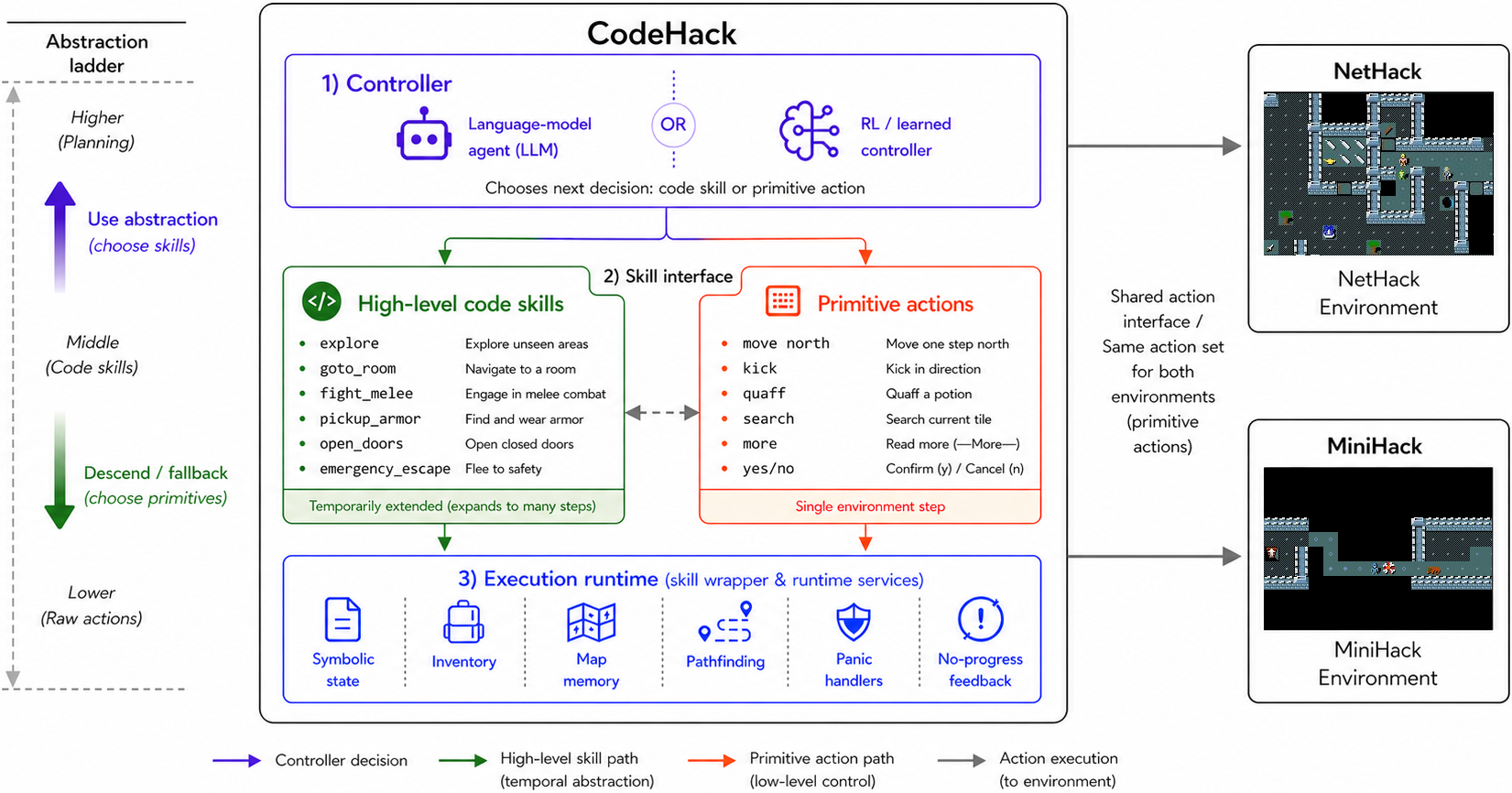}
    \caption{\textbf{\skillLibraryName as a shared control layer for primitive-only, skill-only, and mixed agents.} \textbf{1.} The controller can be either a language-model agent or a learned RL policy, since \skillLibraryName{} supports both natural-language control and a discrete action-space interface. \textbf{2.} Depending on the configured action interface, the controller selects code skills such as \texttt{explore} or \texttt{fight\_melee}, primitive commands such as movement, \texttt{kick}, or \texttt{search}, or both. \textbf{3.} All three interfaces use the same runtime for symbolic state, inventory tracking, map memory, panic handlers, and no-progress feedback. Primitive commands are registered as one-step strategies. Code skills can expand into multiple primitives. \textbf{4.} This runtime supports both NetHack and MiniHack, allowing the same skill library to transfer across environments while preserving a path back to low-level control when abstractions are insufficient.}
    \label{fig:codehack-library}
    \vspace{-1.5em}
\end{figure}

\section{\skillLibraryName: a library of code skills for NetHack}
\label{sec:codehack}

We build \skillLibraryName, a library of code-based skills for NetHack and MiniHack together with a runtime shared by primitive-only, skill-only, and mixed agents. In all three settings, the runtime maintains symbolic state, inventory tracking, and map memory. Primitive commands are registered as one-step strategies and pass through the same execution machinery as code skills. Each code skill is a Python procedure that accesses this structured state to issue primitive NetHack actions, such as movement commands, inventory interactions, or prompt responses, and can adapt to new observations during execution. The language controller receives textual observations, including inventory and summaries of the remembered map, with bounded interaction history, and selects the next action when control returns. \skillLibraryName is not intended to solve NetHack end-to-end through a fixed hand-written policy. Instead, it provides actions that a language model or learned controller can invoke during sequential decision making.~\cref{fig:codehack-library} summarizes this division: the controller selects an action from the configured repertoire, and the runtime executes it. The same runtime is reused across NetHack and MiniHack, enabling direct transfer of the skill library and mixed control over multiple abstraction levels. Additional technical details of the implementation are provided in~\cref{app:codehack}.

We focus on code-based skills because they provide a concrete substrate for studying action abstraction. Each skill has an explicit name, a readable implementation, and a behavioral specification that can be tested directly in NetHack or MiniHack. Compared with latent skills, this makes the action interface easier to inspect, modify, and extend, while giving language models access to semantically grounded actions such as exploration, combat, or item management.

The library implements a broad repertoire of higher-level skills for recurrent NetHack behaviors. These include exploration and navigation, combat and tactical control, inventory management, and recovery handling. Representative examples include \texttt{explore}, \texttt{goto\_room}, \texttt{fight\_melee}, \texttt{pickup\_armor}, \texttt{open\_doors}, \texttt{leave\_shop}, and \texttt{emergency\_escape}. Our NetHack experiments expose a curated repertoire of 78 skills. MiniHack configurations use subsets of this repertoire, augmented with 7 additional skills for lava crossing and levitation. The skill lists and configurations used in the primary zero-shot evaluation are provided in~\cref{app:experimental_skill_set}.

This skill action space is strong but incomplete. Rather than attempting to cover all of NetHack with a fully engineered skill library, we prioritize frequent and reusable behaviors that capture a substantial portion of practical play. This makes the setting a more realistic test of skill-based control: agents benefit from abstraction, but they must still cope with missing skills, edge cases, and local failures.


%% file: sections/experimental_setup2.tex
\begin{figure}[!t]
    \centering
    \includegraphics[width=\linewidth]{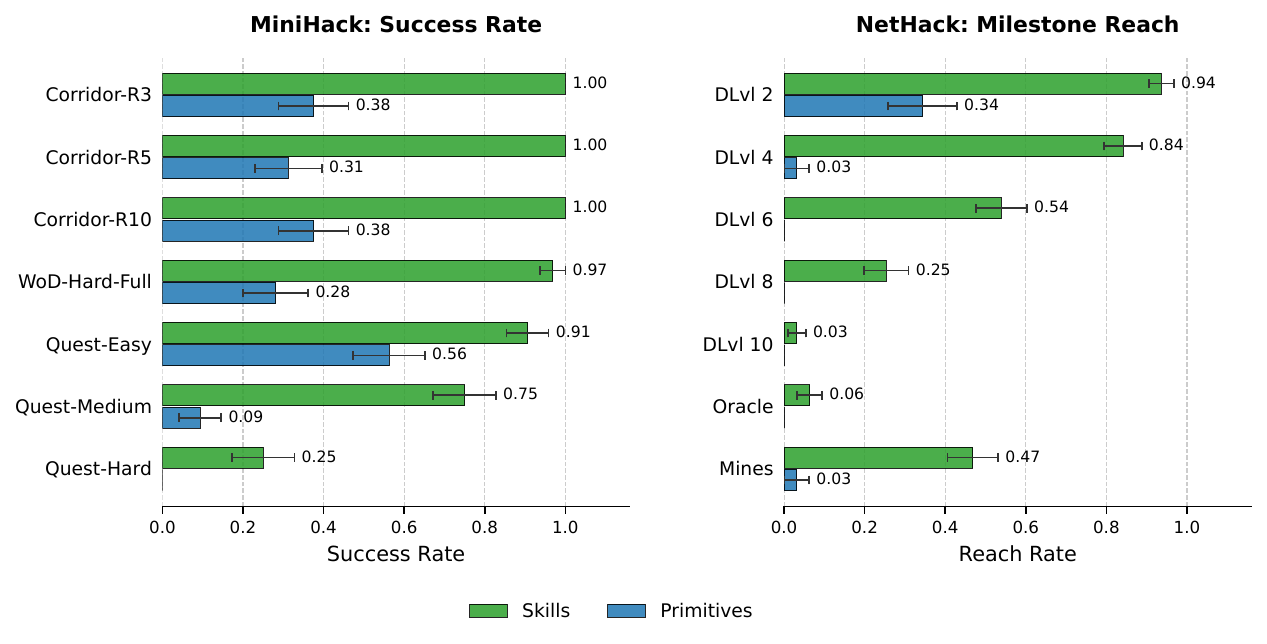}
    \caption{\textbf{Zero-shot results across the two evaluation domains.} \textbf{\emph{(Left)}} MiniHack success rates for GPT-5, used as a controlled testbed for diagnosing failures of primitive control. \textbf{\emph{(Right)}} NetHack milestone reach for GPT-5, showing that skill-based control consistently reaches substantially deeper milestones than primitive-only control. Both plots present averages over 32 episodes per task.}
    \label{fig:zero-shot-overview}
    \vspace{-1.5em}
\end{figure}

\section{Experimental setup}
\label{sec:experimental-setup}

\paragraph{Environments} We use MiniHack~\citep{samvelyan2021minihack} for controlled tests of specific behaviors and the NetHack Learning Environment (NLE)~\citep{kuttler2020nethack} for long-horizon evaluation and learning. Our MiniHack suite comprises seven tasks: \texttt{Corridor-\{R3,R5,R10\}} test navigation and exploration, \texttt{WoD-Hard-Full} tests acquiring and using a wand in combat, and \texttt{Quest-\{Easy,Medium,Hard\}} combine navigation, exploration, item use, and combat in increasingly difficult settings. NetHack requires coordinating these behaviors over thousands of steps while exploring more than 50 levels of a procedurally generated dungeon to retrieve the Amulet of Yendor and return to the surface. All NetHack experiments use \texttt{NetHackScore-v0}. \Cref{app:subsec:environment-details,app:subsec:minihack-task-suite} describe the environments and task configurations.

\paragraph{Environment setup} Both domains are accessed through the \skillLibraryName environment runtime, which wraps the underlying NLE or MiniHack simulator. For each task, we keep the underlying game parameters, observation format, and prompt template fixed across action interfaces, changing the available-action lists and descriptions to match the selected interface. We use the NLE language wrapper~\citep{goodger2023languagewrapper} to convert observations into natural language and an ASCII map. Following the BALROG setup~\citep{paglieri2025balrog}, we then present this information to the LLM through a chat template. At each timestep \(t\), the agent receives a bounded interaction history together with the current observation \(o_t\) inside this chat template, guided by a system prompt that describes the task and the available actions. \Cref{app:subsec:prompt-templates-and-eval-protocol} details the prompt structure, history settings, and handling of malformed outputs. \Cref{app:subsec:environment-details} gives the NetHack episode limits and termination settings.\loose

\paragraph{Action interfaces} We compare three action interfaces through the \skillLibraryName runtime. Primitive-only agents select low-level game commands, skill-only agents select code-based skills, and mixed agents can select either. NetHack uses a repertoire of 78 skills, while MiniHack uses task-specific subsets of this repertoire, augmented with additional skills described in~\cref{app:experimental_skill_set}. The skill implementations remain fixed throughout the experiments.

\paragraph{Evaluation metrics} To evaluate game performance, we use task success rate in MiniHack and score, progression, and dungeon level in NetHack. The NetHack score is built into the game and takes into account a variety of factors such as dungeon depth, enemy kills, and collected gold. 
Since game score is not always aligned with winning, we also report the progression metric introduced by~\citet{paglieri2025balrog}, which estimates progress from the highest dungeon and experience levels reached using a mapping derived from human trajectories (\cref{app:subsec:environment-details}).
Finally, dungeon level tracks the maximum dungeon level the agent reaches anywhere in the episode. We also report token usage and estimated inference cost per episode, using OpenRouter input and output token prices.

\paragraph{Zero-shot evaluation} We compare all three action interfaces on NetHack across 14 models from the GPT~\citep{singh2025openai,agarwal2025gpt}, Llama~\citep{grattafiori2024llama}, Gemma3~\citep{GemmaTeam2025}, Gemma-4, and Qwen-3.5~\citep{team2026qwen3} families. GPT-5, GPT-OSS-120B, Gemma-4, and Qwen-3.5 use reasoning mode, while Llama and Gemma3 use act-only control. On MiniHack, we compare primitive-only and skill-only control using GPT-5.

To test whether primitive actions help compensate for missing skills, we remove one skill family at a time and evaluate Gemma-4-31B on NetHack. We compare skill-only and mixed control using the same reduced library, with primitive actions available under mixed control (\cref{app:skill-coverage}).

\paragraph{Learning experiments} 
We train \texttt{Llama-3.1-8B-Instruct} and \texttt{Qwen-3.5-4B} on NetHack with PPO~\citep{Schulman2017ProximalPO} under all three action interfaces.
Both models use act-only control in this comparison, including their zero-shot baselines. The reward is the change in game score, aggregated over the primitive transitions executed by each skill. PPO uses 32 rollouts per trainer GPU, each containing 16 controller decisions. We compare policies at a common checkpoint after 264 gradient updates and report learning curves against gradient updates and wall-clock time (\cref{app:rl-results}). Skills can execute multiple primitive actions per decision, so equal rollout lengths can involve different numbers of primitive environment transitions.

We also test whether supervised fine-tuning (SFT) improves initialization for skill-only RL. We collect 1{,}024 zero-shot teacher trajectories with \texttt{Gemma-4-31B} and fine-tune both models for one epoch under skill-only control. We evaluate the SFT policies directly and use them to initialize additional PPO runs. We compare PPO with and without SFT initialization after the same number of gradient updates. SFT adds dataset collection and supervised training to the total training cost. \Cref{app:subsec:training-details} gives the training hyperparameters.

%% file: sections/results.tex
\section{Language agents with skills: Experimental results}
\label{sec:experimental-results}

    

\noindent
\begin{minipage}[t]{0.48\textwidth}
    \vspace{0pt} 
    We organize the results around five empirical questions, which we address through six research findings (RF).
    \begin{enumerate}[leftmargin=*]
        \item How do skills affect zero-shot performance across tasks and models? \emph{(\hyperref[sec:rf1]{RF1}, \hyperref[sec:rf2]{RF2})}
        \item How do skills affect inference cost? \emph{(\hyperref[sec:rf3]{RF3})}
        \item What are the benefits and tradeoffs of allowing primitive actions alongside skills? \emph{(\hyperref[sec:rf4]{RF4})}
        \item How do skills affect gains over a fixed RL budget? \emph{(\hyperref[sec:rf5]{RF5})}
        \item Does SFT initialization improve the final RL policy? \emph{(\hyperref[sec:rf6]{RF6})}
    
    \end{enumerate}
\end{minipage}%
\hfill
\input{tables/nethack_zero_shot_averages_v5}\paragraph{RF1: Skills improve success rate across all evaluated MiniHack tasks.} \label{sec:rf1} In \cref{fig:zero-shot-overview} (left), we compare \texttt{GPT-5}'s zero-shot success rates using \skillLibraryName skills versus primitive-only control across seven MiniHack tasks. Skills yield large improvements across all evaluated tasks, averaging 55 percentage points over primitive control. Skills achieve near-perfect success on all three \texttt{Corridor} tasks, which isolate navigation and exploration. On \texttt{WoD-Hard-Full}, success rises from 28\% to 97\%, extending the gains to acquiring and using a wand in combat. Skills also improve \texttt{Quest-Easy} and \texttt{Quest-Medium}, which require combining navigation, item use, and combat, although success remains below that on the \texttt{Corridor} and \texttt{WoD} tasks. Skills enable 25\% success even on \texttt{Quest-Hard}, the hardest task in our suite, where primitive control solves no episodes. Together, these results show that access to higher-level skills makes \texttt{GPT-5} substantially more reliable at tasks requiring navigation, item use, combat, and combinations of these behaviors, although \texttt{Quest-Hard} remains far from consistently solved.

\paragraph{RF2: Skills improve NetHack performance across model families.} \label{sec:rf2} In the 14-model zero-shot sweep, skill-only agents achieve 2.9x the progression, 3.8x the score, and 2.4x the dungeon depth of primitive-only agents on average (\cref{tab:nethack-zero-shot-averages}). Measured by dungeon depth, the relative advantage of skills generally increases with model size. This trend is clearest for Llama, where the skill-to-primitive depth ratio increases from 1.7x at 3B to 3.0x at 70B (\cref{tab:nethack-zero-shot-summary}).

To make these averages more concrete,~\cref{fig:zero-shot-overview} shows milestone reach for \texttt{GPT-5}, the strongest model in our zero-shot sweep. The fraction of episodes reaching successive dungeon levels and landmarks such as the Oracle and Gnomish Mines shows how often the agent advances beyond the opening levels and how far it reaches into the dungeon~\citep{klissarovmaestromotif}. With skills, \texttt{GPT-5} reaches dungeon level 4 in 84\% of episodes, compared with 3\% under primitive control, and the Gnomish Mines in 47\% versus 3\%. These are substantial early-game gains, but even \texttt{GPT-5} with skills remains far from solving NetHack (see \hyperref[sec:discussion]{Discussion}). Per-model results and milestone reach rates averaged across models are provided in~\cref{tab:nethack-zero-shot-summary,tab:nethack-skills-vs-primitives}.


\begin{figure}[t]
    \centering
    \includegraphics[width=\linewidth]{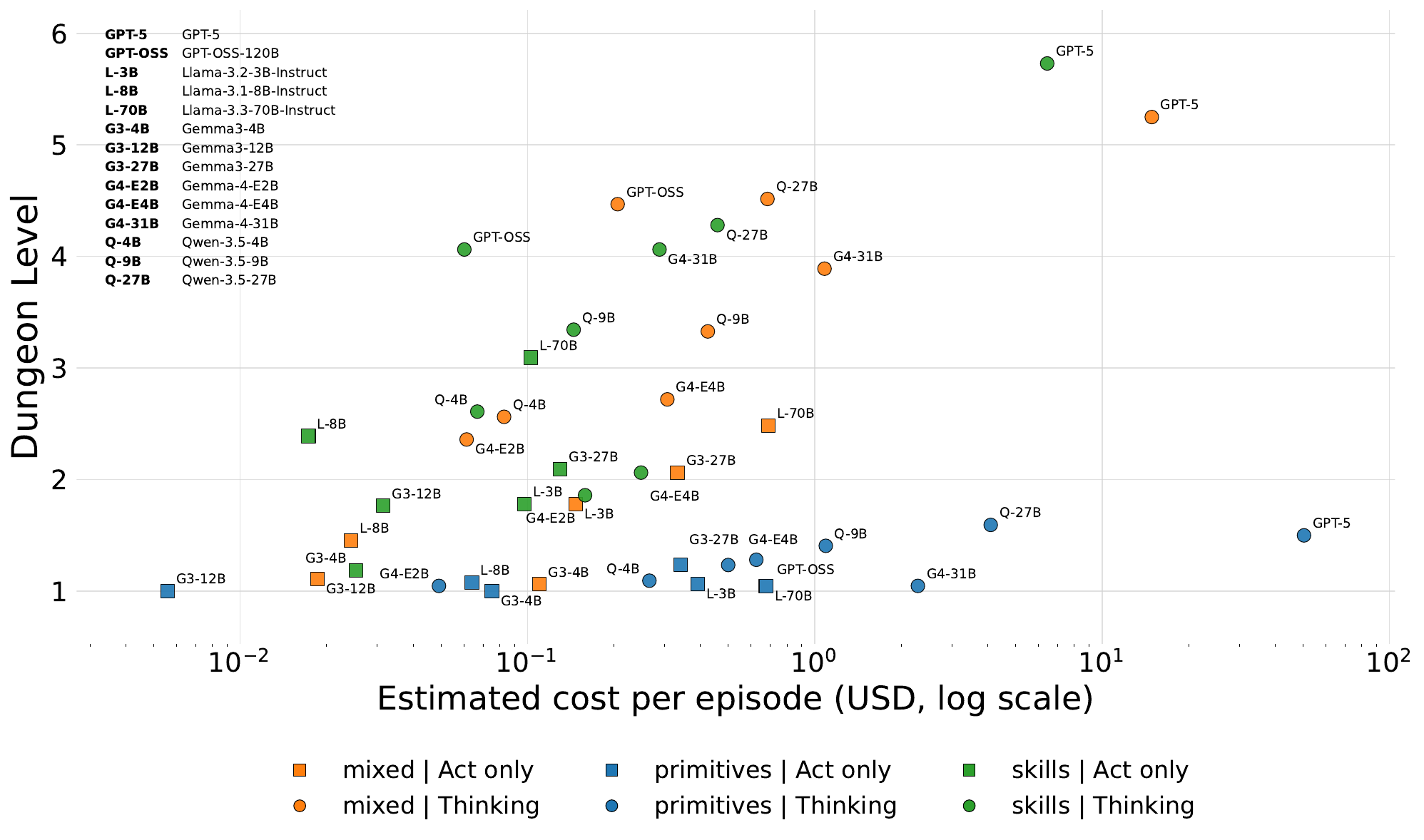}
    \caption{\textbf{Zero-shot NetHack cost-performance frontier.} 
    Each point represents a model evaluated with a particular action interface, showing its average dungeon level reached against its average inference cost per episode. For most models, switching from primitives to skills increases dungeon depth while reducing inference cost. \be{perhaps plot a dashed line for the pareto frontier}}
    \label{fig:nethack-zero-shot-results}
    \vspace{-1.5em}
\end{figure}

\paragraph{RF3: Skills improve the performance-cost frontier.} \label{sec:rf3} \cref{fig:nethack-zero-shot-results} plots dungeon depth against estimated inference cost per episode, using OpenRouter input and output token prices\footnote{\url{https://openrouter.ai}}. We observe two patterns. First, with skill-only or mixed control, models with higher per-episode costs generally reach deeper into the dungeon, whereas primitive-only performance remains largely flat across costs. Second, for most models, switching from primitives to skills moves the corresponding point up and to the left: the same model reaches deeper dungeon levels at a lower inference cost per episode.

\cref{tab:nethack-zero-shot-averages} quantifies these savings: averaged across models, skills reduce inference cost per episode by 86\% and token usage by 74\% relative to primitives (see~\cref{tab:nethack-zero-shot-summary} for the per-model cost breakdown). Skills lower inference cost by reducing the number of LM calls per episode by an average factor of 5.1. This reduction happens because many environment steps are taken by a code skill policy, which can run quickly and cheaply on the CPU, instead of the more expensive language model policy.

\paragraph{RF4: Mixed control mostly preserves the benefits of abstraction while reducing dependence on library coverage.} \label{sec:rf4} Mixed control retains 95\% of skill-only progression, 91\% of score, and 97\% of dungeon depth, remaining much stronger than primitives alone (\cref{tab:nethack-zero-shot-averages}). It occasionally exceeds skill-only performance, but raises average inference cost to 2.3x and token usage to 2x that of skills. Access to primitives lets the controller attempt behaviors that skills cannot provide, while requiring it to decide both what to do and which abstraction level to use.

To examine whether primitives help compensate for incomplete skill coverage, we remove one skill family at a time and compare zero-shot \texttt{Gemma-4-31B} under skill-only and mixed control (\cref{app:skill-coverage}). Removing skill families can reduce performance in both settings, but mixed control is generally less affected. In the skill-only setting, losing essential skills can severely restrict further progress, whereas mixed control can fall back on primitives to continue advancing through the dungeon.

Two selected zero-shot \texttt{GPT-5} trajectories illustrate how primitive fallback works in practice (\cref{app:qualitative-analysis}). In a vault, the library lacks a skill for dropping gold to satisfy the guard's demand. After several unproductive calls, the agent uses primitives to drop the gold and move along the exit corridor, then resumes navigation with \texttt{explore}. In the ranged-combat example, \texttt{fight\_ranged} returns without acting; the agent uses primitives to throw a javelin before using \texttt{pickup\_weapon} to recover its weapons. In both cases, primitives resolve a local problem that the available skills cannot handle, allowing the agent to return to skill-based control.

\input{tables/rl_results_v5}

\paragraph{RF5: RL amplifies the performance gap between skill-based control and primitive-only control.} \label{sec:rf5} We report the evaluation results for the final checkpoint performance of RL in~\cref{tab:rl-results}. We find that both skill-only and mixed control exhibit faster learning than primitive-only control, as demonstrated by a larger performance gain using the same training budget. For example, the gain in dungeon level averaged across models for skill-only and mixed control is 7.2x and 8.6x larger than that for primitive-only control, respectively. We also provide learning curves in ~\cref{fig:rl-progression-score-max-dlvl-vs-grad-steps-llama,fig:rl-progression-score-max-dlvl-vs-wall-clock-time-llama,fig:rl-progression-score-max-dlvl-vs-grad-steps-qwen,fig:rl-progression-score-max-dlvl-vs-wall-clock-time-qwen} of~\cref{app:rl-results}, which provide further evidence that skill and mixed agents improve faster than primitive agents throughout training, both as a function of gradient steps and wall-clock time. \bc{Beyond the training environment, our OOD evaluation on MiniHack reveal that RL improves success rates for both interfaces, while the skill-based controller remains substantially stronger overall and improves not only the Corridor tasks but also the harder Quest tasks (see~\cref{app:ood-minihack-transfer}).} The immediate consequence of faster learning for skills and mixed is that the performance gap between these interfaces and the primitive-only interface is larger at the end of RL training than at the start, validating our finding that RL \emph{amplifies} the performance gap.

\paragraph{RF6: SFT initialization improves final skill-only RL performance.}\label{sec:rf6} To test whether teacher demonstrations can further improve skill-based NetHack control, we combine SFT with RL, following prior work~\citep{wolczyk2024fine,alphago,guo2025deepseek}. At the same RL checkpoint, SFT initialization yields higher mean performance than initialization from the original models (\cref{tab:rl-results}). Mean progression increases from 3.45\% to 4.46\% for \texttt{Llama-3.1-8B-Instruct} and from 4.36\% to 5.25\% for \texttt{Qwen-3.5-4B}, relative improvements of 29\% and 20\%, respectively. Mean scores increase by approximately 51\% for both models. Although the SFT-initialized policies gain less during RL, their stronger starting points translate into better final policies within the evaluated budget. Teacher imitation therefore provides a useful initialization for learning skill-based control, with the additional cost of collecting demonstrations and supervised training.

\davide{Is it possible to run ablations on the skills? Which of these skills are most useful? I'd imagine navigation skills. Would be interesting to check this out, that way we see where are the actual deficits of low-level actions. In my opinion those deficits are likely in the spacial understanding of the map and the navigation skills among complex corridors (you can kind of see this from the minihack experiments as well where mainly low-level LLMs struggle due to poor navigation)}

\be{Broadly, results seem strong! The actual writing of this section will make it a bit challenging for the reader to understand the key takeaways. As the writer, you job is not only to convey the data (figures/tables are fine for that) but also to interpret the data and contextualize it w.r.t. broader scientific questions and prior work. As one metric for doing this: read this section aloud but replace any numbers/long-word with "blah" and see if it still makes sense. \\

One other writing suggestion, which should help strengthen the presentation, is to check that each paragraph starts with an intro sentence that summarizes/introduces the rest of the paragraph. If you just read the first sentence of each paragraph, does it make sense? [This is also a good sanity check for the rest of the paper]}










%% file: tables/nethack_zero_shot_averages_v5.tex
\begin{minipage}[t]{0.49\textwidth}
    \vspace{0pt} 
    \centering
    \setlength{\tabcolsep}{2.1pt}
    \begin{tabular}{lrrr}
        \toprule
        Metric & Primitives & Mixed & Skills \\
        \midrule
        Progression & 0.69 {\tiny\textcolor{gray}{$\pm$ 0.16}} & 1.88 {\tiny\textcolor{gray}{$\pm$ 0.33}} & 1.98 {\tiny\textcolor{gray}{$\pm$ 0.33}} \\
        Score & 83.9 {\tiny\textcolor{gray}{$\pm$ 20.2}} & 291.0 {\tiny\textcolor{gray}{$\pm$ 71.0}} & 318.3 {\tiny\textcolor{gray}{$\pm$ 89.2}} \\
        Dlvl & 1.19 {\tiny\textcolor{gray}{$\pm$ 0.05}} & 2.79 {\tiny\textcolor{gray}{$\pm$ 0.35}} & 2.88 {\tiny\textcolor{gray}{$\pm$ 0.34}} \\
        Cost/ep. (\$) & 4.35 {\tiny\textcolor{gray}{$\pm$ 3.56}} & 1.36 {\tiny\textcolor{gray}{$\pm$ 1.04}} & 0.59 {\tiny\textcolor{gray}{$\pm$ 0.45}} \\
        Tok./ep. (M) & 6.72 {\tiny\textcolor{gray}{$\pm$ 1.13}} & 3.54 {\tiny\textcolor{gray}{$\pm$ 0.53}} & 1.74 {\tiny\textcolor{gray}{$\pm$ 0.31}} \\
        \bottomrule
    \end{tabular}
    \captionof{table}{\textbf{Summary of the 14-model zero-shot NetHack sweep.} Values are unweighted averages across models with gray standard errors across model-level means. Progression is in percent; lower is better for cost and token usage.~\cref{app:zero-shot-results} gives the full per-model breakdown.}
    \label{tab:nethack-zero-shot-averages}
\end{minipage}

%% file: tables/rl_results_v5.tex
\begin{table}[t]
    \small
    \centering
    \caption{\textbf{Results across learning regimes for \texttt{Llama-3.1-8B-Instruct} and \texttt{Qwen-3.5-4B}.} Both models use act-only control. We report final score, progression, and dungeon level reached. Evaluation checkpoints are picked at 264 gradient steps for all RL results. Values are mean ± 1 standard error (SE). For zero-shot and SFT, SE is computed across 64 evaluation episodes, while for RL (Base) and RL (SFT) it is computed across 3 training seeds (each seed averages over its own 64 evaluation episodes).}
    \label{tab:rl-results}
    \begin{tabular}{llllll}
        \toprule
        \textbf{Regime} & \textbf{Interface} & \textbf{Model} & \textbf{Score} & \textbf{Progression} & \textbf{Dungeon Level} \\
        \midrule
        \multirow{6}{*}{Zero-shot}
            & Primitives & \texttt{Llama-3.1-8B-Instruct} & 22.0 \stderr{2.9} & 0.12 \stderr{0.05} & 1.08 \stderr{0.03} \\
            & Primitives & \texttt{Qwen-3.5-4B} & 36.3 \stderr{6.5} & 0.29 \stderr{0.08} & 1.17 \stderr{0.06} \\
            & Mixed & \texttt{Llama-3.1-8B-Instruct} & 78.6 \stderr{8.0} & 0.90 \stderr{0.11} & 1.48 \stderr{0.08} \\
            & Mixed & \texttt{Qwen-3.5-4B} & 92.0 \stderr{11.1} & 0.94 \stderr{0.11} & 1.81 \stderr{0.12} \\
            & Skills & \texttt{Llama-3.1-8B-Instruct} & 178.4 \stderr{15.5} & 1.59 \stderr{0.10} & 2.48 \stderr{0.15} \\
            & Skills & \texttt{Qwen-3.5-4B} & 215.9 \stderr{22.1} & 1.49 \stderr{0.13} & 2.45 \stderr{0.18} \\
        \midrule
        \multirow{6}{*}{RL (Base)}
            & Primitives & \texttt{Llama-3.1-8B-Instruct} & 203.1 \stderr{40.7} & 1.54 \stderr{0.23} & 1.46 \stderr{0.14} \\
            & Primitives & \texttt{Qwen-3.5-4B} & 334.9 \stderr{42.6} & 1.94 \stderr{0.01} & 1.54 \stderr{0.03} \\
            & Mixed & \texttt{Llama-3.1-8B-Instruct} & 477.0 \stderr{44.5} & 3.37 \stderr{0.02} & 4.80 \stderr{0.12} \\
            & Mixed & \texttt{Qwen-3.5-4B} & 612.6 \stderr{32.1} & 3.64 \stderr{0.15} & 4.97 \stderr{0.14} \\
            & Skills & \texttt{Llama-3.1-8B-Instruct} & 549.5 \stderr{40.7} & 3.45 \stderr{0.38} & 4.85 \stderr{0.26} \\
            & Skills & \texttt{Qwen-3.5-4B} & 674.5 \stderr{49.9} & 4.36 \stderr{0.24} & 5.51 \stderr{0.20} \\
        \midrule
        \multirow{2}{*}{SFT}
            & Skills & \texttt{Llama-3.1-8B-Instruct} & 542.4 \stderr{55.2} & 3.49 \stderr{0.31} & 4.58 \stderr{0.31} \\
            & Skills & \texttt{Qwen-3.5-4B} & 625.8 \stderr{54.7} & 3.41 \stderr{0.25} & 4.70 \stderr{0.27} \\
        \midrule
        \multirow{2}{*}{RL (SFT)}
            & Skills & \texttt{Llama-3.1-8B-Instruct} & 831.0 \stderr{67.1} & 4.46 \stderr{0.09} & 5.52 \stderr{0.18} \\
            & Skills & \texttt{Qwen-3.5-4B} & 1015.7 \stderr{93.1} & 5.25 \stderr{0.49} & 5.86 \stderr{0.39} \\
        \bottomrule
    \end{tabular}
    \vspace{-1.5em}
\end{table}

%% file: sections/related_work.tex
\section{Related work}
\label{sec:related-work}




\be{I'd recommend moving this to before the results section. Especially since hierarchy/abstraction has been so extensively studied in other parts of the literature, it'd be useful to signal to readers that you've done your homework and read the literature.}

\paragraph{Language models for low-level control.}
Prior work has studied entry points for language models into standard reinforcement learning and control frameworks, including using them as reward functions~\citep{Ma2023EurekaHR,Klissarov2023MotifIM}, for writing code for low-level policies~\citep{kwon2023language,huang2023voxposer}, and (most relevant to our discussion) as a high-level policy that sequences skills~\citep{ahn2022can,wangvoyager,shentu2024llms}.
In robotics,~\citet{liang2023code} use a language model to recursively generate code defining task-specific low-level policies that achieve specified goals. Our work similarly uses code to define policies, but differs by having the LLM call upon an existing set of code policies and sequence them to perform long-horizon tasks.
Our skill-based controller is reminiscent of~\citep{ahn2022can}, which likewise uses
the language model directly as the controller over skills; the aim of our work is different in two ways: (1) to study the value of those skills as compared to primitives and (2) to understand the impact of finetuning the LLM through SFT and RL.
In games,~\citet{klissarovmaestromotif} focus on learning parametric skills using AI feedback and composing them through code defining the high-level policy, whereas our work instead uses code as low-level skills and leverages a possibly fine-tuned LLM as a high-level policy.~\citet{wangvoyager} also use the language model to call code functions that act as skills, but do not study the value of those skills as compared to primitives and instead focus on updating the set of skills for continual progress in an open-ended environment.
Several prior works use NetHack as a rich testbed for evaluating language agents~\citep{paglieri2025balrog,jeurissen2024playing,piterbarg2024diff}. Amongst these, our work is most similar to~\citet{jeurissen2024playing}, which likewise uses code-based skills to play NetHack with LLMs. However, they do not provide a systematic comparison between primitive and skill-based play. The \skillLibraryName runtime developed in this paper is designed to run with both code skills and low-level primitives, allowing us to carefully ablate the choice of action interface while keeping everything else fixed.


\todo{To be included:
\begin{enumerate}
    \item Language models as zero-shot trajectory generators~\citep{kwon2024language}.
    \item Executable code actions elicit better LLM agents~\citep{wang2024executable}. 
    \item HERAKLES: Hierarchical skill compilation for open-ended LLM agents~\citep{carta2025herakles}.
    \item More robotics works? e.g. Inner Monologue: Embodied Reasoning through Planning with Language Models~\citep{huang2023inner}
    \item Ghost in the Minecraft: https://arxiv.org/pdf/2305.17144
\end{enumerate}}


\paragraph{Hierarchical RL.} Foundational work on temporal abstractions in reinforcement learning goes back to feudal RL~\citep{NIPS1992_d14220ee}, the hierarchy of abstract machines~\citep{parr1997reinforcement}, and options~\citep{SUTTON1999181}. The latter has spurred rich follow-up work, including studies of option discovery and learning~\citep{mcgovern2001automatic,stolle_learning,konidaris2009skill}, as well as methods that aim to simultaneously learn the high-level controller along with the options for low-level control~\citep{bacon2017option,nachum2018data,khetarpal2020options,Colas2020LanguageAA,wulfmeier2021data}. A related line of work aims to learn low-level controllers without access to a reward function (i.e., in an unsupervised fashion)~\citep{machado2017laplacian,eysenbachdiversity,hansenfast,sharmadynamics,parkmetra}.
While our work is conceptually motivated by the arguments for hierarchy put forth in this prior work, it is structurally most similar to ~\citet{parr1997reinforcement} in that we assume skills are given as input as a form of prior knowledge. By making this assumption, we can precisely quantify how much a strong repertoire of skills can accelerate learning, abstracting away (the important and still open) problem of how such skills might be learned.  As ~\citet{SUTTON1999181} introduce the options framework, they also show that learning and planning with predefined options can drastically improve sample efficiency. We take inspiration from their findings and add to the understanding of the trade-offs of learning with an existing options set in the context of open-ended environments by leveraging large language models.
In addition, there have also been several related works that, similar to our work, study the benefits of hierarchy. \citet{park2023hiql} highlight that hierarchy helps improve the signal-to-noise ratio, which helps with planning. \citet{nachum2019does} find that most of the HRL benefits  come from improved exploration. \citet{frans2018meta} define a skill discovery objective that directly optimizes for transferability across tasks.
Perhaps most relatedly,~\citet{piterbarg2023nethack} finds hierarchy to help in NetHack, but unlike our work focuses mostly on the imitation learning setting and does not use language models. For an extensive survey of HRL, we refer the reader to~\citet{klissarov2025discovering}.

\todo{Mention horizon reduction work somewhere?~\citep{parkhorizon}}







%% file: sections/conclusion.tex
\section{Discussion}
\label{sec:discussion}
By providing a systematic comparison across skill, mixed, and primitive control, our work suggests that the action interface can strongly influence the performance of language agents in long-horizon environments requiring low-level control. At the same time, the current results that use \skillLibraryName{} do not come close to solving NetHack. The skill library is hand-designed and necessarily incomplete, so its benefits depend on whether the provided abstractions match the situations encountered by the agent. Some failure modes include situations requiring fine-grained tactical control, rare game knowledge, long-term resource management, or strategic planning beyond what the current skills encode. To deal with these failures, future language agents will need to be able to proficiently move up and down the abstraction ladder by knowing when to use higher-level skills vs. when it's more appropriate to use the low-level primitives (see~\cref{app:qualitative-analysis} for examples). We believe our finding that mixed control can already retain most of the benefits of skill-only control is a promising result along this path.


A promising direction for future work is for LLMs to construct both action skills and tools for querying remembered information. Agents could identify repeated failures, implement missing action skills, and test and refine them through interaction~\citep{liang2023code,wangvoyager}. Alongside these skills, they could construct tools that track previously observed entities and retrieve relevant information on demand, reducing the need to repeatedly process a growing observation history. We see this combination of skill construction, selective information access, and primitive fallback as a plausible path toward solving NetHack, allowing agents to both move up and down the abstraction ladder and to extend this ladder through experience.

%% file: sections/appendixA.tex
\section{Technical Details of \skillLibraryName}
\label{app:codehack}

This appendix provides additional implementation detail for \skillLibraryName, the code-based skill library used throughout the paper. Whereas~\cref{sec:codehack} introduces the library at a high level, the goal of this appendix is to make the artifact easier to understand, reproduce, and extend.

\subsection{Library overview}

\skillLibraryName is implemented as a control layer between the primitive NetHack action interface and higher-level controllers such as language models or reinforcement learning agents. Its role is to expose temporally extended, semantically named actions while preserving a path back to lower-level control when finer intervention is needed.

We developed \skillLibraryName iteratively, drawing on human NetHack knowledge to identify reusable behaviors for navigation, exploration, combat, item management, and recovery. We implemented these behaviors as Python procedures and tested them through end-to-end tests, human playthroughs, and \texttt{GPT-5} playthroughs. Failures observed during testing guided debugging and refinement of the skill implementations. For the experiments, we prioritized reliable routines that were broadly useful across the evaluated tasks. 


\paragraph{Design principles.} \skillLibraryName is designed around three practical design principles for long-horizon control: \textbf{(1) Semantic meaning}: skills should correspond to recognizable behaviors such as exploration, combat, equipment management, or terrain interaction rather than arbitrary low-level command bundles. \textbf{(2) Reusability}: skills should be reusable across many states and episodes, so that they function as stable units of abstraction rather than narrow scripts tied to a single layout. \textbf{(3) Mobility across abstraction levels}: the interface should support moving up and down the abstraction ladder, so that high-level skills accelerate long-horizon behavior without removing the possibility of more local intervention when abstractions are insufficient. The runtime therefore includes explicit mechanisms for moving back down the abstraction ladder when a high-level skill is not appropriate. Skills can be interrupted by panic handlers when the local state changes unexpectedly, for example when a newly reachable hostile monster appears or the agent loses HP, and the wrapper can report when a selected skill makes no progress. These signals expose abstraction failures early and let the outer controller choose a different high-level behavior or, under mixed control, revert to more local control. The following subsections describe the runtime, state tracking, and execution mechanics.

\subsection{Wrapper and control interface}

At runtime, \skillLibraryName is introduced by wrapping an NLE or MiniHack environment with a skill-aware wrapper. This wrapper instantiates the internal \skillLibraryName runtime, registers a configurable set of strategies, registers panic handlers, and can optionally expose selected primitive commands as one-step strategies. Once wrapped, the outer controller interacts with the environment by choosing from an action space of named skills rather than directly emitting raw NetHack commands.

Skill execution follows a call-and-return discipline. A controller selects one skill, the corresponding Python procedure is invoked, and the procedure continues issuing primitive actions until it returns, triggers a panic, or the episode terminates. After each primitive step, the runtime updates its state so that the skill can respond to new observations before choosing subsequent actions. For example, selecting \texttt{goto\_room} calls a function that may execute several movement commands based on persistent map memory. Control returns only after the target room is reached, the skill decides that it cannot proceed, an interrupt fires, or the episode ends. The wrapper records how many primitive environment transitions were consumed during the selected skill and exposes this count as \texttt{env\_steps}; it also aggregates the discounted reward accumulated during those internal steps and stores auxiliary skill-level statistics in the returned info dictionary. An additional feedback wrapper can attach a short text message when a selected skill makes zero primitive steps, giving the outer controller an immediate indication that the chosen abstraction was inapplicable or failed locally.

Because the runtime can register both higher-level strategies and selected primitive commands, the same underlying environment can be exposed as a primitive-only, skill-only, or mixed control interface. This is the main experimental advantage of the wrapper: it lets us vary abstraction level without rewriting the environment or the execution machinery.

\subsection{Internal state representation}

The central runtime abstraction in \skillLibraryName is a state manager that maintains a structured, persistent view of the game state on top of raw NLE observations. At every primitive step, this runtime updates its cache, so skills can reason over symbolic entities such as rooms, items, monsters, and reachability instead of repeatedly parsing raw tensors from scratch. The internal state includes the following components:

\begin{itemize}
\item \textbf{Current observation state}, including the current message, glyph map, terminal characters and colors, cursor position, and \texttt{blstats}.
\item \textbf{Persistent level memory}, stored in a \texttt{Level} object that records seen tiles, walkable structure, map objects, visited positions, known traps, door locations, and simple terrain-derived features.
\item \textbf{Inventory state}, maintained by an \texttt{InventoryManager} that parses inventory glyphs and strings into typed items grouped by categories such as armor, rings, potions, food, tools, and wands.
\item \textbf{Character state}, including role-specific skill information, currently known spells, and equipment-dependent combat properties.
\item \textbf{Navigation state}, maintained by a pathfinder over currently known walkable positions, with costs adjusted to avoid local hazards such as monsters or traps.
\item \textbf{Combat and trap utilities}, used by higher-level skills that reason about tactical movement, attack choices, or environmental hazards.
\end{itemize}

Persistent level memory is especially important for navigation and exploration. A skill can depend not only on what is visible now, but also on previously observed map structure such as earlier rooms, corridors, stairs, doors, and hazards.

\subsection{Panic handlers and interruption}
\label{app:panic_handlers}

Skill execution is intentionally interruptible. After each primitive step, \skillLibraryName runs a configurable list of panic handlers over the updated symbolic state. The current implementation includes handlers such as \texttt{lost\_hp}, which aborts when the agent takes damage, and \texttt{enemy\_appeared}, which aborts when a newly reachable hostile monster enters the local state.

This mechanism reduces the brittleness of longer procedures by preventing them from blindly continuing through clearly adverse local changes. It also creates a clean handoff back to the outer controller: once a panic fires, the controller can choose a different high-level skill or drop to lower-level actions that are better matched to the new local state.

\subsection{Experimental skill set}
\label{app:experimental_skill_set}

We list here the skills exposed in the main experimental action space, followed by additional skills used in MiniHack-only settings. We group them by broad behavioral role.

\paragraph{Exploration, navigation, and search.}
\texttt{explore}, \texttt{goto\_corridor}, \texttt{goto\_corridor\_east}, \texttt{goto\_corridor\_north}, \texttt{goto\_corridor\_south}, \texttt{goto\_corridor\_west}, \texttt{goto\_room}, \texttt{goto\_room\_east}, \texttt{goto\_room\_north}, \texttt{goto\_room\_south}, \texttt{goto\_room\_west}, \texttt{descend\_stairs}, \texttt{ascend\_stairs}, \texttt{goto\_corpse}, \texttt{search\_corridor\_for\_hidden\_doors}, and \texttt{search\_room\_for\_hidden\_doors}.

\paragraph{Combat and tactical control.}
\texttt{engrave\_elbereth}, \texttt{fight\_engulfed}, \texttt{fight\_melee}, \texttt{fight\_ranged}, \texttt{zap\_monster}, \texttt{approach\_monster}, \texttt{goto\_choke\_point}, \texttt{wait\_for\_monster}, and \texttt{escape\_trap}.

\paragraph{Doors, terrain, and special environment interaction.}
\texttt{open\_doors}, \texttt{open\_doors\_kick}, \texttt{open\_doors\_key}, \texttt{goto\_boulder}, \texttt{push\_boulder\_east}, \texttt{push\_boulder\_north}, \texttt{push\_boulder\_south}, \texttt{push\_boulder\_west}, \texttt{leave\_shop}, \texttt{identify\_items\_altar}, and \texttt{dip\_for\_excalibur}.

\paragraph{Item acquisition and equipment management.}
\texttt{examine\_items}, \texttt{pickup\_amulet}, \texttt{pickup\_armor}, \texttt{pickup\_coin}, \texttt{pickup\_food}, \texttt{pickup\_gem}, \texttt{pickup\_potion}, \texttt{pickup\_ring}, \texttt{pickup\_scroll}, \texttt{pickup\_spellbook}, \texttt{pickup\_tool}, \texttt{pickup\_wand}, \texttt{pickup\_weapon}, \texttt{pickup\_corpse}, \texttt{puton\_ring}, \texttt{puton\_amulet}, \texttt{wear\_boots}, \texttt{wear\_cloak}, \texttt{wear\_gloves}, \texttt{wear\_helm}, \texttt{wear\_shield}, \texttt{wear\_shirt}, and \texttt{wear\_suit}.

\paragraph{Food, prompts, recovery, and cleanup.}
\texttt{eat\_corpse\_floor}, \texttt{eat\_corpse\_inventory}, \texttt{eat\_food\_inventory}, \texttt{yes}, \texttt{no}, \texttt{cancel}, \texttt{more}, \texttt{emergency\_escape}, \texttt{fix\_trouble}, \texttt{pray}, \texttt{heal}, \texttt{rest\_until\_full\_health}, \texttt{drop\_cursed\_items}, \texttt{drop\_unequipped\_armor}, \texttt{drop\_unequipped\_weapons}, \texttt{drop\_unidentified\_potions}, \texttt{drop\_unidentified\_scrolls}, \texttt{drop\_unidentified\_spellbooks}, and \texttt{drop\_unidentified\_wands}.

\paragraph{MiniHack-only skills.}
\texttt{cross\_lava\_river}, \texttt{freeze\_lava\_horn}, \texttt{freeze\_lava\_river},  \texttt{acquire\_levitation}, \texttt{approach\_lava\_river}, \texttt{levitate\_over\_lava\_river}, and \texttt{freeze\_lava\_wand}

%% file: sections/appendixB.tex
\section{Additional experimental details}
\label{app:exp-details}

This appendix provides additional details for~\cref{sec:experimental-setup}. We describe the NetHack and MiniHack environments in more detail, define the progress metric used for NetHack, list the MiniHack tasks used in our study, summarize the prompt templates used by the language agent, and report the main training hyperparameters for SFT and RL.

\subsection{Environment details}
\label{app:subsec:environment-details}

\paragraph{NetHack.}
NetHack is a classic and highly complex terminal roguelike game built around dungeon exploration, tactical combat, inventory management, and long-horizon planning. In the full game, the player must descend through more than 50 procedurally generated dungeon levels, retrieve the Amulet of Yendor, and then ascend back to the surface through several extremely difficult late-game regions. The NetHack Learning Environment (NLE) \citep{kuttler2020nethack} wraps the original game in a synchronous RL interface while leaving the underlying game dynamics unchanged. Because the environment is both procedurally generated and stochastic, it is designed to stress exploration, planning, credit assignment, and systematic generalization rather than memorization of fixed layouts.

At the level of the raw simulator, NLE is naturally multimodal. The default observation space includes symbolic views of the visible dungeon map together with textual and structured information such as the current message, the hero's bottom-line statistics, and the inventory. More concretely, the NLE paper describes the default observation as including \texttt{glyphs}, \texttt{chars}, \texttt{colors}, \texttt{specials}, \texttt{blstats}, \texttt{message}, and several inventory arrays (\texttt{inv\_glyphs}, \texttt{inv\_strs}, \texttt{inv\_letters}, and \texttt{inv\_oclasses}). In our setup, these low-level observations are not passed directly to the language model. Instead, we use the NLE language wrapper to convert them into natural-language descriptions and an ASCII map, producing the text-based interface described in~\cref{sec:experimental-setup} and~\cref{app:exp-details}.

The original NLE paper defines a full action space of 93 discrete actions, consisting of 77 command actions and 16 movement actions. However, interacting with NetHack through its terminal interface often requires \emph{action chaining}: many semantically simple behaviors are implemented as sequences of low-level key presses. For example, commands such as throwing, zapping, or applying an item typically require an initial command followed by one or more menu selections, inventory letters, prompt confirmations, or directions. This substantially increases the difficulty of primitive control because an agent must not only decide \emph{what} to do, but also produce the correct sequence of interface-level actions needed to complete the command. In our code-wrapper setup, this is reflected in the primitive language interface, which exposes an expanded set of textual action strings including command names, directional movement, prompt responses, and menu-selection characters.

Our main NetHack training environment is \texttt{NetHackScore-v0}. In this task, the environment reward is the change in the in-game NetHack score between successive timesteps. As emphasized by the NLE paper, this score is a sensible proxy for incremental progress because it depends on factors such as dungeon depth, enemy kills, gold collection, and object knowledge. At the same time, score is not perfectly aligned with actually solving the game: expert human players can ascend while deliberately keeping score relatively low. For this reason, in addition to score we also report the BALROG NLE-progress metric described below.

Procedural generation also matters for the interpretation of our results. Since each episode is generated from a fresh random seed, the agent is highly unlikely to encounter exactly the same level twice. The goal is therefore not to memorize solutions to specific layouts, but to learn reusable behavioral patterns that transfer across unseen seeds. Moreover, progression in NetHack is not strictly linear: the agent may backtrack, revisit earlier levels, or enter branching sub-dungeons such as the Gnomish Mines. This means that even within a single episode, the agent must preserve and reuse knowledge across changing contexts rather than merely advance through a fixed curriculum of stages. In our experiments, NetHack serves as the main long-horizon domain of study, and our SFT and RL experiments focus on this setting.

Unless noted otherwise, NetHack runs use the \texttt{val-hum-law-fem} character, disable autopickup, keep the pet enabled, allow unrestricted menu and \texttt{yn} responses, skip repeated \texttt{--More--} prompts automatically, and use a 5{,}000-step episode limit with a 150-step no-progress timeout.

\paragraph{NetHack progress metric.}
For NetHack, in addition to raw environment score, we report the NLE-progress metric introduced in BALROG \citep{paglieri2025balrog}. The motivation is that in-game NetHack score does not always align with meaningful progress toward winning the game. BALROG therefore defines a data-driven progression metric from human-played NetHack trajectories. Let \(d^\star\) denote the highest dungeon level reached in an episode and \(x^\star\) the highest experience level reached. BALROG maps each of these to a probability of eventual human ascension using empirical progression curves, with Dungeon Level 1 and Experience Level 1 defined as \(0\%\) progress and ascension defined as \(100\%\). The episode-level progress score is then
\[
\mathrm{progress} = \max\bigl(p_{\mathrm{DLvl}}(d^\star),\; p_{\mathrm{xlvl}}(x^\star)\bigr),
\]
where \(p_{\mathrm{DLvl}}\) and \(p_{\mathrm{xlvl}}\) are the BALROG progression curves. We use this metric as our main measure of human-like game progress in NetHack.

\paragraph{MiniHack.}
MiniHack \citep{samvelyan2021minihack} is a flexible sandbox built on top of NLE that allows researchers to define controlled NetHack-style tasks with custom layouts, monsters, terrain, and objects. Following BALROG \citep{paglieri2025balrog}, we use MiniHack as a complementary zero-shot testbed because it preserves the mechanics of NetHack while making failures easier to interpret. To interface with language models, we use the same NLE language wrapper as in NetHack, so both domains are presented through a shared text-based observation and action interface.

\subsection{MiniHack task suite}
\label{app:subsec:minihack-task-suite}

Our MiniHack evaluation suite covers seven tasks spanning exploration, combat, inventory usage, and long-horizon subgoal sequencing. The standard Corridor and Quest families come directly from MiniHack, while our Corridor-R10 evaluation level is a custom extension of the Corridor family and simulates exploration of a single NetHack level.

\paragraph{Corridor-R3 (\texttt{MiniHack-Corridor-R3-v0}).}

\begin{figure}[t]
\centering
\includegraphics[width=\linewidth]{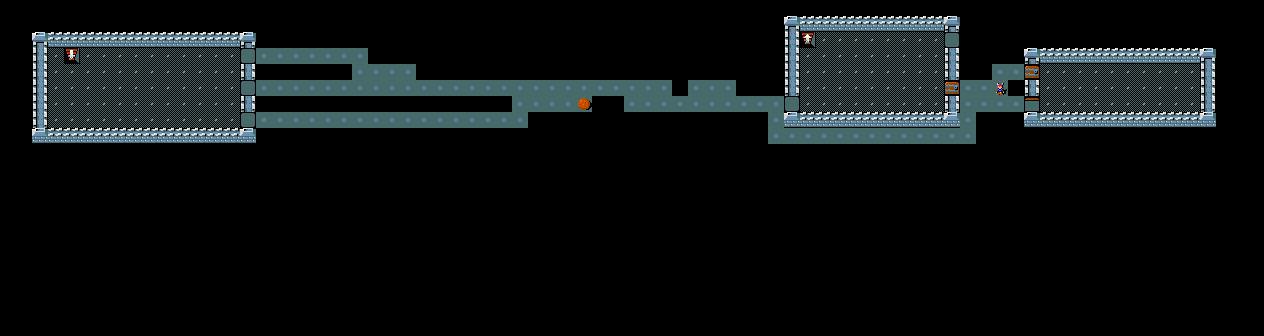}
\caption{Example layout from \texttt{MiniHack-Corridor-R3-v0}.}
\label{fig:minihack-corridor-r3}
\end{figure}

Corridor-R3 belongs to the MiniHack Corridor family of exploration tasks. The objective is to reach the staircase located in one of the rooms, but the room positions, room sizes, and connecting corridors are procedurally generated. In the R3 version, the level is composed of three rooms. To solve it reliably, the agent has to navigate rooms and corridors, open doors, including kicking closed ones when needed, and search for hidden doors and hidden corridors that may block progress. This creates a short-horizon exploration problem in which the agent must execute the core mechanics of NetHack navigation in a relatively small layout. Corridor-R3 is useful for exposing brittle action repetition and basic exploration failures under primitive control.~\cref{fig:minihack-corridor-r3} shows a representative state.

\begin{figure}[t]
\centering
\includegraphics[width=\linewidth]{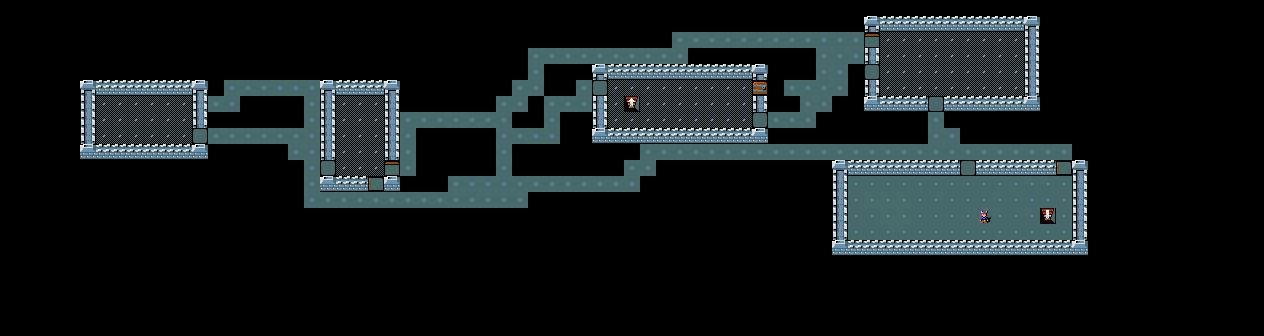}
\caption{Example layout from \texttt{MiniHack-Corridor-R5-v0}}
\label{fig:minihack-corridor-r5}
\end{figure}

\paragraph{Corridor-R5 (\texttt{MiniHack-Corridor-R5-v0}).}
Corridor-R5 uses the same basic construction as Corridor-R3, but increases the number of rooms to five. This directly increases the amount of exploration needed before the staircase can be found, and it raises the cost of wasted movement or local oscillation. Compared with R3, the task places more pressure on maintaining directional progress, remembering which parts of the map have already been explored, and recovering from short detours without becoming trapped in repetitive low-level behaviors.~\cref{fig:minihack-corridor-r5} shows a representative state.

\begin{figure}[t]
\centering
\includegraphics[width=\linewidth]{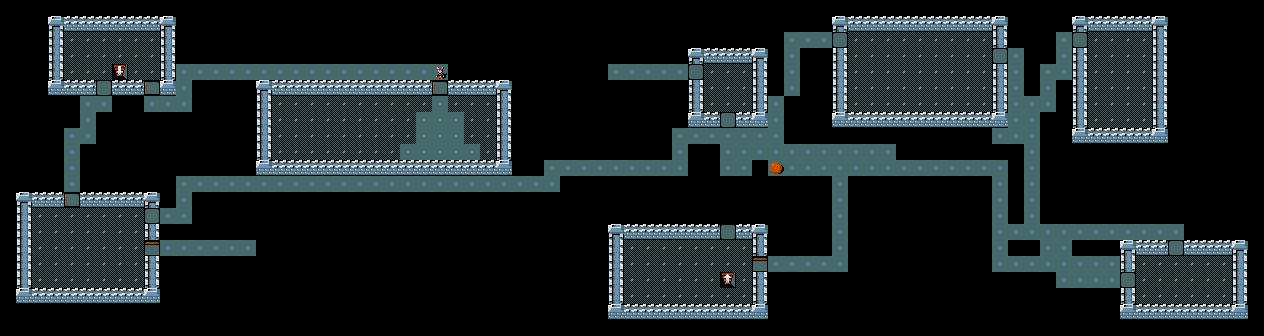}
\caption{Example layout from \texttt{MiniHack-Corridor-R10-v0}}
\label{fig:minihack-corridor-r10}
\end{figure}

\paragraph{Corridor-R10 (\texttt{MiniHack-Corridor-R10-v0}).}
Corridor-R10 is a custom extension of the standard Corridor family that further increases the number of rooms. This task is particularly useful because it tests whether an agent can explore a normal NetHack-style corridor layout while disentangling exploration from survival. The challenge is not to win difficult fights or manage complex item interactions, but to make steady exploratory progress over a longer navigation horizon.~\cref{fig:minihack-corridor-r10} shows a representative state.

\begin{figure}[t]
\centering
\includegraphics[width=\linewidth]{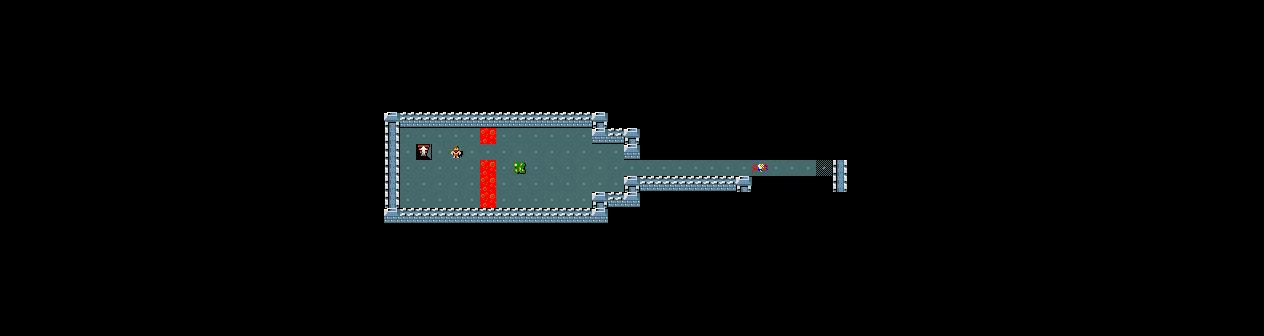}
\caption{Example layout for \texttt{MiniHack-Quest-Easy-v0}.}
\label{fig:minihack-quest-easy}
\end{figure}

\paragraph{Quest-Easy (\texttt{MiniHack-Quest-Easy-v0}).}
This task is the simplest member of the Quest family. The layout is relatively simple and mostly fixed, so the main difficulty is not large-scale search but correctly sequencing a short multi-stage plan. The agent must explore the map, acquire or use an object that allows it to cross a lava river, survive a small amount of combat, and then reach the staircase. Compared with pure navigation tasks, Quest-Easy introduces basic item usage and simple combat.~\cref{fig:minihack-quest-easy} shows a representative state.

\begin{figure}[t]
\centering
\includegraphics[width=\linewidth]{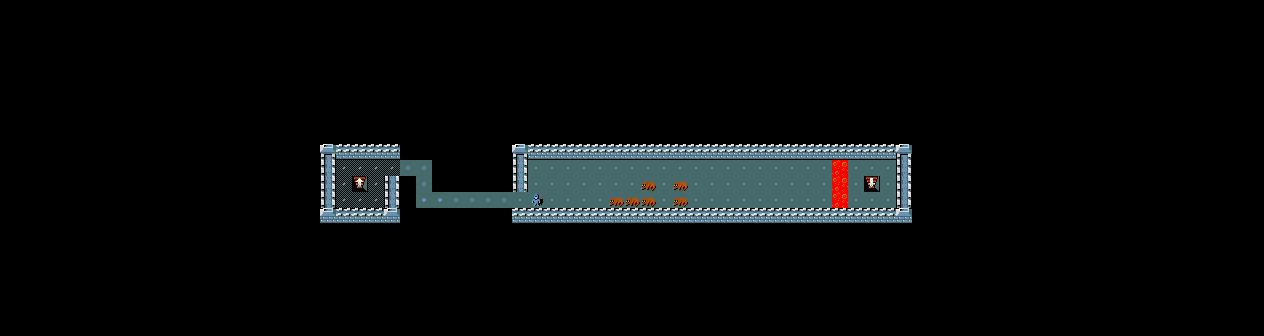}
\caption{Example layout for \texttt{MiniHack-Quest-Medium-v0}.}
\label{fig:minihack-quest-medium}
\end{figure}

\paragraph{Quest-Medium (\texttt{MiniHack-Quest-Medium-v0}).}
Quest-Medium increases the difficulty of combat and subgoal coordination. In this variant, the agent must fight a swarm of monsters in a narrow corridor, where careless forward movement can easily lead to being surrounded. At the same time, the environment tests whether the agent can recognize that an item may be useful for a later subgoal rather than immediately: after surviving the encounter with monsters, it must use the item found at the beginning of the level, and use it to cross the lava river. Only then can it reach the staircase. Successful behavior therefore requires both corridor-aware combat and delayed item use.~\cref{fig:minihack-quest-medium} shows a representative state.

\begin{figure}[t]
\centering
\includegraphics[width=\linewidth]{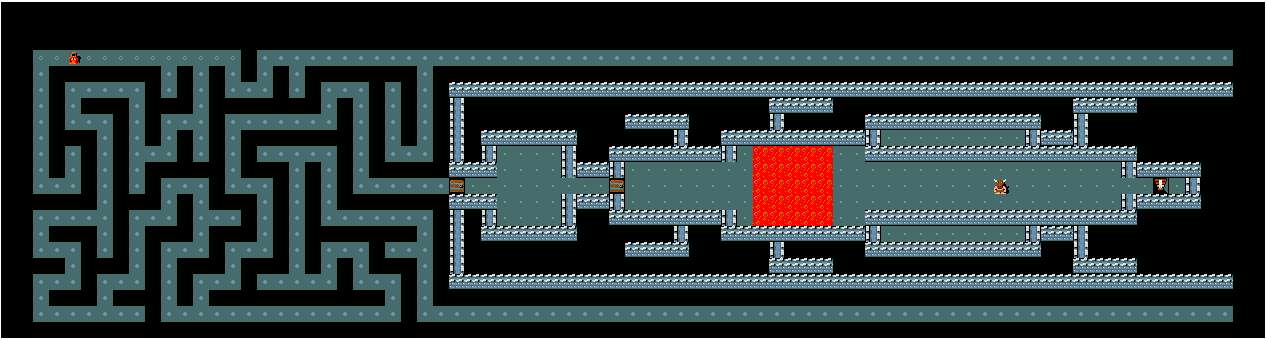}
\caption{Example layout for \texttt{MiniHack-Quest-Hard-v0}.}
\label{fig:minihack-quest-hard}
\end{figure}

\paragraph{Quest-Hard (\texttt{MiniHack-Quest-Hard-v0}).}
Quest-Hard is the most demanding task in this family. It begins with a large procedurally generated maze that must be solved before the rest of the quest can even begin. If the agent spends too much time exploring it will die because of hunger forcing the agent to be efficient. After escaping the maze, the agent still needs to collect and use items to cross the lava river, and finally use a wand of death to kill a minotaur guarding the staircase. Overall, Quest-Hard is a long-horizon problem that requires chaining together multiple skills, including efficient exploration, item collection, lava crossing, and combat.~\cref{fig:minihack-quest-hard} shows a representative state.

\begin{figure}[t]
\centering
\includegraphics[width=\linewidth]{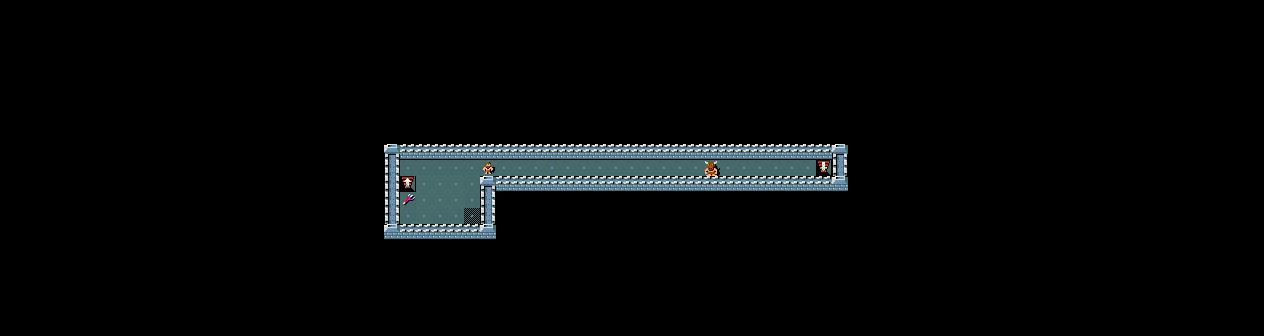}
\caption{Example layout for \texttt{MiniHack-WoD-Hard-Full-v0}.}
\label{fig:minihack-wod-hard-full}
\end{figure}

\paragraph{WoD-Hard-Full (\texttt{MiniHack-WoD-Hard-Full-v0}).}
This task belongs to the Wand-of-Death family. The central mechanic is that the agent must use a wand of death correctly to eliminate a dangerous monster that blocks progress to the staircase. In the hard variant, the wand must first be found or acquired before it can be used, and used to kill the minotaur. As a result, the task combines item pickup, item use, and combat. This is useful because correct wand-of-death usage is itself an important subtask inside \texttt{MiniHack-Wod-Hard-Full-v0}.~\cref{fig:minihack-wod-hard-full} shows a representative state.

\subsection{Prompt templates and evaluation protocol}
\label{app:subsec:prompt-templates-and-eval-protocol}

For each task, we use the same observation format and prompt template across \emph{primitive}, \emph{skill}, and \emph{mixed} settings. The available-action lists and descriptions change to match the selected interface.

\paragraph{System prompt.}
The system prompt has the following structure:
\begin{verbatim}
You are an expert NetHack player. Your task is to: {task_goal}

<task_instruction>
- Analyze the observation.
- Choose the best immediate action from the command/tool list.
- Provide ONLY the action command inside <action> tags.
</task_instruction>

<commands>...</commands>    (primitive or mixed settings)
<tools>...</tools>          (skill or mixed settings)
<tips>...</tips>
<output_format><action>...</action></output_format>
\end{verbatim}

In the primitive setting, \texttt{<commands>} contains the low-level action list. In the skill setting, \texttt{<tools>} contains the available \skillLibraryName{} skills together with short docstring-based descriptions. In the mixed setting, both blocks are present.

\paragraph{User turn structure.}
At each timestep \(t\), the model receives a chat-style user message containing recent action history and the current text observation:
\begin{verbatim}
<action_history>...</action_history>
<current_state>
  <message_log>...</message_log>
  <cursor>...</cursor>
  <map description>...</map description>
  <stats>...</stats>
  <map>...</map>
  <language_observation>...</language_observation>
  <inventory>...</inventory>
  <feedback>...</feedback> (optional)
</current_state>
\end{verbatim}

The \texttt{<feedback>} block is used to tell the agent when its previous output was malformed or mapped to an invalid action. The model must answer with a single action string enclosed in \texttt{<action>} tags.

\paragraph{History and fallback.}
In the main setup used for zero-shot evaluation, SFT, and RL, the agent keeps five interaction turns of history and is given the two most recent observations. If the model output does not contain a valid \texttt{<action>} block, the environment replaces it with a default action and tells the agent that its previous action was invalid. This fallback mechanism prevents brittle formatting failures from terminating evaluation episodes.

\subsection{Training details}
\label{app:subsec:training-details}

~\cref{tab:training-hparams} summarizes the main hyperparameters used in SFT and RL. Unless noted otherwise, the reported SFT and RL experiments use a 5{,}000-step episode limit and set the code-wrapper's maximum skill horizon to 100{,}000 primitive steps.

During SFT, we train a LoRA adapter on top of the base language model. In the current setup, this uses rank-stabilized LoRA with rank 128 and scaling 64. After SFT, the learned adapter is merged back into the base model. This merged model is then used as the initialization for the SFT-primed RL experiments described below.

For RL from SFT, the model is initialized from the SFT-primed model obtained by merging the SFT adapter into the base model. For RL from base, the model is trained directly from the base model without SFT warm-starting. The default RL configuration uses rank-stabilized LoRA with rank 128 and scaling 64, 32 rollouts per trainer GPU, 16 controller decisions per rollout, 350 gradient steps, \(\lambda = 0.95\), clip range 0.2, and policy and value learning rates of $2 \times 10^{-6}$.

We use Ray to orchestrate data collection and training, vLLM for inference, and PyTorch for training. Each environment worker maintains a separate environment instance and sends its observation and interaction history to vLLM to generate the next action. We collect fixed-length rollouts, storing the prompt, generated tokens, reward, termination flags, and the sequence log-probability returned by vLLM. Once a batch is complete, the trainer computes value estimates and generalized advantage estimates, then updates the policy and value models using PPO minibatches.

We use double buffering to overlap data collection with training: while the trainer updates on one batch, the environment workers collect the next. This reduces idle time but introduces policy lag, as the collected data may come from an earlier version of the policy. Numerical differences between vLLM and the PyTorch training backend can also produce different action probabilities, even with the same model weights. Such training inference mismatch has been identified as a source of instability~\citep{liu-li-2025-rl-collapse,li2025trust}. To account for both sources of off-policy mismatch, we compute the PPO importance ratio using the log-probabilities recorded by vLLM during collection.

Each reinforcement learning and supervised fine-tuning run was completed in under one day using 8xH100 GPUs, while each zero-shot and evaluation run was completed in under one day using a single H100 GPU.

\input{tables/training_hparams_v1}

%% file: tables/training_hparams_v1.tex
\begin{table}[t]
\centering
\small
\begin{tabular}{|p{0.18\linewidth}|p{0.34\linewidth}|p{0.38\linewidth}|}
\hline
\textbf{Stage} & \textbf{Hyperparameter} & \textbf{Value} \\
\hline
SFT & Teacher model & Gemma-4-31B \\
\hline
SFT & Teacher episodes & 1{,}024 zero-shot episodes \\
\hline
SFT & Trainable parameters & LoRA adapter only \\
\hline
SFT & LoRA rank / alpha / dropout & 128 / 64 / 0.0 \\
\hline
SFT & Learning rate & \(5 \times 10^{-5}\) \\
\hline
SFT & Local minibatch size & 64 \\
\hline
SFT & Per-device train batch size & 2 \\
\hline
SFT & Training epochs & 1 \\
\hline
SFT & Evaluation frequency & every 8{,}192 samples \\
\hline
SFT & Max generated action tokens & 32 in the non-thinking setup \\
\hline
RL & Algorithm & PPO \\
\hline
RL & Initializations compared & base model and merged SFT model \\
\hline
RL & Trainable parameters & LoRA adapters for both policy and value \\
\hline
RL & LoRA rank / alpha / dropout & 128 / 64 / 0.0 \\
\hline
RL & Policy / value learning rates & \(2 \times 10^{-6}\) / \(2 \times 10^{-6}\) \\
\hline
RL & KL coefficient & 0.0 \\
\hline
RL & Rollouts per trainer GPU & 32 \\
\hline
RL & Controller decisions per rollout & 16 \\
\hline
RL & Local minibatch size & 256 \\
\hline
RL & Per-device train batch size & 8 \\
\hline
RL & Gradient steps & 350 \\
\hline
RL & GAE \(\lambda\) & 0.95 \\
\hline
RL & PPO clip range & 0.2 \\
\hline
RL & Entropy coefficient & 0.0 \\
\hline
RL & Number of seeds & 3 \\
\hline
\end{tabular}
\caption{Main SFT and RL hyperparameters used in our experiments.}
\label{tab:training-hparams}
\end{table}

%% file: sections/appendixC.tex
\clearpage

\section{Full zero-shot NetHack results}
\label{app:zero-shot-results}

For completeness,~\cref{tab:nethack-zero-shot-summary} reports the full aggregate zero-shot NetHack results for every model and action interface used in our sweep. We use reasoning mode for GPT-5, GPT-OSS-120B, Gemma-4, and Qwen-3.5, and act-only mode for Llama and Gemma3.

\input{tables/nethack_zero_shot_summary_v4}

\clearpage

\input{tables/nethack_skills_vs_primitives_v5}

\paragraph{Additional zero-shot tradeoff plots.}
For completeness, we also report the remaining pairwise zero-shot tradeoff plots relating NetHack performance to inference cost and token usage. The main text already includes maximum dungeon level versus cost (~\cref{fig:nethack-zero-shot-results}); here we additionally show maximum dungeon level versus tokens, progression versus cost and tokens, and score versus cost and tokens.

\begin{figure}[t]
    \centering
    \includegraphics[width=\linewidth]{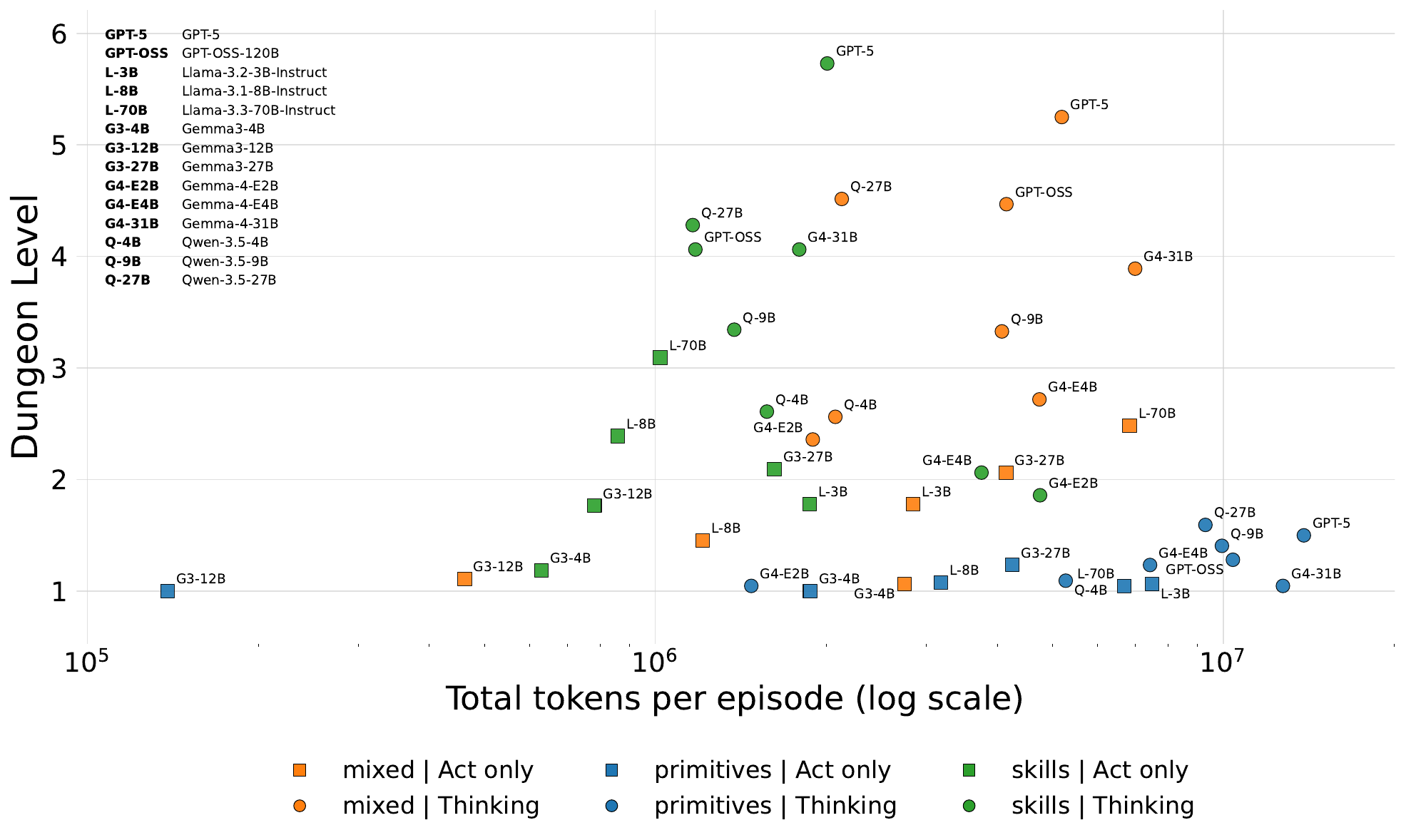}
    \caption{\textbf{Additional zero-shot NetHack tradeoff plot: average maximum dungeon level reached versus tokens per episode.} As in the main-text cost-based view, skill-based control defines a stronger performance-efficiency frontier than primitive-only control, while mixed control is typically intermediate.}
    \label{fig:nethack-zero-shot-dlvl-cost}
\end{figure}

\begin{figure}[t]
    \centering
    \includegraphics[width=\linewidth]{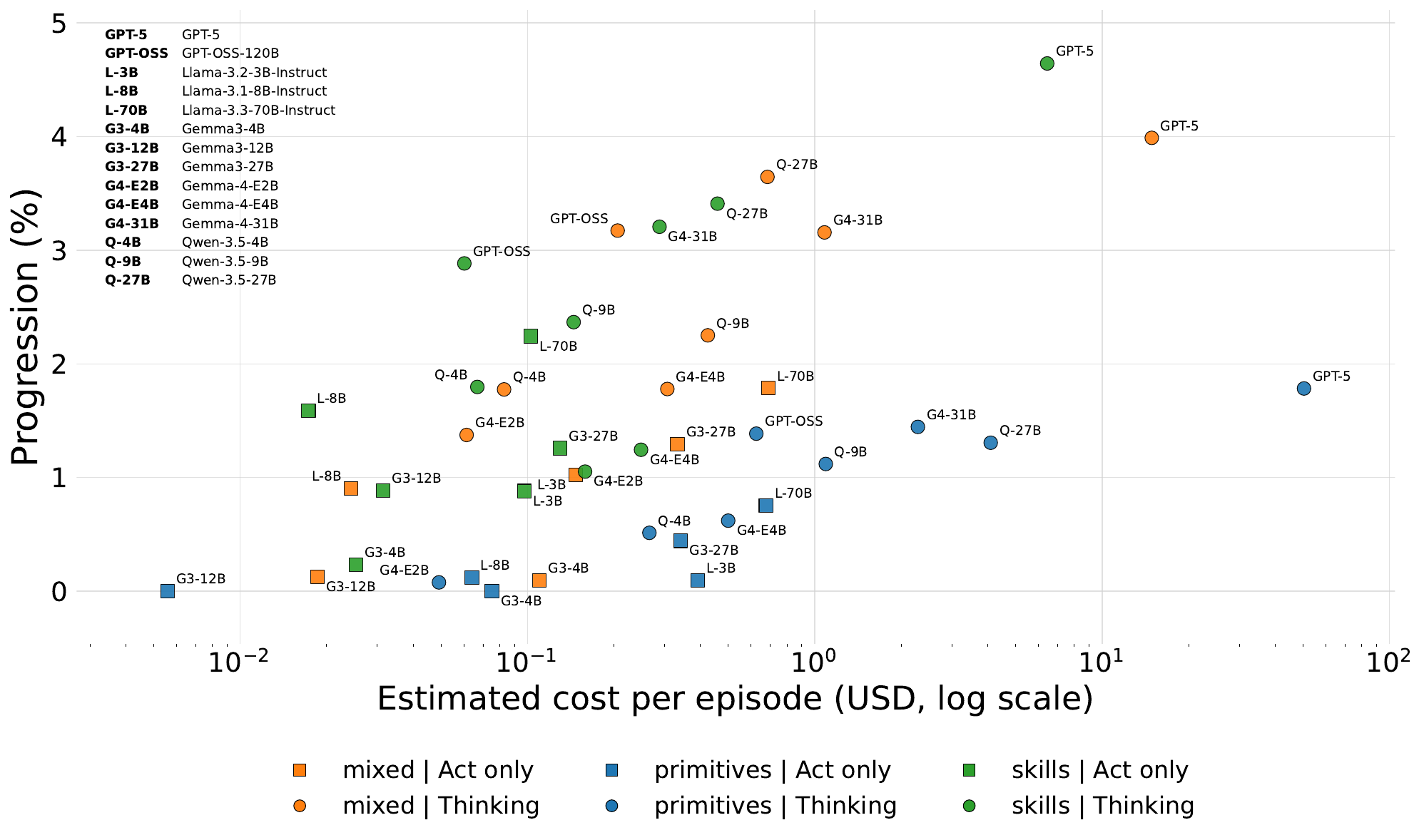}
    \caption{\textbf{Additional zero-shot NetHack tradeoff plot: progression versus inference cost per episode.} Skill-based control generally achieves higher progression at lower or comparable cost than primitive-only control.}
    \label{fig:nethack-zero-shot-progression-cost}
\end{figure}

\begin{figure}[t]
    \centering
    \includegraphics[width=\linewidth]{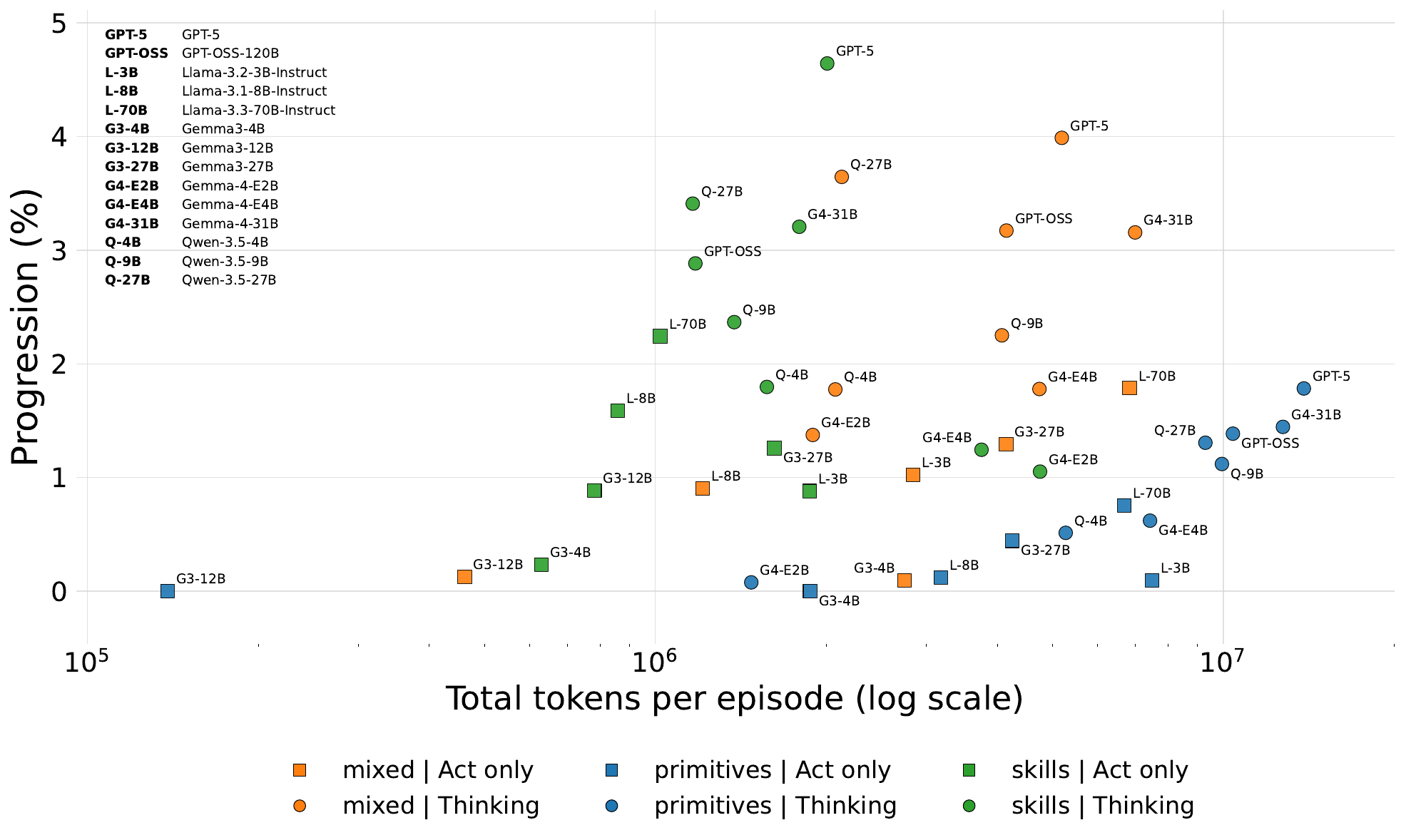}
    \caption{\textbf{Additional zero-shot NetHack tradeoff plot: progression versus token usage per episode.} Skills remain favorable when efficiency is measured in tokens rather than monetary cost.}
    \label{fig:nethack-zero-shot-progression-tokens}
\end{figure}

\begin{figure}[t]
    \centering
    \includegraphics[width=\linewidth]{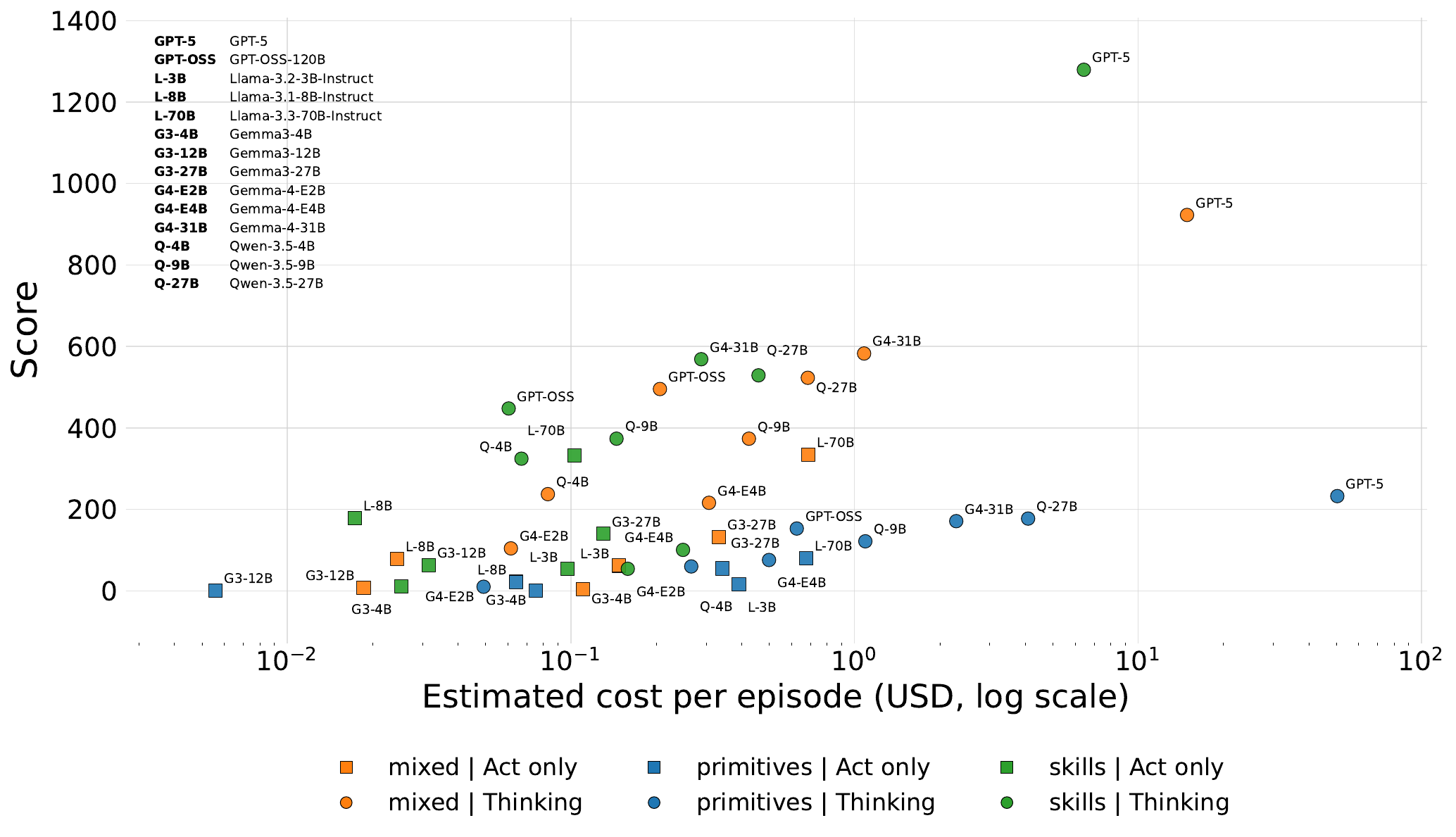}
    \caption{\textbf{Additional zero-shot NetHack tradeoff plot: score versus inference cost per episode.} Across models, skill-based control tends to produce higher scores at lower or comparable cost than primitive-only control.}
    \label{fig:nethack-zero-shot-score-cost}
\end{figure}

\begin{figure}[t]
    \centering
    \includegraphics[width=\linewidth]{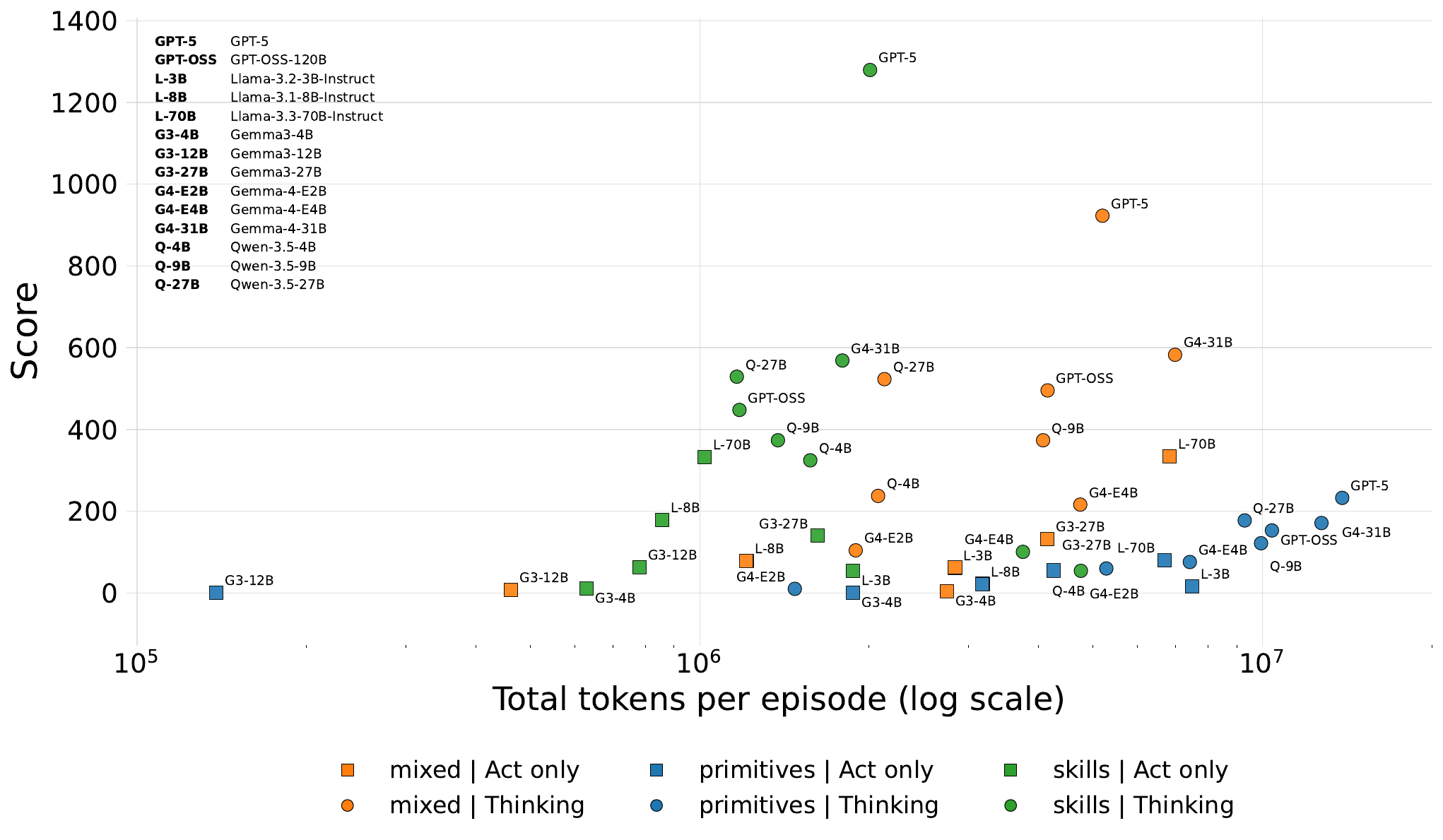}
    \caption{\textbf{Additional zero-shot NetHack tradeoff plot: score versus token usage per episode.} The same qualitative advantage of skills persists when efficiency is measured in total input and output tokens.}
    \label{fig:nethack-zero-shot-score-tokens}
\end{figure}

\clearpage
\subsection{Sensitivity to incomplete skill coverage}
\label{app:skill-coverage}

We ask how skill-only and mixed controllers respond when the supplied library lacks a family of useful behaviors. Here, \emph{coverage} refers to whether the available skills provide the behaviors needed in the situations the agent encounters. Starting from the full 78-skill library, we remove one of the five behavioral families listed in~\cref{app:experimental_skill_set} at a time, keeping the language model and runtime fixed. We evaluate zero-shot \texttt{Gemma-4-31B} on NetHack under skill-only and mixed control over 64 episodes each. In mixed control, primitive commands remain available after each removal, allowing the agent to attempt behaviors no longer supplied as skills.

\begin{figure}[htbp]
    \centering
    \includegraphics[width=\linewidth]{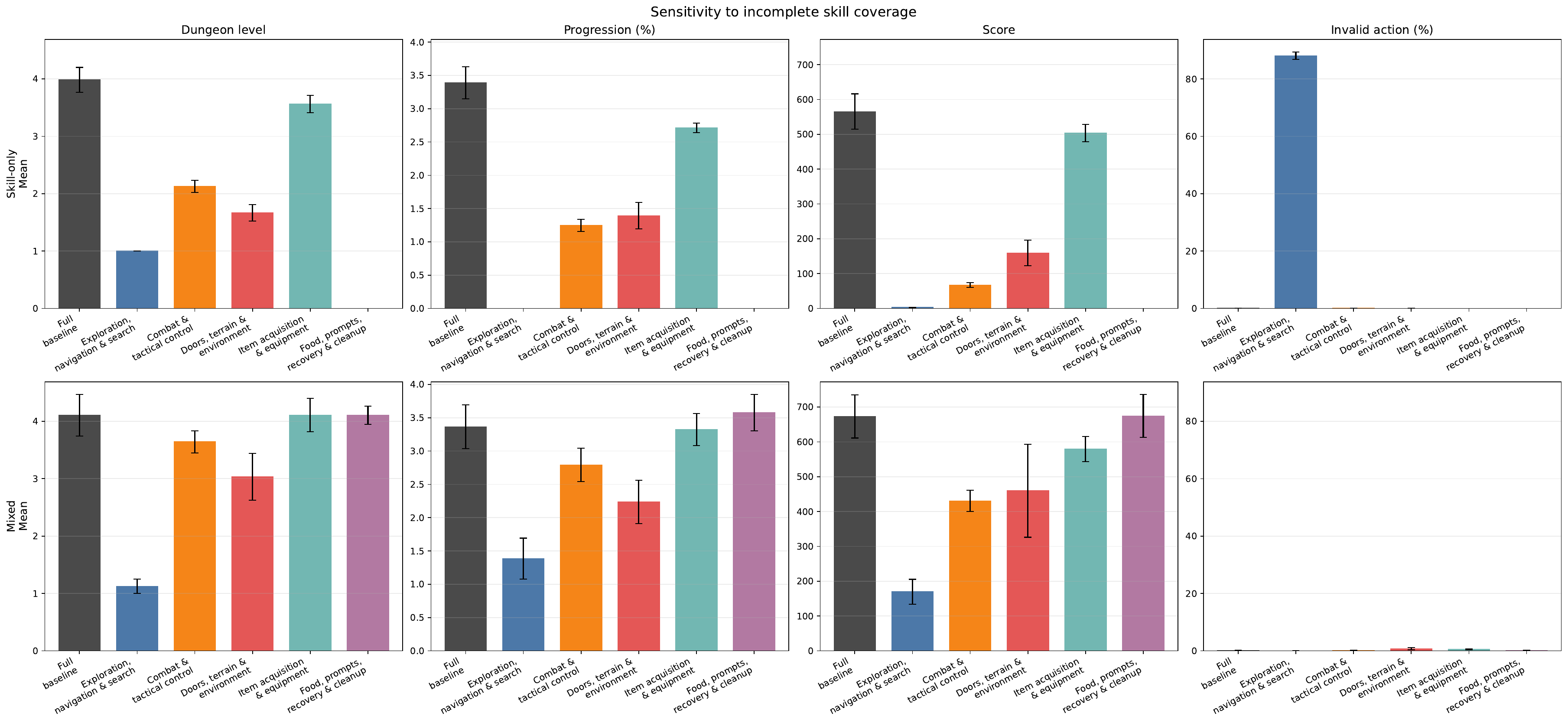}
    \caption{\textbf{Sensitivity to incomplete skill coverage for zero-shot \texttt{Gemma-4-31B} on NetHack.} The full baseline uses all 78 skills; each other condition removes the named skill family. Mixed control retains primitive actions. Bars show means over 64 episodes per condition. The comparison measures sensitivity to missing skill families and the extent to which primitive alternatives preserve performance.}
    \label{fig:skill-coverage}
\end{figure}

The skill-only results in~\cref{fig:skill-coverage} show the importance of covering essential behaviors. Removing exploration, navigation, and search also removes staircase skills and severely restricts the agent's ability to explore and advance through the dungeon. The resulting drop in progression demonstrates dependence on coverage of these behaviors.

Mixed control lets us examine the value of supplying a skill family when primitive alternatives remain available. Family removals are generally less damaging in this setting, consistent with primitive fallback helping the controller cope with incomplete coverage. Removing exploration/navigation/search still produces the largest decrease in progression and dungeon level, followed by doors/terrain/environment and then combat/tactical control. Access to primitives therefore mitigates the loss of supplied skills, but does not guarantee successful recovery.

\clearpage
\subsection{Qualitative analysis of mixed control}
\label{app:qualitative-analysis}

To examine how agents move between abstraction levels, we analyze zero-shot \texttt{GPT-5} trajectories on NetHack under mixed control. We identify sequences in which primitives complete menus, recover from failed skills, or supply behaviors missing from the library, after which the agent returns to skills. Two examples from the videos on our \href{https://bartekcupial.github.io/abstraction-ladder/\#rollouts}{project page} illustrate this behavior (\cref{tab:qualitative-mixed-control}).

\input{tables/qualitative_mixed_control_v1}

The ranged-combat example shows recovery after an existing skill fails to complete a local subproblem. The vault example shows how primitives supply a behavior absent from the skill library. In both cases, the agent resolves the immediate problem and returns to higher-level control. These sequences support the motivation for mixed control: primitive fallback can enable the agent to move down the abstraction ladder when a skill is insufficient and back up once the problem is resolved.

Mixed control also introduces a routing problem: the controller must decide both what to do and whether to use a skill or a primitive. The vault sequence illustrates that this choice can be inefficient. Before dropping the gold, \texttt{GPT-5} opens and cancels unrelated menus, calls \texttt{explore} without advancing the environment, and moves north and back south. This is consistent with our observation that \texttt{GPT-5} sometimes switches too late or uses primitives aimlessly. The selected zero-shot examples show that useful recoveries are possible; their frequency and how RL changes these decisions require further analysis.

%% file: tables/nethack_zero_shot_summary_v4.tex
\begingroup
    \tiny
    \setlength{\tabcolsep}{0.85pt}
\begin{longtable}{llrrrrrrrrrrr}
    \caption{Aggregate NetHack zero-shot results for the same 14 models and thinking modes as~\cref{tab:nethack-zero-shot-averages}. Gray $\pm$ terms are standard errors across episodes. Progression is in percent; milestone reach values are fractions. All settings use 64 evaluation episodes except \texttt{GPT-5}, which uses 32 episodes per action interface due to cost. Mines reach is computed from episode traces.}\label{tab:nethack-zero-shot-summary} \\
        \toprule
        Model & Action & Progression & Score & Dlvl & L2 & L4 & L6 & L8 & Oracle & Mines & Cost/ep. (\$) & Tokens/ep. \\
        \midrule
    \endfirsthead
        \toprule
        Model & Action & Progression & Score & Dlvl & L2 & L4 & L6 & L8 & Oracle & Mines & Cost/ep. (\$) & Tokens/ep. \\
        \midrule
    \endhead
        \midrule
        \multicolumn{13}{r}{Continued on next page} \\
    \endfoot
        \bottomrule
    \endlastfoot
        GPT-5 & skills & 4.64 {\tiny\scalebox{0.7}{\textcolor{gray}{$\pm$ 0.38}}} & 1279.29 {\tiny\scalebox{0.7}{\textcolor{gray}{$\pm$ 151.62}}} & 5.73 {\tiny\scalebox{0.7}{\textcolor{gray}{$\pm$ 0.29}}} & 0.94 {\tiny\scalebox{0.7}{\textcolor{gray}{$\pm$ 0.03}}} & 0.84 {\tiny\scalebox{0.7}{\textcolor{gray}{$\pm$ 0.05}}} & 0.54 {\tiny\scalebox{0.7}{\textcolor{gray}{$\pm$ 0.06}}} & 0.25 {\tiny\scalebox{0.7}{\textcolor{gray}{$\pm$ 0.06}}} & 0.06 {\tiny\scalebox{0.7}{\textcolor{gray}{$\pm$ 0.03}}} & 0.47 {\tiny\scalebox{0.7}{\textcolor{gray}{$\pm$ 0.06}}} & 6.44 & 2,007,700 \\
         & mixed & 3.99 {\tiny\scalebox{0.7}{\textcolor{gray}{$\pm$ 0.54}}} & 922.84 {\tiny\scalebox{0.7}{\textcolor{gray}{$\pm$ 179.63}}} & 5.25 {\tiny\scalebox{0.7}{\textcolor{gray}{$\pm$ 0.42}}} & 0.94 {\tiny\scalebox{0.7}{\textcolor{gray}{$\pm$ 0.04}}} & 0.78 {\tiny\scalebox{0.7}{\textcolor{gray}{$\pm$ 0.07}}} & 0.38 {\tiny\scalebox{0.7}{\textcolor{gray}{$\pm$ 0.09}}} & 0.22 {\tiny\scalebox{0.7}{\textcolor{gray}{$\pm$ 0.07}}} & 0.06 {\tiny\scalebox{0.7}{\textcolor{gray}{$\pm$ 0.04}}} & 0.53 {\tiny\scalebox{0.7}{\textcolor{gray}{$\pm$ 0.09}}} & 14.88 & 5,196,745 \\
         & primitives & 1.78 {\tiny\scalebox{0.7}{\textcolor{gray}{$\pm$ 0.13}}} & 232.56 {\tiny\scalebox{0.7}{\textcolor{gray}{$\pm$ 22.59}}} & 1.50 {\tiny\scalebox{0.7}{\textcolor{gray}{$\pm$ 0.14}}} & 0.34 {\tiny\scalebox{0.7}{\textcolor{gray}{$\pm$ 0.09}}} & 0.03 {\tiny\scalebox{0.7}{\textcolor{gray}{$\pm$ 0.03}}} & 0.00 {\tiny\scalebox{0.7}{\textcolor{gray}{$\pm$ 0.00}}} & 0.00 {\tiny\scalebox{0.7}{\textcolor{gray}{$\pm$ 0.00}}} & 0.00 {\tiny\scalebox{0.7}{\textcolor{gray}{$\pm$ 0.00}}} & 0.03 {\tiny\scalebox{0.7}{\textcolor{gray}{$\pm$ 0.03}}} & 50.41 & 13,866,723 \\
        \midrule
        GPT-OSS-120B & skills & 2.88 {\tiny\scalebox{0.7}{\textcolor{gray}{$\pm$ 0.17}}} & 447.72 {\tiny\scalebox{0.7}{\textcolor{gray}{$\pm$ 39.81}}} & 4.06 {\tiny\scalebox{0.7}{\textcolor{gray}{$\pm$ 0.23}}} & 0.95 {\tiny\scalebox{0.7}{\textcolor{gray}{$\pm$ 0.03}}} & 0.61 {\tiny\scalebox{0.7}{\textcolor{gray}{$\pm$ 0.06}}} & 0.22 {\tiny\scalebox{0.7}{\textcolor{gray}{$\pm$ 0.05}}} & 0.05 {\tiny\scalebox{0.7}{\textcolor{gray}{$\pm$ 0.03}}} & 0.03 {\tiny\scalebox{0.7}{\textcolor{gray}{$\pm$ 0.02}}} & 0.45 {\tiny\scalebox{0.7}{\textcolor{gray}{$\pm$ 0.06}}} & 0.06 & 1,176,438 \\
         & mixed & 3.17 {\tiny\scalebox{0.7}{\textcolor{gray}{$\pm$ 0.22}}} & 495.48 {\tiny\scalebox{0.7}{\textcolor{gray}{$\pm$ 41.31}}} & 4.47 {\tiny\scalebox{0.7}{\textcolor{gray}{$\pm$ 0.25}}} & 0.95 {\tiny\scalebox{0.7}{\textcolor{gray}{$\pm$ 0.03}}} & 0.67 {\tiny\scalebox{0.7}{\textcolor{gray}{$\pm$ 0.06}}} & 0.33 {\tiny\scalebox{0.7}{\textcolor{gray}{$\pm$ 0.06}}} & 0.06 {\tiny\scalebox{0.7}{\textcolor{gray}{$\pm$ 0.03}}} & 0.05 {\tiny\scalebox{0.7}{\textcolor{gray}{$\pm$ 0.03}}} & 0.36 {\tiny\scalebox{0.7}{\textcolor{gray}{$\pm$ 0.06}}} & 0.21 & 4,151,999 \\
         & primitives & 1.39 {\tiny\scalebox{0.7}{\textcolor{gray}{$\pm$ 0.11}}} & 152.97 {\tiny\scalebox{0.7}{\textcolor{gray}{$\pm$ 14.46}}} & 1.28 {\tiny\scalebox{0.7}{\textcolor{gray}{$\pm$ 0.08}}} & 0.20 {\tiny\scalebox{0.7}{\textcolor{gray}{$\pm$ 0.05}}} & 0.02 {\tiny\scalebox{0.7}{\textcolor{gray}{$\pm$ 0.02}}} & 0.00 {\tiny\scalebox{0.7}{\textcolor{gray}{$\pm$ 0.00}}} & 0.00 {\tiny\scalebox{0.7}{\textcolor{gray}{$\pm$ 0.00}}} & 0.00 {\tiny\scalebox{0.7}{\textcolor{gray}{$\pm$ 0.00}}} & 0.00 {\tiny\scalebox{0.7}{\textcolor{gray}{$\pm$ 0.00}}} & 0.63 & 10,399,676 \\
        \midrule
        Llama-3.2-3B-Instruct & skills & 0.88 {\tiny\scalebox{0.7}{\textcolor{gray}{$\pm$ 0.11}}} & 53.78 {\tiny\scalebox{0.7}{\textcolor{gray}{$\pm$ 7.03}}} & 1.78 {\tiny\scalebox{0.7}{\textcolor{gray}{$\pm$ 0.11}}} & 0.52 {\tiny\scalebox{0.7}{\textcolor{gray}{$\pm$ 0.06}}} & 0.03 {\tiny\scalebox{0.7}{\textcolor{gray}{$\pm$ 0.02}}} & 0.00 {\tiny\scalebox{0.7}{\textcolor{gray}{$\pm$ 0.00}}} & 0.00 {\tiny\scalebox{0.7}{\textcolor{gray}{$\pm$ 0.00}}} & 0.00 {\tiny\scalebox{0.7}{\textcolor{gray}{$\pm$ 0.00}}} & 0.06 {\tiny\scalebox{0.7}{\textcolor{gray}{$\pm$ 0.03}}} & 0.10 & 1,870,996 \\
         & mixed & 1.02 {\tiny\scalebox{0.7}{\textcolor{gray}{$\pm$ 0.10}}} & 62.09 {\tiny\scalebox{0.7}{\textcolor{gray}{$\pm$ 6.90}}} & 1.78 {\tiny\scalebox{0.7}{\textcolor{gray}{$\pm$ 0.10}}} & 0.62 {\tiny\scalebox{0.7}{\textcolor{gray}{$\pm$ 0.06}}} & 0.02 {\tiny\scalebox{0.7}{\textcolor{gray}{$\pm$ 0.02}}} & 0.00 {\tiny\scalebox{0.7}{\textcolor{gray}{$\pm$ 0.00}}} & 0.00 {\tiny\scalebox{0.7}{\textcolor{gray}{$\pm$ 0.00}}} & 0.00 {\tiny\scalebox{0.7}{\textcolor{gray}{$\pm$ 0.00}}} & 0.08 {\tiny\scalebox{0.7}{\textcolor{gray}{$\pm$ 0.03}}} & 0.15 & 2,842,893 \\
         & primitives & 0.10 {\tiny\scalebox{0.7}{\textcolor{gray}{$\pm$ 0.05}}} & 15.94 {\tiny\scalebox{0.7}{\textcolor{gray}{$\pm$ 2.83}}} & 1.06 {\tiny\scalebox{0.7}{\textcolor{gray}{$\pm$ 0.03}}} & 0.06 {\tiny\scalebox{0.7}{\textcolor{gray}{$\pm$ 0.03}}} & 0.00 {\tiny\scalebox{0.7}{\textcolor{gray}{$\pm$ 0.00}}} & 0.00 {\tiny\scalebox{0.7}{\textcolor{gray}{$\pm$ 0.00}}} & 0.00 {\tiny\scalebox{0.7}{\textcolor{gray}{$\pm$ 0.00}}} & 0.00 {\tiny\scalebox{0.7}{\textcolor{gray}{$\pm$ 0.00}}} & 0.02 {\tiny\scalebox{0.7}{\textcolor{gray}{$\pm$ 0.02}}} & 0.39 & 7,495,379 \\
        \midrule
        Llama-3.1-8B-Instruct & skills & 1.59 {\tiny\scalebox{0.7}{\textcolor{gray}{$\pm$ 0.10}}} & 178.39 {\tiny\scalebox{0.7}{\textcolor{gray}{$\pm$ 15.52}}} & 2.48 {\tiny\scalebox{0.7}{\textcolor{gray}{$\pm$ 0.15}}} & 0.81 {\tiny\scalebox{0.7}{\textcolor{gray}{$\pm$ 0.05}}} & 0.11 {\tiny\scalebox{0.7}{\textcolor{gray}{$\pm$ 0.04}}} & 0.03 {\tiny\scalebox{0.7}{\textcolor{gray}{$\pm$ 0.02}}} & 0.00 {\tiny\scalebox{0.7}{\textcolor{gray}{$\pm$ 0.00}}} & 0.00 {\tiny\scalebox{0.7}{\textcolor{gray}{$\pm$ 0.00}}} & 0.17 {\tiny\scalebox{0.7}{\textcolor{gray}{$\pm$ 0.05}}} & 0.02 & 858,096 \\
         & mixed & 0.90 {\tiny\scalebox{0.7}{\textcolor{gray}{$\pm$ 0.11}}} & 78.62 {\tiny\scalebox{0.7}{\textcolor{gray}{$\pm$ 7.98}}} & 1.48 {\tiny\scalebox{0.7}{\textcolor{gray}{$\pm$ 0.08}}} & 0.42 {\tiny\scalebox{0.7}{\textcolor{gray}{$\pm$ 0.06}}} & 0.00 {\tiny\scalebox{0.7}{\textcolor{gray}{$\pm$ 0.00}}} & 0.00 {\tiny\scalebox{0.7}{\textcolor{gray}{$\pm$ 0.00}}} & 0.00 {\tiny\scalebox{0.7}{\textcolor{gray}{$\pm$ 0.00}}} & 0.00 {\tiny\scalebox{0.7}{\textcolor{gray}{$\pm$ 0.00}}} & 0.06 {\tiny\scalebox{0.7}{\textcolor{gray}{$\pm$ 0.03}}} & 0.02 & 1,210,851 \\
         & primitives & 0.12 {\tiny\scalebox{0.7}{\textcolor{gray}{$\pm$ 0.05}}} & 21.98 {\tiny\scalebox{0.7}{\textcolor{gray}{$\pm$ 2.91}}} & 1.08 {\tiny\scalebox{0.7}{\textcolor{gray}{$\pm$ 0.03}}} & 0.08 {\tiny\scalebox{0.7}{\textcolor{gray}{$\pm$ 0.03}}} & 0.00 {\tiny\scalebox{0.7}{\textcolor{gray}{$\pm$ 0.00}}} & 0.00 {\tiny\scalebox{0.7}{\textcolor{gray}{$\pm$ 0.00}}} & 0.00 {\tiny\scalebox{0.7}{\textcolor{gray}{$\pm$ 0.00}}} & 0.00 {\tiny\scalebox{0.7}{\textcolor{gray}{$\pm$ 0.00}}} & 0.00 {\tiny\scalebox{0.7}{\textcolor{gray}{$\pm$ 0.00}}} & 0.06 & 3,181,590 \\
        \midrule
        Llama-3.3-70B-Instruct & skills & 2.24 {\tiny\scalebox{0.7}{\textcolor{gray}{$\pm$ 0.17}}} & 332.19 {\tiny\scalebox{0.7}{\textcolor{gray}{$\pm$ 37.04}}} & 3.09 {\tiny\scalebox{0.7}{\textcolor{gray}{$\pm$ 0.20}}} & 0.83 {\tiny\scalebox{0.7}{\textcolor{gray}{$\pm$ 0.05}}} & 0.39 {\tiny\scalebox{0.7}{\textcolor{gray}{$\pm$ 0.06}}} & 0.09 {\tiny\scalebox{0.7}{\textcolor{gray}{$\pm$ 0.04}}} & 0.00 {\tiny\scalebox{0.7}{\textcolor{gray}{$\pm$ 0.00}}} & 0.02 {\tiny\scalebox{0.7}{\textcolor{gray}{$\pm$ 0.02}}} & 0.28 {\tiny\scalebox{0.7}{\textcolor{gray}{$\pm$ 0.06}}} & 0.10 & 1,020,211 \\
         & mixed & 1.79 {\tiny\scalebox{0.7}{\textcolor{gray}{$\pm$ 0.17}}} & 334.45 {\tiny\scalebox{0.7}{\textcolor{gray}{$\pm$ 40.69}}} & 2.48 {\tiny\scalebox{0.7}{\textcolor{gray}{$\pm$ 0.18}}} & 0.66 {\tiny\scalebox{0.7}{\textcolor{gray}{$\pm$ 0.06}}} & 0.28 {\tiny\scalebox{0.7}{\textcolor{gray}{$\pm$ 0.06}}} & 0.03 {\tiny\scalebox{0.7}{\textcolor{gray}{$\pm$ 0.02}}} & 0.00 {\tiny\scalebox{0.7}{\textcolor{gray}{$\pm$ 0.00}}} & 0.00 {\tiny\scalebox{0.7}{\textcolor{gray}{$\pm$ 0.00}}} & 0.27 {\tiny\scalebox{0.7}{\textcolor{gray}{$\pm$ 0.06}}} & 0.69 & 6,842,115 \\
         & primitives & 0.76 {\tiny\scalebox{0.7}{\textcolor{gray}{$\pm$ 0.12}}} & 79.84 {\tiny\scalebox{0.7}{\textcolor{gray}{$\pm$ 10.18}}} & 1.05 {\tiny\scalebox{0.7}{\textcolor{gray}{$\pm$ 0.03}}} & 0.03 {\tiny\scalebox{0.7}{\textcolor{gray}{$\pm$ 0.02}}} & 0.00 {\tiny\scalebox{0.7}{\textcolor{gray}{$\pm$ 0.00}}} & 0.00 {\tiny\scalebox{0.7}{\textcolor{gray}{$\pm$ 0.00}}} & 0.00 {\tiny\scalebox{0.7}{\textcolor{gray}{$\pm$ 0.00}}} & 0.00 {\tiny\scalebox{0.7}{\textcolor{gray}{$\pm$ 0.00}}} & 0.00 {\tiny\scalebox{0.7}{\textcolor{gray}{$\pm$ 0.00}}} & 0.68 & 6,700,777 \\
        \midrule
        Gemma3-4B & skills & 0.23 {\tiny\scalebox{0.7}{\textcolor{gray}{$\pm$ 0.07}}} & 10.38 {\tiny\scalebox{0.7}{\textcolor{gray}{$\pm$ 3.68}}} & 1.19 {\tiny\scalebox{0.7}{\textcolor{gray}{$\pm$ 0.07}}} & 0.14 {\tiny\scalebox{0.7}{\textcolor{gray}{$\pm$ 0.04}}} & 0.02 {\tiny\scalebox{0.7}{\textcolor{gray}{$\pm$ 0.02}}} & 0.00 {\tiny\scalebox{0.7}{\textcolor{gray}{$\pm$ 0.00}}} & 0.00 {\tiny\scalebox{0.7}{\textcolor{gray}{$\pm$ 0.00}}} & 0.00 {\tiny\scalebox{0.7}{\textcolor{gray}{$\pm$ 0.00}}} & 0.00 {\tiny\scalebox{0.7}{\textcolor{gray}{$\pm$ 0.00}}} & 0.03 & 629,528 \\
         & mixed & 0.10 {\tiny\scalebox{0.7}{\textcolor{gray}{$\pm$ 0.05}}} & 4.50 {\tiny\scalebox{0.7}{\textcolor{gray}{$\pm$ 1.64}}} & 1.06 {\tiny\scalebox{0.7}{\textcolor{gray}{$\pm$ 0.03}}} & 0.06 {\tiny\scalebox{0.7}{\textcolor{gray}{$\pm$ 0.03}}} & 0.00 {\tiny\scalebox{0.7}{\textcolor{gray}{$\pm$ 0.00}}} & 0.00 {\tiny\scalebox{0.7}{\textcolor{gray}{$\pm$ 0.00}}} & 0.00 {\tiny\scalebox{0.7}{\textcolor{gray}{$\pm$ 0.00}}} & 0.00 {\tiny\scalebox{0.7}{\textcolor{gray}{$\pm$ 0.00}}} & 0.00 {\tiny\scalebox{0.7}{\textcolor{gray}{$\pm$ 0.00}}} & 0.11 & 2,747,439 \\
         & primitives & 0.00 {\tiny\scalebox{0.7}{\textcolor{gray}{$\pm$ 0.00}}} & 0.41 {\tiny\scalebox{0.7}{\textcolor{gray}{$\pm$ 0.27}}} & 1.00 {\tiny\scalebox{0.7}{\textcolor{gray}{$\pm$ 0.00}}} & 0.00 {\tiny\scalebox{0.7}{\textcolor{gray}{$\pm$ 0.00}}} & 0.00 {\tiny\scalebox{0.7}{\textcolor{gray}{$\pm$ 0.00}}} & 0.00 {\tiny\scalebox{0.7}{\textcolor{gray}{$\pm$ 0.00}}} & 0.00 {\tiny\scalebox{0.7}{\textcolor{gray}{$\pm$ 0.00}}} & 0.00 {\tiny\scalebox{0.7}{\textcolor{gray}{$\pm$ 0.00}}} & 0.00 {\tiny\scalebox{0.7}{\textcolor{gray}{$\pm$ 0.00}}} & 0.08 & 1,872,859 \\
        \midrule
        Gemma3-12B & skills & 0.89 {\tiny\scalebox{0.7}{\textcolor{gray}{$\pm$ 0.12}}} & 63.27 {\tiny\scalebox{0.7}{\textcolor{gray}{$\pm$ 10.98}}} & 1.77 {\tiny\scalebox{0.7}{\textcolor{gray}{$\pm$ 0.13}}} & 0.47 {\tiny\scalebox{0.7}{\textcolor{gray}{$\pm$ 0.06}}} & 0.09 {\tiny\scalebox{0.7}{\textcolor{gray}{$\pm$ 0.04}}} & 0.00 {\tiny\scalebox{0.7}{\textcolor{gray}{$\pm$ 0.00}}} & 0.00 {\tiny\scalebox{0.7}{\textcolor{gray}{$\pm$ 0.00}}} & 0.00 {\tiny\scalebox{0.7}{\textcolor{gray}{$\pm$ 0.00}}} & 0.06 {\tiny\scalebox{0.7}{\textcolor{gray}{$\pm$ 0.03}}} & 0.03 & 781,424 \\
         & mixed & 0.13 {\tiny\scalebox{0.7}{\textcolor{gray}{$\pm$ 0.06}}} & 7.53 {\tiny\scalebox{0.7}{\textcolor{gray}{$\pm$ 2.67}}} & 1.11 {\tiny\scalebox{0.7}{\textcolor{gray}{$\pm$ 0.05}}} & 0.08 {\tiny\scalebox{0.7}{\textcolor{gray}{$\pm$ 0.03}}} & 0.00 {\tiny\scalebox{0.7}{\textcolor{gray}{$\pm$ 0.00}}} & 0.00 {\tiny\scalebox{0.7}{\textcolor{gray}{$\pm$ 0.00}}} & 0.00 {\tiny\scalebox{0.7}{\textcolor{gray}{$\pm$ 0.00}}} & 0.00 {\tiny\scalebox{0.7}{\textcolor{gray}{$\pm$ 0.00}}} & 0.00 {\tiny\scalebox{0.7}{\textcolor{gray}{$\pm$ 0.00}}} & 0.02 & 461,737 \\
         & primitives & 0.00 {\tiny\scalebox{0.7}{\textcolor{gray}{$\pm$ 0.00}}} & 0.58 {\tiny\scalebox{0.7}{\textcolor{gray}{$\pm$ 0.28}}} & 1.00 {\tiny\scalebox{0.7}{\textcolor{gray}{$\pm$ 0.00}}} & 0.00 {\tiny\scalebox{0.7}{\textcolor{gray}{$\pm$ 0.00}}} & 0.00 {\tiny\scalebox{0.7}{\textcolor{gray}{$\pm$ 0.00}}} & 0.00 {\tiny\scalebox{0.7}{\textcolor{gray}{$\pm$ 0.00}}} & 0.00 {\tiny\scalebox{0.7}{\textcolor{gray}{$\pm$ 0.00}}} & 0.00 {\tiny\scalebox{0.7}{\textcolor{gray}{$\pm$ 0.00}}} & 0.00 {\tiny\scalebox{0.7}{\textcolor{gray}{$\pm$ 0.00}}} & 0.01 & 138,330 \\
        \midrule
        Gemma3-27B & skills & 1.26 {\tiny\scalebox{0.7}{\textcolor{gray}{$\pm$ 0.14}}} & 140.17 {\tiny\scalebox{0.7}{\textcolor{gray}{$\pm$ 21.58}}} & 2.09 {\tiny\scalebox{0.7}{\textcolor{gray}{$\pm$ 0.15}}} & 0.62 {\tiny\scalebox{0.7}{\textcolor{gray}{$\pm$ 0.06}}} & 0.11 {\tiny\scalebox{0.7}{\textcolor{gray}{$\pm$ 0.04}}} & 0.02 {\tiny\scalebox{0.7}{\textcolor{gray}{$\pm$ 0.02}}} & 0.00 {\tiny\scalebox{0.7}{\textcolor{gray}{$\pm$ 0.00}}} & 0.00 {\tiny\scalebox{0.7}{\textcolor{gray}{$\pm$ 0.00}}} & 0.19 {\tiny\scalebox{0.7}{\textcolor{gray}{$\pm$ 0.05}}} & 0.13 & 1,620,361 \\
         & mixed & 1.29 {\tiny\scalebox{0.7}{\textcolor{gray}{$\pm$ 0.15}}} & 131.70 {\tiny\scalebox{0.7}{\textcolor{gray}{$\pm$ 18.91}}} & 2.06 {\tiny\scalebox{0.7}{\textcolor{gray}{$\pm$ 0.16}}} & 0.59 {\tiny\scalebox{0.7}{\textcolor{gray}{$\pm$ 0.06}}} & 0.12 {\tiny\scalebox{0.7}{\textcolor{gray}{$\pm$ 0.04}}} & 0.02 {\tiny\scalebox{0.7}{\textcolor{gray}{$\pm$ 0.02}}} & 0.02 {\tiny\scalebox{0.7}{\textcolor{gray}{$\pm$ 0.02}}} & 0.02 {\tiny\scalebox{0.7}{\textcolor{gray}{$\pm$ 0.02}}} & 0.14 {\tiny\scalebox{0.7}{\textcolor{gray}{$\pm$ 0.04}}} & 0.33 & 4,147,507 \\
         & primitives & 0.44 {\tiny\scalebox{0.7}{\textcolor{gray}{$\pm$ 0.10}}} & 55.09 {\tiny\scalebox{0.7}{\textcolor{gray}{$\pm$ 8.64}}} & 1.23 {\tiny\scalebox{0.7}{\textcolor{gray}{$\pm$ 0.07}}} & 0.19 {\tiny\scalebox{0.7}{\textcolor{gray}{$\pm$ 0.05}}} & 0.02 {\tiny\scalebox{0.7}{\textcolor{gray}{$\pm$ 0.02}}} & 0.00 {\tiny\scalebox{0.7}{\textcolor{gray}{$\pm$ 0.00}}} & 0.00 {\tiny\scalebox{0.7}{\textcolor{gray}{$\pm$ 0.00}}} & 0.00 {\tiny\scalebox{0.7}{\textcolor{gray}{$\pm$ 0.00}}} & 0.00 {\tiny\scalebox{0.7}{\textcolor{gray}{$\pm$ 0.00}}} & 0.34 & 4,253,730 \\
        \midrule
        Gemma-4-E2B & skills & 1.05 {\tiny\scalebox{0.7}{\textcolor{gray}{$\pm$ 0.11}}} & 54.17 {\tiny\scalebox{0.7}{\textcolor{gray}{$\pm$ 6.39}}} & 1.86 {\tiny\scalebox{0.7}{\textcolor{gray}{$\pm$ 0.11}}} & 0.62 {\tiny\scalebox{0.7}{\textcolor{gray}{$\pm$ 0.06}}} & 0.05 {\tiny\scalebox{0.7}{\textcolor{gray}{$\pm$ 0.03}}} & 0.00 {\tiny\scalebox{0.7}{\textcolor{gray}{$\pm$ 0.00}}} & 0.00 {\tiny\scalebox{0.7}{\textcolor{gray}{$\pm$ 0.00}}} & 0.00 {\tiny\scalebox{0.7}{\textcolor{gray}{$\pm$ 0.00}}} & 0.09 {\tiny\scalebox{0.7}{\textcolor{gray}{$\pm$ 0.04}}} & 0.16 & 4,758,126 \\
         & mixed & 1.37 {\tiny\scalebox{0.7}{\textcolor{gray}{$\pm$ 0.11}}} & 104.25 {\tiny\scalebox{0.7}{\textcolor{gray}{$\pm$ 8.97}}} & 2.36 {\tiny\scalebox{0.7}{\textcolor{gray}{$\pm$ 0.14}}} & 0.75 {\tiny\scalebox{0.7}{\textcolor{gray}{$\pm$ 0.05}}} & 0.16 {\tiny\scalebox{0.7}{\textcolor{gray}{$\pm$ 0.05}}} & 0.00 {\tiny\scalebox{0.7}{\textcolor{gray}{$\pm$ 0.00}}} & 0.00 {\tiny\scalebox{0.7}{\textcolor{gray}{$\pm$ 0.00}}} & 0.02 {\tiny\scalebox{0.7}{\textcolor{gray}{$\pm$ 0.02}}} & 0.16 {\tiny\scalebox{0.7}{\textcolor{gray}{$\pm$ 0.05}}} & 0.06 & 1,893,335 \\
         & primitives & 0.08 {\tiny\scalebox{0.7}{\textcolor{gray}{$\pm$ 0.04}}} & 10.08 {\tiny\scalebox{0.7}{\textcolor{gray}{$\pm$ 3.10}}} & 1.05 {\tiny\scalebox{0.7}{\textcolor{gray}{$\pm$ 0.03}}} & 0.05 {\tiny\scalebox{0.7}{\textcolor{gray}{$\pm$ 0.03}}} & 0.00 {\tiny\scalebox{0.7}{\textcolor{gray}{$\pm$ 0.00}}} & 0.00 {\tiny\scalebox{0.7}{\textcolor{gray}{$\pm$ 0.00}}} & 0.00 {\tiny\scalebox{0.7}{\textcolor{gray}{$\pm$ 0.00}}} & 0.00 {\tiny\scalebox{0.7}{\textcolor{gray}{$\pm$ 0.00}}} & 0.00 {\tiny\scalebox{0.7}{\textcolor{gray}{$\pm$ 0.00}}} & 0.05 & 1,475,575 \\
        \midrule
        Gemma-4-E4B & skills & 1.24 {\tiny\scalebox{0.7}{\textcolor{gray}{$\pm$ 0.12}}} & 100.58 {\tiny\scalebox{0.7}{\textcolor{gray}{$\pm$ 12.88}}} & 2.06 {\tiny\scalebox{0.7}{\textcolor{gray}{$\pm$ 0.14}}} & 0.64 {\tiny\scalebox{0.7}{\textcolor{gray}{$\pm$ 0.06}}} & 0.16 {\tiny\scalebox{0.7}{\textcolor{gray}{$\pm$ 0.05}}} & 0.00 {\tiny\scalebox{0.7}{\textcolor{gray}{$\pm$ 0.00}}} & 0.00 {\tiny\scalebox{0.7}{\textcolor{gray}{$\pm$ 0.00}}} & 0.02 {\tiny\scalebox{0.7}{\textcolor{gray}{$\pm$ 0.02}}} & 0.09 {\tiny\scalebox{0.7}{\textcolor{gray}{$\pm$ 0.04}}} & 0.25 & 3,753,052 \\
         & mixed & 1.78 {\tiny\scalebox{0.7}{\textcolor{gray}{$\pm$ 0.15}}} & 216.17 {\tiny\scalebox{0.7}{\textcolor{gray}{$\pm$ 22.76}}} & 2.72 {\tiny\scalebox{0.7}{\textcolor{gray}{$\pm$ 0.21}}} & 0.69 {\tiny\scalebox{0.7}{\textcolor{gray}{$\pm$ 0.06}}} & 0.28 {\tiny\scalebox{0.7}{\textcolor{gray}{$\pm$ 0.06}}} & 0.06 {\tiny\scalebox{0.7}{\textcolor{gray}{$\pm$ 0.03}}} & 0.00 {\tiny\scalebox{0.7}{\textcolor{gray}{$\pm$ 0.00}}} & 0.02 {\tiny\scalebox{0.7}{\textcolor{gray}{$\pm$ 0.02}}} & 0.16 {\tiny\scalebox{0.7}{\textcolor{gray}{$\pm$ 0.05}}} & 0.31 & 4,746,369 \\
         & primitives & 0.62 {\tiny\scalebox{0.7}{\textcolor{gray}{$\pm$ 0.11}}} & 75.73 {\tiny\scalebox{0.7}{\textcolor{gray}{$\pm$ 11.06}}} & 1.23 {\tiny\scalebox{0.7}{\textcolor{gray}{$\pm$ 0.07}}} & 0.19 {\tiny\scalebox{0.7}{\textcolor{gray}{$\pm$ 0.05}}} & 0.02 {\tiny\scalebox{0.7}{\textcolor{gray}{$\pm$ 0.02}}} & 0.00 {\tiny\scalebox{0.7}{\textcolor{gray}{$\pm$ 0.00}}} & 0.00 {\tiny\scalebox{0.7}{\textcolor{gray}{$\pm$ 0.00}}} & 0.00 {\tiny\scalebox{0.7}{\textcolor{gray}{$\pm$ 0.00}}} & 0.00 {\tiny\scalebox{0.7}{\textcolor{gray}{$\pm$ 0.00}}} & 0.50 & 7,431,707 \\
        \midrule
        Gemma-4-31B & skills & 3.21 {\tiny\scalebox{0.7}{\textcolor{gray}{$\pm$ 0.30}}} & 568.67 {\tiny\scalebox{0.7}{\textcolor{gray}{$\pm$ 70.24}}} & 4.06 {\tiny\scalebox{0.7}{\textcolor{gray}{$\pm$ 0.29}}} & 0.91 {\tiny\scalebox{0.7}{\textcolor{gray}{$\pm$ 0.04}}} & 0.50 {\tiny\scalebox{0.7}{\textcolor{gray}{$\pm$ 0.06}}} & 0.28 {\tiny\scalebox{0.7}{\textcolor{gray}{$\pm$ 0.06}}} & 0.11 {\tiny\scalebox{0.7}{\textcolor{gray}{$\pm$ 0.04}}} & 0.09 {\tiny\scalebox{0.7}{\textcolor{gray}{$\pm$ 0.04}}} & 0.30 {\tiny\scalebox{0.7}{\textcolor{gray}{$\pm$ 0.06}}} & 0.29 & 1,792,420 \\
         & mixed & 3.16 {\tiny\scalebox{0.7}{\textcolor{gray}{$\pm$ 0.24}}} & 582.70 {\tiny\scalebox{0.7}{\textcolor{gray}{$\pm$ 66.09}}} & 3.89 {\tiny\scalebox{0.7}{\textcolor{gray}{$\pm$ 0.27}}} & 0.89 {\tiny\scalebox{0.7}{\textcolor{gray}{$\pm$ 0.04}}} & 0.50 {\tiny\scalebox{0.7}{\textcolor{gray}{$\pm$ 0.06}}} & 0.25 {\tiny\scalebox{0.7}{\textcolor{gray}{$\pm$ 0.05}}} & 0.06 {\tiny\scalebox{0.7}{\textcolor{gray}{$\pm$ 0.03}}} & 0.06 {\tiny\scalebox{0.7}{\textcolor{gray}{$\pm$ 0.03}}} & 0.34 {\tiny\scalebox{0.7}{\textcolor{gray}{$\pm$ 0.06}}} & 1.08 & 6,996,392 \\
         & primitives & 1.45 {\tiny\scalebox{0.7}{\textcolor{gray}{$\pm$ 0.12}}} & 171.11 {\tiny\scalebox{0.7}{\textcolor{gray}{$\pm$ 16.01}}} & 1.05 {\tiny\scalebox{0.7}{\textcolor{gray}{$\pm$ 0.03}}} & 0.05 {\tiny\scalebox{0.7}{\textcolor{gray}{$\pm$ 0.03}}} & 0.00 {\tiny\scalebox{0.7}{\textcolor{gray}{$\pm$ 0.00}}} & 0.00 {\tiny\scalebox{0.7}{\textcolor{gray}{$\pm$ 0.00}}} & 0.00 {\tiny\scalebox{0.7}{\textcolor{gray}{$\pm$ 0.00}}} & 0.00 {\tiny\scalebox{0.7}{\textcolor{gray}{$\pm$ 0.00}}} & 0.00 {\tiny\scalebox{0.7}{\textcolor{gray}{$\pm$ 0.00}}} & 2.28 & 12,738,072 \\
        \midrule
        Qwen-3.5-4B & skills & 1.80 {\tiny\scalebox{0.7}{\textcolor{gray}{$\pm$ 0.13}}} & 324.48 {\tiny\scalebox{0.7}{\textcolor{gray}{$\pm$ 33.11}}} & 2.61 {\tiny\scalebox{0.7}{\textcolor{gray}{$\pm$ 0.17}}} & 0.78 {\tiny\scalebox{0.7}{\textcolor{gray}{$\pm$ 0.05}}} & 0.20 {\tiny\scalebox{0.7}{\textcolor{gray}{$\pm$ 0.05}}} & 0.05 {\tiny\scalebox{0.7}{\textcolor{gray}{$\pm$ 0.03}}} & 0.00 {\tiny\scalebox{0.7}{\textcolor{gray}{$\pm$ 0.00}}} & 0.02 {\tiny\scalebox{0.7}{\textcolor{gray}{$\pm$ 0.02}}} & 0.27 {\tiny\scalebox{0.7}{\textcolor{gray}{$\pm$ 0.06}}} & 0.07 & 1,571,941 \\
         & mixed & 1.77 {\tiny\scalebox{0.7}{\textcolor{gray}{$\pm$ 0.10}}} & 237.41 {\tiny\scalebox{0.7}{\textcolor{gray}{$\pm$ 24.97}}} & 2.56 {\tiny\scalebox{0.7}{\textcolor{gray}{$\pm$ 0.15}}} & 0.83 {\tiny\scalebox{0.7}{\textcolor{gray}{$\pm$ 0.05}}} & 0.22 {\tiny\scalebox{0.7}{\textcolor{gray}{$\pm$ 0.05}}} & 0.02 {\tiny\scalebox{0.7}{\textcolor{gray}{$\pm$ 0.02}}} & 0.00 {\tiny\scalebox{0.7}{\textcolor{gray}{$\pm$ 0.00}}} & 0.00 {\tiny\scalebox{0.7}{\textcolor{gray}{$\pm$ 0.00}}} & 0.28 {\tiny\scalebox{0.7}{\textcolor{gray}{$\pm$ 0.06}}} & 0.08 & 2,074,919 \\
         & primitives & 0.51 {\tiny\scalebox{0.7}{\textcolor{gray}{$\pm$ 0.11}}} & 59.91 {\tiny\scalebox{0.7}{\textcolor{gray}{$\pm$ 11.67}}} & 1.09 {\tiny\scalebox{0.7}{\textcolor{gray}{$\pm$ 0.04}}} & 0.08 {\tiny\scalebox{0.7}{\textcolor{gray}{$\pm$ 0.03}}} & 0.00 {\tiny\scalebox{0.7}{\textcolor{gray}{$\pm$ 0.00}}} & 0.00 {\tiny\scalebox{0.7}{\textcolor{gray}{$\pm$ 0.00}}} & 0.00 {\tiny\scalebox{0.7}{\textcolor{gray}{$\pm$ 0.00}}} & 0.00 {\tiny\scalebox{0.7}{\textcolor{gray}{$\pm$ 0.00}}} & 0.00 {\tiny\scalebox{0.7}{\textcolor{gray}{$\pm$ 0.00}}} & 0.27 & 5,280,187 \\
        \midrule
        Qwen-3.5-9B & skills & 2.37 {\tiny\scalebox{0.7}{\textcolor{gray}{$\pm$ 0.14}}} & 373.59 {\tiny\scalebox{0.7}{\textcolor{gray}{$\pm$ 29.49}}} & 3.34 {\tiny\scalebox{0.7}{\textcolor{gray}{$\pm$ 0.21}}} & 0.89 {\tiny\scalebox{0.7}{\textcolor{gray}{$\pm$ 0.04}}} & 0.36 {\tiny\scalebox{0.7}{\textcolor{gray}{$\pm$ 0.06}}} & 0.11 {\tiny\scalebox{0.7}{\textcolor{gray}{$\pm$ 0.04}}} & 0.03 {\tiny\scalebox{0.7}{\textcolor{gray}{$\pm$ 0.02}}} & 0.02 {\tiny\scalebox{0.7}{\textcolor{gray}{$\pm$ 0.02}}} & 0.33 {\tiny\scalebox{0.7}{\textcolor{gray}{$\pm$ 0.06}}} & 0.14 & 1,377,070 \\
         & mixed & 2.25 {\tiny\scalebox{0.7}{\textcolor{gray}{$\pm$ 0.17}}} & 373.58 {\tiny\scalebox{0.7}{\textcolor{gray}{$\pm$ 33.40}}} & 3.33 {\tiny\scalebox{0.7}{\textcolor{gray}{$\pm$ 0.20}}} & 0.89 {\tiny\scalebox{0.7}{\textcolor{gray}{$\pm$ 0.04}}} & 0.42 {\tiny\scalebox{0.7}{\textcolor{gray}{$\pm$ 0.06}}} & 0.11 {\tiny\scalebox{0.7}{\textcolor{gray}{$\pm$ 0.04}}} & 0.02 {\tiny\scalebox{0.7}{\textcolor{gray}{$\pm$ 0.02}}} & 0.03 {\tiny\scalebox{0.7}{\textcolor{gray}{$\pm$ 0.02}}} & 0.28 {\tiny\scalebox{0.7}{\textcolor{gray}{$\pm$ 0.06}}} & 0.42 & 4,076,366 \\
         & primitives & 1.12 {\tiny\scalebox{0.7}{\textcolor{gray}{$\pm$ 0.13}}} & 121.66 {\tiny\scalebox{0.7}{\textcolor{gray}{$\pm$ 13.13}}} & 1.41 {\tiny\scalebox{0.7}{\textcolor{gray}{$\pm$ 0.12}}} & 0.23 {\tiny\scalebox{0.7}{\textcolor{gray}{$\pm$ 0.05}}} & 0.05 {\tiny\scalebox{0.7}{\textcolor{gray}{$\pm$ 0.03}}} & 0.02 {\tiny\scalebox{0.7}{\textcolor{gray}{$\pm$ 0.02}}} & 0.00 {\tiny\scalebox{0.7}{\textcolor{gray}{$\pm$ 0.00}}} & 0.00 {\tiny\scalebox{0.7}{\textcolor{gray}{$\pm$ 0.00}}} & 0.03 {\tiny\scalebox{0.7}{\textcolor{gray}{$\pm$ 0.02}}} & 1.09 & 9,946,628 \\
        \midrule
        Qwen-3.5-27B & skills & 3.41 {\tiny\scalebox{0.7}{\textcolor{gray}{$\pm$ 0.43}}} & 529.08 {\tiny\scalebox{0.7}{\textcolor{gray}{$\pm$ 56.37}}} & 4.28 {\tiny\scalebox{0.7}{\textcolor{gray}{$\pm$ 0.29}}} & 0.94 {\tiny\scalebox{0.7}{\textcolor{gray}{$\pm$ 0.03}}} & 0.53 {\tiny\scalebox{0.7}{\textcolor{gray}{$\pm$ 0.06}}} & 0.30 {\tiny\scalebox{0.7}{\textcolor{gray}{$\pm$ 0.06}}} & 0.06 {\tiny\scalebox{0.7}{\textcolor{gray}{$\pm$ 0.03}}} & 0.06 {\tiny\scalebox{0.7}{\textcolor{gray}{$\pm$ 0.03}}} & 0.38 {\tiny\scalebox{0.7}{\textcolor{gray}{$\pm$ 0.06}}} & 0.46 & 1,163,767 \\
         & mixed & 3.65 {\tiny\scalebox{0.7}{\textcolor{gray}{$\pm$ 0.40}}} & 523.05 {\tiny\scalebox{0.7}{\textcolor{gray}{$\pm$ 46.49}}} & 4.52 {\tiny\scalebox{0.7}{\textcolor{gray}{$\pm$ 0.29}}} & 0.98 {\tiny\scalebox{0.7}{\textcolor{gray}{$\pm$ 0.02}}} & 0.61 {\tiny\scalebox{0.7}{\textcolor{gray}{$\pm$ 0.06}}} & 0.30 {\tiny\scalebox{0.7}{\textcolor{gray}{$\pm$ 0.06}}} & 0.11 {\tiny\scalebox{0.7}{\textcolor{gray}{$\pm$ 0.04}}} & 0.05 {\tiny\scalebox{0.7}{\textcolor{gray}{$\pm$ 0.03}}} & 0.38 {\tiny\scalebox{0.7}{\textcolor{gray}{$\pm$ 0.06}}} & 0.68 & 2,129,585 \\
         & primitives & 1.31 {\tiny\scalebox{0.7}{\textcolor{gray}{$\pm$ 0.12}}} & 177.41 {\tiny\scalebox{0.7}{\textcolor{gray}{$\pm$ 20.04}}} & 1.59 {\tiny\scalebox{0.7}{\textcolor{gray}{$\pm$ 0.12}}} & 0.36 {\tiny\scalebox{0.7}{\textcolor{gray}{$\pm$ 0.06}}} & 0.06 {\tiny\scalebox{0.7}{\textcolor{gray}{$\pm$ 0.03}}} & 0.00 {\tiny\scalebox{0.7}{\textcolor{gray}{$\pm$ 0.00}}} & 0.00 {\tiny\scalebox{0.7}{\textcolor{gray}{$\pm$ 0.00}}} & 0.00 {\tiny\scalebox{0.7}{\textcolor{gray}{$\pm$ 0.00}}} & 0.05 {\tiny\scalebox{0.7}{\textcolor{gray}{$\pm$ 0.03}}} & 4.09 & 9,302,815 \\
\end{longtable}
\endgroup

%% file: tables/nethack_skills_vs_primitives_v5.tex
\begin{table}[t]
    \centering
    \small
    \setlength{\tabcolsep}{3pt}
    \caption{Average zero-shot NetHack performance and resource use across 14 models with both primitive and skill runs for every metric. Gray $\pm$ terms are standard errors across paired model-level means. Ratios and percent changes compare the skills average against the primitives average; lower cost and token values indicate reduced resource use. Ratios with a zero primitive baseline are undefined and shown as --.}
    \label{tab:nethack-skills-vs-primitives}
    \begin{tabular}{lrrrr}
        \toprule
        Metric & Primitives avg & Skills avg & Skills / primitives & \% change \\
        \midrule
        Progression (\%) & 0.690 {\tiny\textcolor{gray}{$\pm$ 0.164}} & 1.978 {\tiny\textcolor{gray}{$\pm$ 0.325}} & 2.866$\times$ & +187\% \\
        Score & 83.948 {\tiny\textcolor{gray}{$\pm$ 20.157}} & 318.268 {\tiny\textcolor{gray}{$\pm$ 89.226}} & 3.791$\times$ & +279\% \\
        Max DLvl & 1.188 {\tiny\textcolor{gray}{$\pm$ 0.052}} & 2.880 {\tiny\textcolor{gray}{$\pm$ 0.340}} & 2.425$\times$ & +143\% \\
        L2 reach & 0.133 {\tiny\textcolor{gray}{$\pm$ 0.032}} & 0.719 {\tiny\textcolor{gray}{$\pm$ 0.062}} & 5.411$\times$ & +441\% \\
        L4 reach & 0.013 {\tiny\textcolor{gray}{$\pm$ 0.005}} & 0.286 {\tiny\textcolor{gray}{$\pm$ 0.068}} & 21.320$\times$ & +2032\% \\
        L6 reach & 0.001 {\tiny\textcolor{gray}{$\pm$ 0.001}} & 0.117 {\tiny\textcolor{gray}{$\pm$ 0.043}} & 104.540$\times$ & +10354\% \\
        L8 reach & 0.000 {\tiny\textcolor{gray}{$\pm$ 0.000}} & 0.036 {\tiny\textcolor{gray}{$\pm$ 0.019}} & -- & -- \\
        L10 reach & 0.000 {\tiny\textcolor{gray}{$\pm$ 0.000}} & 0.004 {\tiny\textcolor{gray}{$\pm$ 0.003}} & -- & -- \\
        Oracle reach & 0.000 {\tiny\textcolor{gray}{$\pm$ 0.000}} & 0.022 {\tiny\textcolor{gray}{$\pm$ 0.008}} & -- & -- \\
        Gnomish Mines reach & 0.009 {\tiny\textcolor{gray}{$\pm$ 0.004}} & 0.224 {\tiny\textcolor{gray}{$\pm$ 0.040}} & 25.125$\times$ & +2413\% \\
        Cost / episode (\$) & 4.348 {\tiny\textcolor{gray}{$\pm$ 3.556}} & 0.590 {\tiny\textcolor{gray}{$\pm$ 0.451}} & 0.136$\times$ & -86\% \\
        Total tokens / episode & 6,720,289 {\tiny\textcolor{gray}{$\pm$ 1,133,932}} & 1,741,509 {\tiny\textcolor{gray}{$\pm$ 310,306}} & 0.259$\times$ & -74\% \\
        \bottomrule
    \end{tabular}
\end{table}

%% file: tables/qualitative_mixed_control_v1.tex
\begin{table}[htbp]
    \centering
    \caption{\textbf{Primitive fallback in two zero-shot \texttt{GPT-5} trajectories.} Numbers are controller-decision indices shown in the linked videos.}
    \label{tab:qualitative-mixed-control}
    \begingroup
    \small
    \setlength{\tabcolsep}{4pt}
    \renewcommand{\arraystretch}{1.15}
    \begin{tabular}{@{}p{0.15\linewidth}p{0.25\linewidth}p{0.32\linewidth}p{0.20\linewidth}@{}}
        \toprule
        \textbf{Example} & \textbf{Skill limitation} & \textbf{Primitive fallback} & \textbf{Return to skills} \\
        \midrule
        \raggedright \href{https://bartekcupial.github.io/abstraction-ladder/gpt5_video_showcase/01_nethack_mixed_control.mp4}{Ranged combat}
        & \raggedright 732: \texttt{fight\_ranged} reports no ranged weapon or ammunition and executes zero environment steps.
        & \raggedright 733--735: \texttt{throw}, \texttt{i}, \texttt{east} select a javelin and kill the gas spore. 736--739: \texttt{east} moves to the dropped weapons.
        & \raggedright 740: \texttt{pickup\_weapon} recovers the weapons. \tabularnewline
        \midrule
        \raggedright \href{https://bartekcupial.github.io/abstraction-ladder/gpt5_video_showcase/04_vault_drop_gold_recovery.mp4}{Vault guard}
        & \raggedright 480: \texttt{explore} executes zero environment steps. 483: the guard repeats its demand to drop the gold; no drop-gold skill is available.
        & \raggedright 484--485: \texttt{drop}, \texttt{\$} drop all 650 gold pieces. 486--487: \texttt{east} follows the corridor.
        & \raggedright 488: \texttt{explore} resumes navigation. \tabularnewline
        \bottomrule
    \end{tabular}
    \endgroup
\end{table}

%% file: sections/appendixD.tex
\clearpage

\section{Full RL results}
\label{app:rl-results}

\begin{figure}[t]
    \centering
    \includegraphics[width=\linewidth]{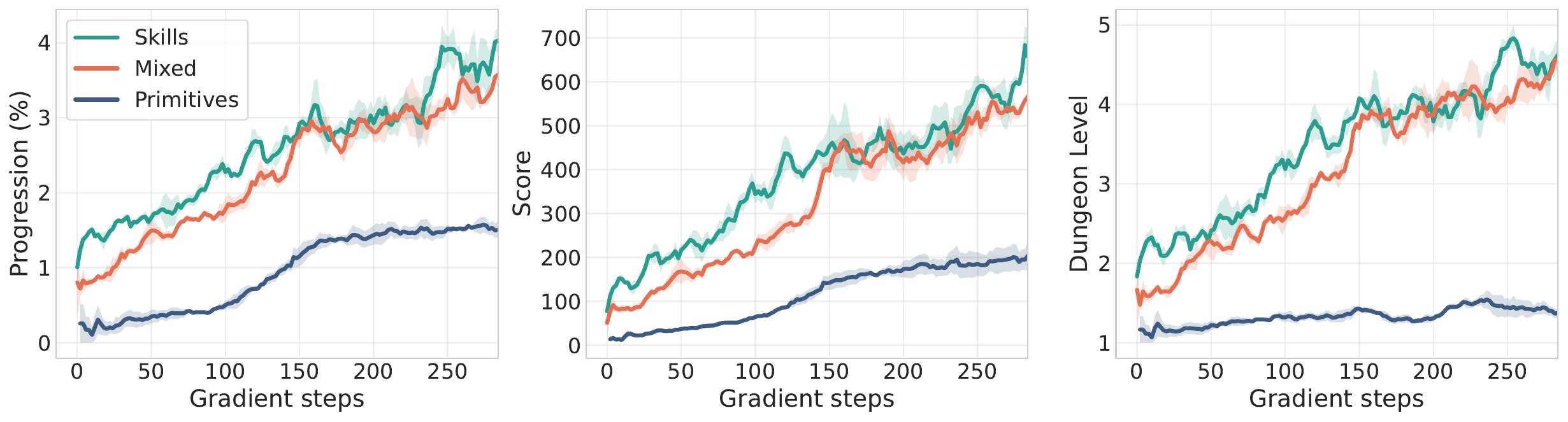}
    \caption{\textbf{Full training curves for \texttt{Llama-3.1-8B-Instruct} (gradient steps).} We plot progression (left), score (middle), and dungeon level (right) vs.\ gradient steps. All curves are averages across three seeds, and error bands indicate one standard error.}
    \label{fig:rl-progression-score-max-dlvl-vs-grad-steps-llama}
\end{figure}

\begin{figure}[t]
    \centering
    \includegraphics[width=\linewidth]{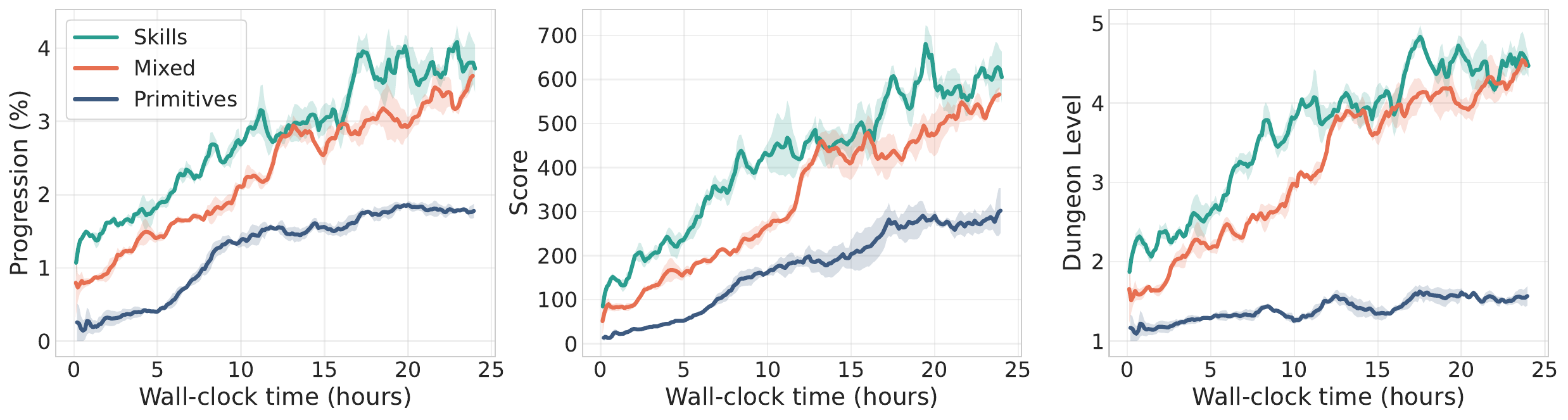}
    \caption{\textbf{Full training curves for \texttt{Llama-3.1-8B-Instruct} (wall-clock time).} We plot progression (left), score (middle), and dungeon level (right) vs. wall-clock time. All curves are averages across three seeds, and error bands indicate one standard error.}
    \label{fig:rl-progression-score-max-dlvl-vs-wall-clock-time-llama}
\end{figure}

\begin{figure}[t]
    \centering
    \includegraphics[width=\linewidth]{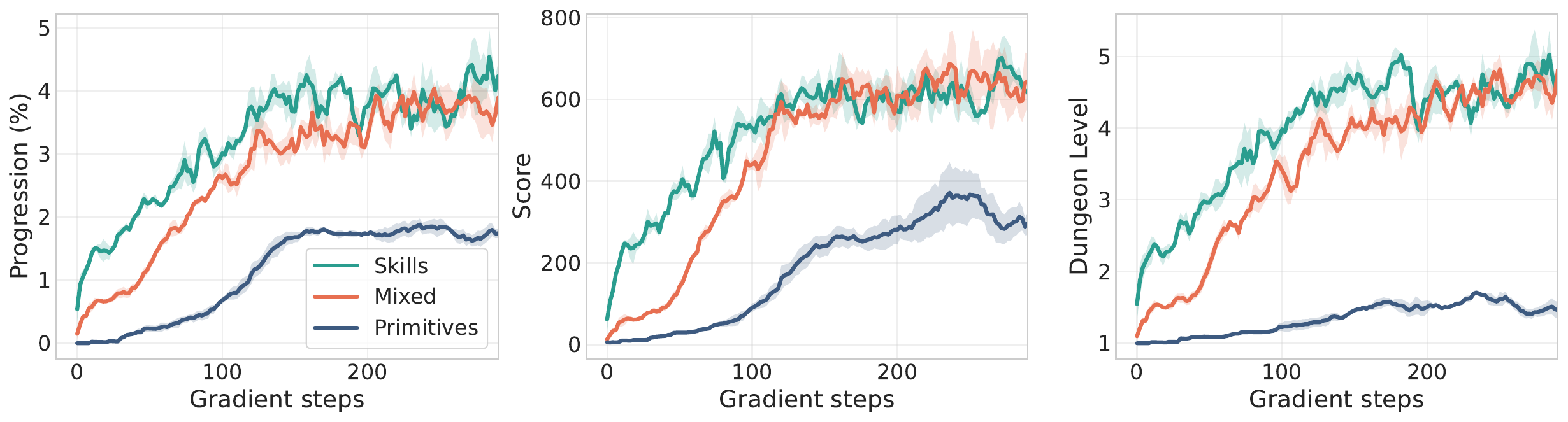}
    \caption{\textbf{Full training curves for \texttt{Qwen-3.5-4B} (gradient steps).} We plot progression (left), score (middle), and dungeon level (right) vs.\ gradient steps. All curves are averages across three seeds (except for mixed, which uses 2 seeds), and error bands indicate one standard error.}
    \label{fig:rl-progression-score-max-dlvl-vs-grad-steps-qwen}
\end{figure}

\begin{figure}[t]
    \centering
    \includegraphics[width=\linewidth]{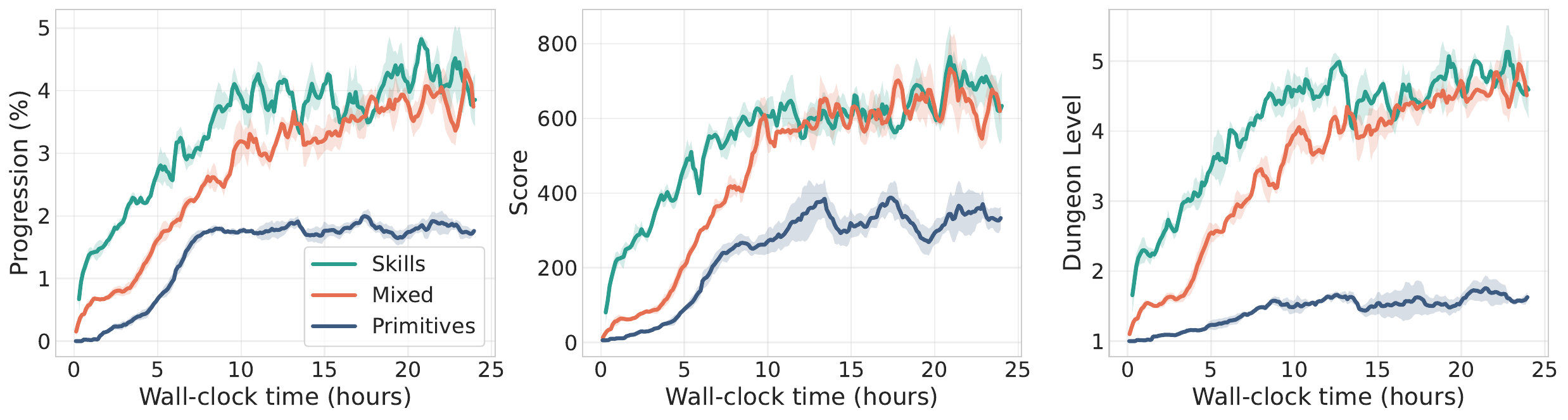}
    \caption{\textbf{Full training curves for \texttt{Qwen-3.5-4B} (wall-clock time).} We plot progression (left), score (middle), and dungeon level (right) vs. wall-clock time. All curves are averages across three seeds (except for mixed, which uses 2 seeds), and error bands indicate one standard error.}
    \label{fig:rl-progression-score-max-dlvl-vs-wall-clock-time-qwen}
\end{figure}

\cref{fig:rl-progression-score-max-dlvl-vs-grad-steps-llama,fig:rl-progression-score-max-dlvl-vs-wall-clock-time-llama} and~\cref{fig:rl-progression-score-max-dlvl-vs-grad-steps-qwen,fig:rl-progression-score-max-dlvl-vs-wall-clock-time-qwen} give the full RL training curves for \texttt{Llama-3.1-8B-Instruct} and \texttt{Qwen-3.5-4B}, respectively, all under the three action interfaces. The first figure plots learning against gradient steps, while the second replots the same runs against wall-clock time. Together they show that the qualitative ranking is stable under both views: skill-based and mixed controllers improve faster than primitive control throughout training, and the mixed and skill interfaces maintain a large performance advantage by the end of training. The action-distribution plots in~\crefrange{fig:rl-action-kind-share}{fig:rl-top-actions-primitives} complement that performance view by showing how RL changes the distribution of skills, primitives, and invalid outputs.

\clearpage

\subsection{Action-distribution analysis for Llama-3.1-8B-Instruct}
\label{app:rl-action-analysis}

\begin{figure*}[t]
    \centering
    \includegraphics[width=0.98\textwidth]{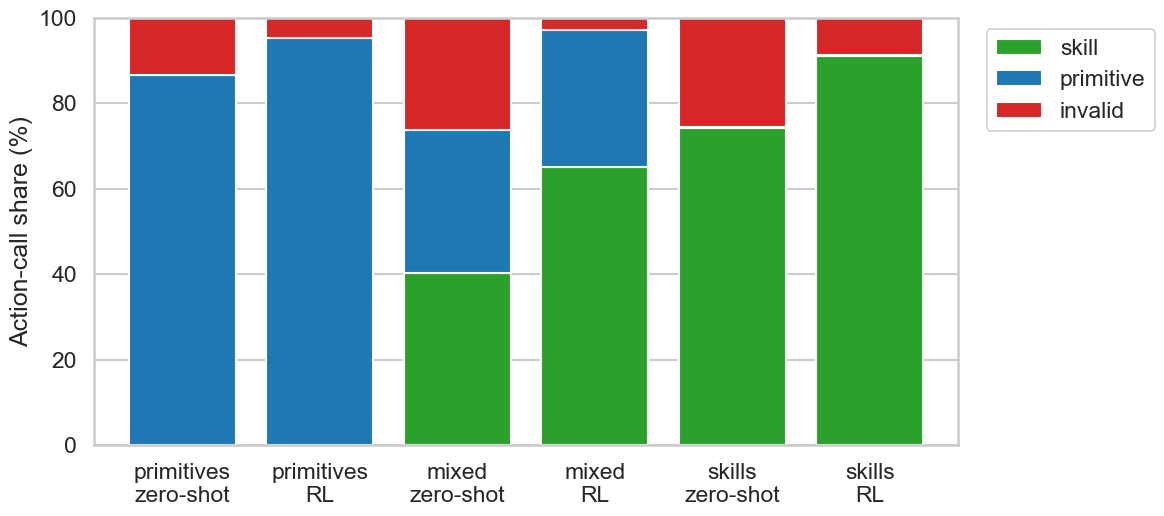}
    \caption{\textbf{Aggregate action-type mix before and after RL on Llama-3.1-8B-Instruct.} Bars show the share of parsed calls attributed to skills, primitives, and invalid outputs, aggregated across all evaluation episodes in each stage. RL shifts the mixed controller toward more successful skill use and sharply reduces invalid calls, but does not eliminate primitive fallback: the mixed primitive share remains close to one third of all calls.}
    \label{fig:rl-action-kind-share}
\end{figure*}

\begin{figure*}[t]
    \centering
    \includegraphics[width=0.75\textwidth]{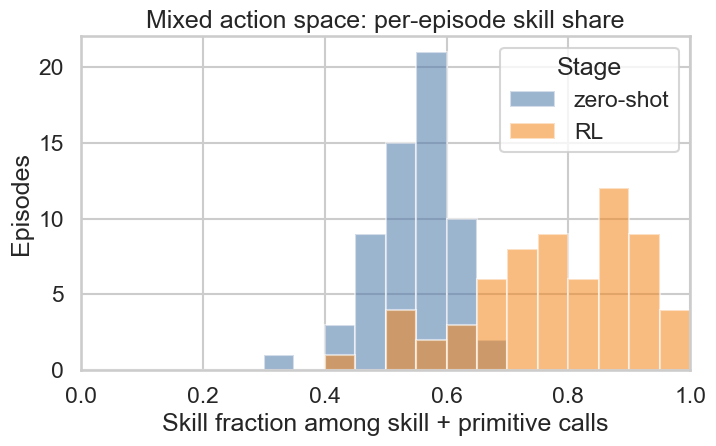}
    \caption{\textbf{Per-episode skill and primitive fractions before and after RL.} Each histogram is computed over per-episode call fractions derived from parsed action strings. We do not show primitive-only and skill-only interfaces, because the distributions collapse to 0 or 1 by construction. The important change is in the mixed interface: RL leaves the aggregate primitive share nearly unchanged (33.3\% to 32.0\%), but the typical episode becomes much more skill-heavy because the model outputs fewer invalid actions. The median mixed-episode skill fraction increases from 41.1\% to 77.7\%, while the median primitive fraction falls from 32.9\% to 20.2\%.}
    \label{fig:rl-episode-skill-primitive-hist}
\end{figure*}

\begin{figure*}[t]
    \centering
    \includegraphics[width=0.98\textwidth]{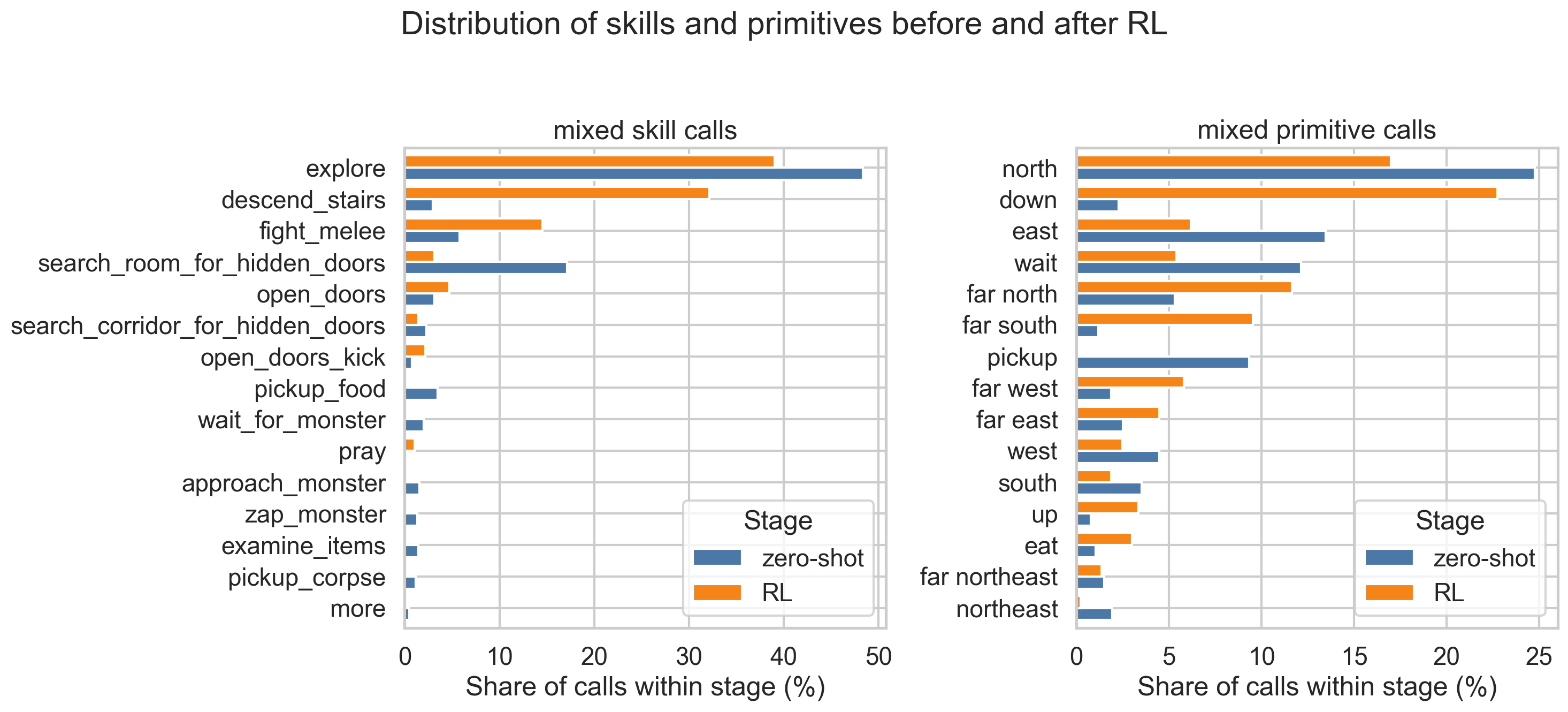}
    \caption{\textbf{Top mixed-controller skills and primitives before and after RL.} The left panel shows skill calls and the right panel shows primitive calls within the mixed interface. In each panel, we plot the top-$k$ call types with $k=15$, selected and ordered by the sum of their before-RL and after-RL call counts. After RL, mixed control concentrates its skill usage on a small core of reusable routines, especially \texttt{explore}, \texttt{descend\_stairs}, \texttt{fight\_melee}, \texttt{open\_doors}, and hidden-door search, while still retaining a nontrivial set of movement and staircase primitives.}
    \label{fig:rl-mixed-skill-vs-primitive-top-actions}
\end{figure*}

\begin{figure*}[t]
    \centering
    \includegraphics[width=0.65\textwidth]{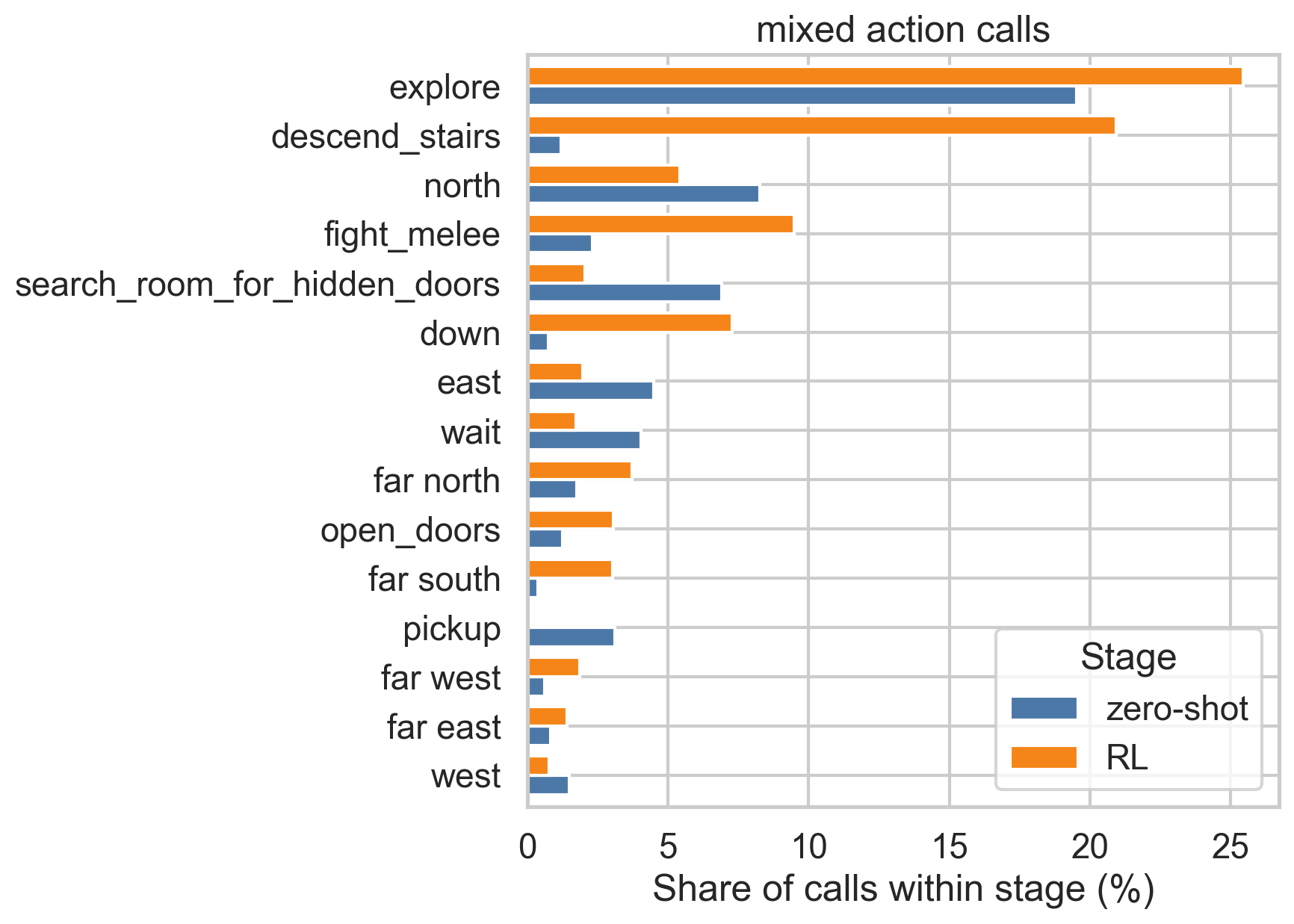}
    \caption{\textbf{Top action calls in the mixed interface before and after RL.} We plot the top-$k$ actions with $k=15$, selected and ordered by the sum of their before-RL and after-RL call counts within the mixed interface. When skills and primitives are pooled together, \texttt{explore} remains the single most common action both before and after training. RL elevates high-value skills such as \texttt{descend\_stairs} and \texttt{fight\_melee}, while several directional primitives decline in relative share, consistent with the controller relying more often on reusable skills rather than long sequences of local movement decisions.}
    \label{fig:rl-top-actions-mixed}
    \vspace{-2em}
\end{figure*}

\begin{figure*}[t]
    \centering
    \includegraphics[width=0.65\textwidth]{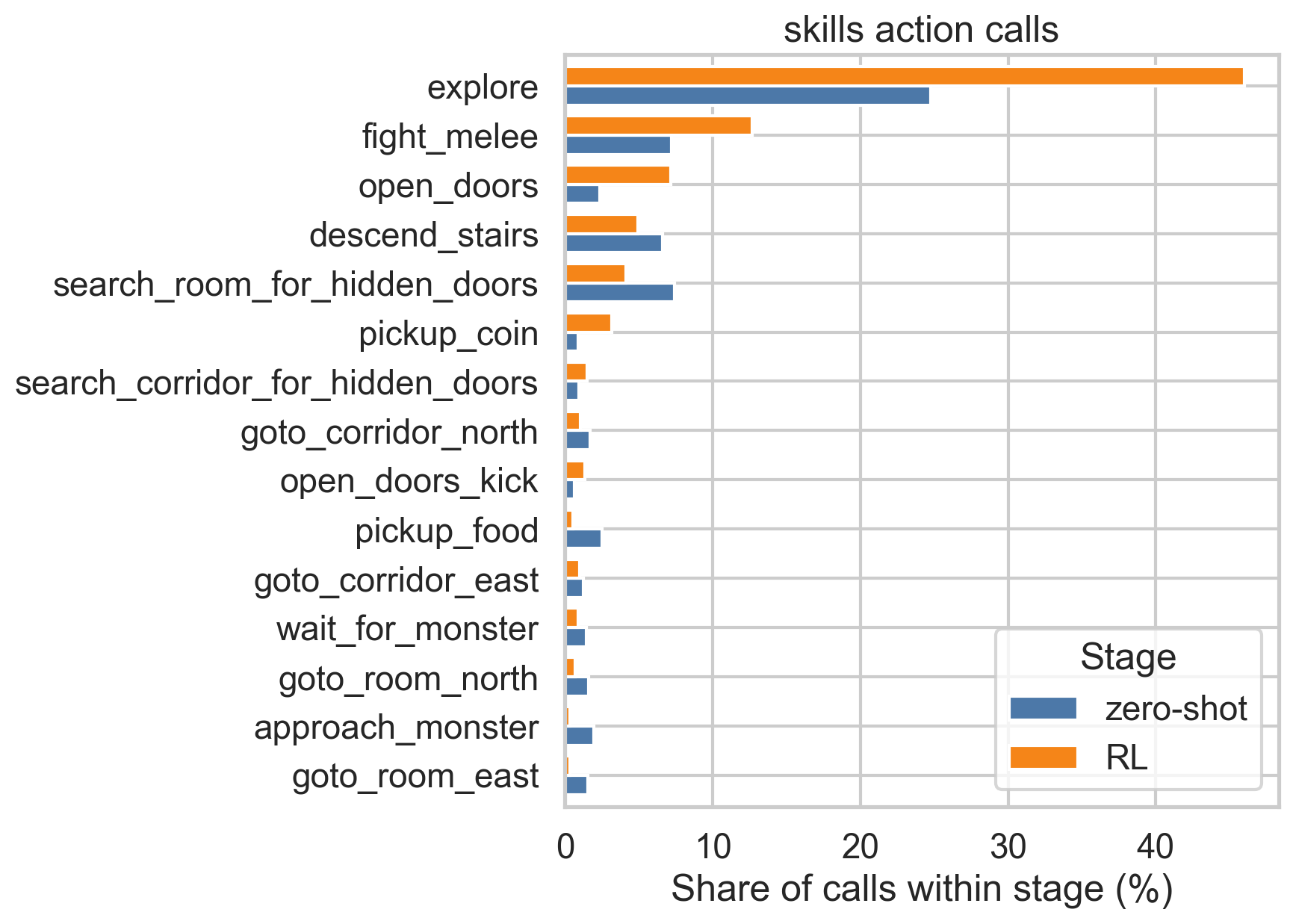}
    \caption{\textbf{Top action calls in the skill-only interface before and after RL.} We plot the top-$k$ actions with $k=15$, selected and ordered by the sum of their before-RL and after-RL call counts within the skill-only interface. Skill-only RL becomes even more concentrated on a handful of reusable routines, with \texttt{explore} dominating and combat, door handling, staircase navigation, and hidden-door search accounting for much of the remaining mass. This suggests that much of the learned improvement comes from invoking a small core of robust behaviors more effectively, rather than from broad uniform use of the entire library.}
    \label{fig:rl-top-actions-skills}
    \vspace{-2em}
\end{figure*}

\begin{figure*}[t]
    \centering
    \includegraphics[width=0.65\textwidth]{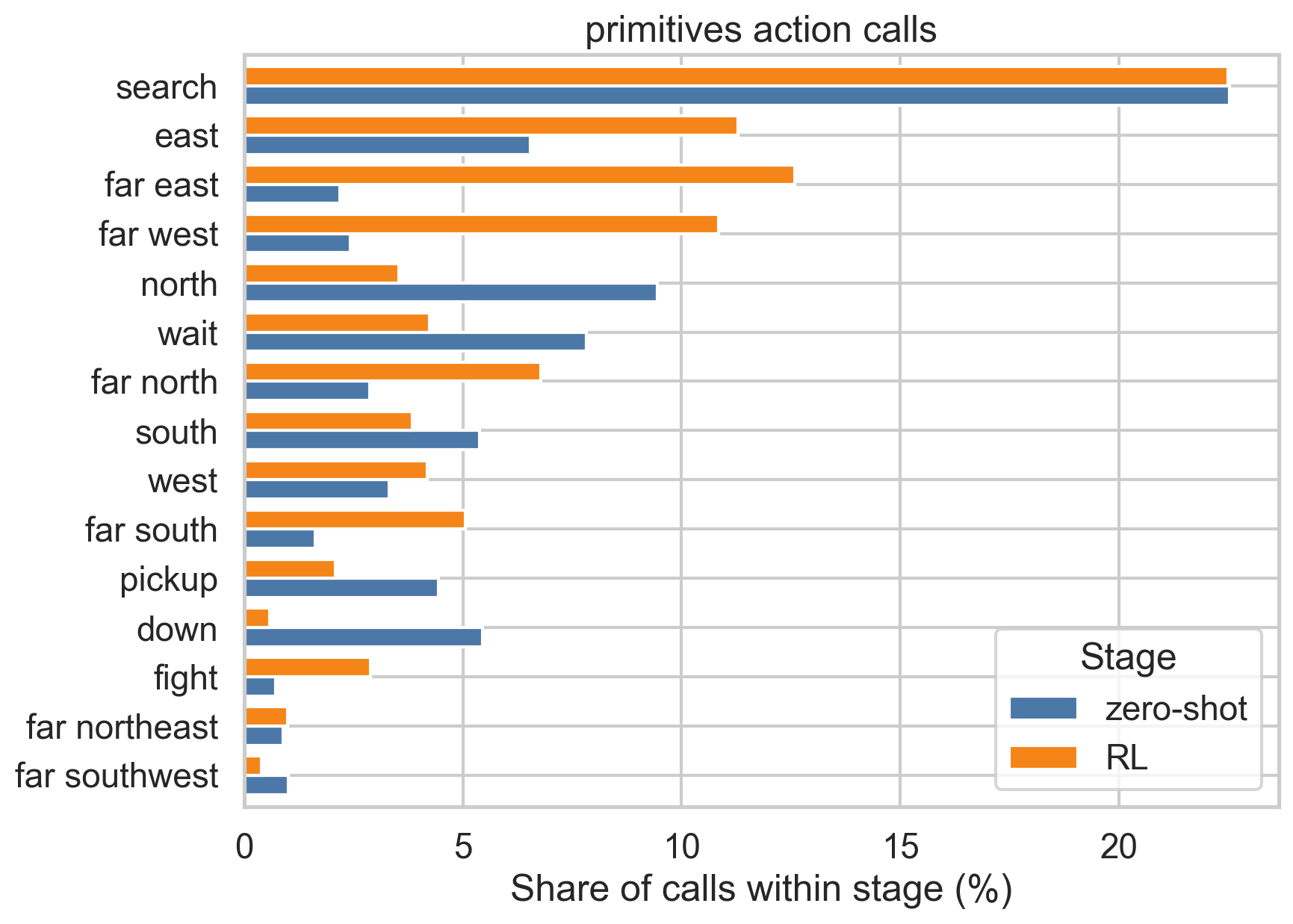}
    \caption{\textbf{Top action calls in the primitive-only interface before and after RL.} We plot the top-$k$ actions with $k=15$, selected and ordered by the sum of their before-RL and after-RL call counts within the primitive-only interface. Primitive-only control remains dominated by local movement and \texttt{search}. Training changes the mix of directions and reduces invalid calls (~\cref{fig:rl-action-kind-share}), but it does not create the same kind of reusable behavioral compression available in the skill-based interfaces.}
    \label{fig:rl-top-actions-primitives}
    \vspace{-2em}
\end{figure*}

To better understand how RL changes the use of the abstraction ladder, we compare parsed action distributions before and after RL for Llama-3.1-8B-Instruct.~\cref{fig:rl-action-kind-share,fig:rl-episode-skill-primitive-hist} summarize the aggregate and per-episode shift between skills, primitives, and invalid outputs, while~\crefrange{fig:rl-mixed-skill-vs-primitive-top-actions}{fig:rl-top-actions-primitives} break that shift down into the most frequent calls in the mixed, skill-only, and primitive-only interfaces. Three patterns are especially clear. First, mixed-control RL does not simply eliminate primitive use: by aggregate share, primitive calls remain almost unchanged in the mixed interface (33.3\% in zero-shot vs.\ 32.0\% after RL). Second, the typical mixed episode still becomes much more skill-heavy after training. The median skill fraction rises from 41.1\% to 77.7\%, while the median primitive fraction falls from 32.9\% to 20.2\%. Third, RL concentrates behavior around a small set of simple but important routines, while sharply reducing malformed or otherwise invalid calls.

\subsection{OOD MiniHack transfer for Llama-3.1-8B-Instruct}
\label{app:ood-minihack-transfer}

\begin{figure*}[t]
    \centering
    \includegraphics[width=0.75\textwidth]{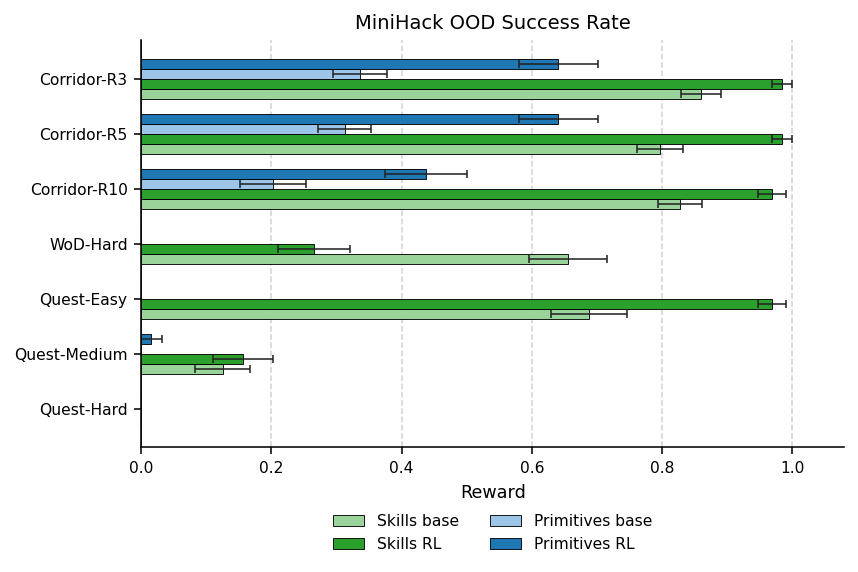}
    \caption{\textbf{OOD MiniHack transfer before and after NetHack RL.} Bars compare the Llama-3.1-8B-Instruct for primitives and skills before RL and after RL on held-out MiniHack tasks. We report average success rates. Both interfaces transfer from NetHack to MiniHack, but RL changes them differently: primitive gains are concentrated on the Corridor tasks, while skill-based RL pushes the Corridor family to near-perfect success and also improves the Quest tasks. The main exception is \texttt{WoD-Hard}, where the skill controller becomes worse after RL.}
    \label{fig:ood-minihack-success-rate}
\end{figure*}

~\cref{fig:ood-minihack-success-rate} compares MiniHack success before and after RL training on NetHack. RL improves average success for both action interfaces, but the gains differ across tasks. For primitive control, improvements are concentrated on the Corridor tasks, with one additional success on \texttt{Quest-Medium} after RL. For skill-based control, RL brings Corridor success close to saturation and improves \texttt{Quest-Easy} and \texttt{Quest-Medium}. However, \texttt{Quest-Hard} remains unsolved, and \texttt{WoD-Hard} success decreases. Averaged across tasks, success rises from 56.5\% to 61.8\% for skills and from 12.2\% to 24.8\% for primitives. Thus, primitives show a larger average absolute improvement, while skills retain substantially higher success overall.

\clearpage